\documentclass[preprint,12pt]{elsarticle}

\usepackage{amsmath,amssymb}
\usepackage{pdflscape}
\usepackage{algorithm,algpseudocode,float}
\usepackage{lipsum}
\usepackage{caption}
\usepackage{booktabs}
\usepackage{multirow}
\usepackage{bm}
\usepackage{xcolor}
\usepackage{changepage}
\usepackage{subcaption}
\usepackage{adjustbox}
\usepackage{array}
\usepackage{enumitem}

\usepackage[colorlinks=true, linkcolor=blue, urlcolor=blue, citecolor=blue]{hyperref}
\usepackage{cleveref}
\usepackage{circuitikz}

\biboptions{numbers,sort&compress}

\crefname{figure}{\protect\textcolor{blue}{Fig.}}{\protect\textcolor{blue}{Figs.}}
\DeclareCaptionLabelFormat{bold}{\textbf{#1~#2}}
\newcommand{\beq}{\begin{equation}}
	\newcommand{\eeq}{\end{equation}}
\newcommand{\beqa}{\begin{eqnarray}}
	\newcommand{\eeqa}{\end{eqnarray}}
\newcommand{\beqas}{\begin{eqnarray*}}
	\newcommand{\eeqas}{\end{eqnarray*}}
\newcommand{\ba}{\begin{array}}
	\newcommand{\ea}{\end{array}}
\newcommand{\bi}{\begin{itemize}}
	\newcommand{\ei}{\end{itemize}}

\makeatletter
\newenvironment{breakablealgorithm}
{% \begin{breakablealgorithm}
		\begin{center}
			\refstepcounter{algorithm}% New algorithm
			\hrule height.8pt depth0pt \kern2pt% \@fs@pre for \@fs@ruled
			\renewcommand{\caption}[2][\relax]{% Make a new \caption
				{\raggedright\textbf{\ALG@name~\thealgorithm} ##2\par}%
				\ifx\relax##1\relax % #1 is \relax
				\addcontentsline{loa}{algorithm}{\protect\numberline{\thealgorithm}##2}%
				\else % #1 is not \relax
				\addcontentsline{loa}{algorithm}{\protect\numberline{\thealgorithm}##1}%
				\fi
				\kern2pt\hrule\kern2pt
			}
		}{% \end{breakablealgorithm}
		\kern2pt\hrule\relax% \@fs@post for \@fs@ruled
	\end{center}
}
\makeatother

\journal{Alexandria Engineering Journal}

\begin{document}

\begin{frontmatter}

\title{Memory Enhanced Fractional-Order Dung Beetle Optimization for Photovoltaic Parameter Identification}

\author[xidian]{Yiwei Li} %% Author name
\author[xidian]{Zhihua Allen-Zhao\corref{cor1}} %% Author name
\ead{allenzhaozh@gmail.com}
\author[xidian]{Yuncheng Xu}
\author[xidian]{Sanyang Liu}
\cortext[cor1]{Corresponding author}

%% Author affiliation
\affiliation[xidian]{organization={School of Mathematics and Statistics, Xidian University},%Department and Organization
	city={Xi'an},
	postcode={710071},
	state={Shaanxi},
	country={China}}

%% Abstract
\begin{abstract}
Accurate parameter identification in photovoltaic (PV) models is crucial for performance evaluation but remains challenging due to their nonlinear, multimodal, and high-dimensional nature.
Although the Dung Beetle Optimization (DBO) algorithm has shown potential in addressing such problems,
it often suffers from premature convergence.
To overcome these issues,
this paper proposes a Memory Enhanced Fractional-Order Dung Beetle Optimization (MFO-DBO) algorithm that integrates three coordinated strategies.
Firstly, fractional-order (FO) calculus introduces memory into the search process, enhancing convergence stability and solution quality.
Secondly, a fractional-order logistic chaotic map improves population diversity during initialization.
Thirdly, a chaotic perturbation mechanism helps elite solutions escape local optima.
Numerical results on the CEC2017 benchmark suite and the PV parameter identification problem demonstrate that
MFO-DBO consistently outperforms advanced DBO variants, CEC competition winners, FO-based optimizers, enhanced classical algorithms,
and recent metaheuristics in terms of accuracy, robustness, convergence speed,
while also maintaining an excellent balance between exploration and exploitation compared to the standard DBO algorithm.

\end{abstract}

%% Keywords

\end{frontmatter}

\section{Introduction} \label{sec1}

Photovoltaic (PV) systems are pivotal components of contemporary renewable energy infrastructure,
and their precise modeling is essential for performance assessment under diverse environmental conditions \citep{song2025parameter}.
Predominant PV models, such as the single-diode model (SDM), double-diode model (DDM), and comprehensive PV module models, encapsulate the nonlinear current-voltage (I-V) characteristics of solar cells through critical parameters,
including photogenerated current, saturation current, series resistance, and shunt resistance.
Nonetheless, the estimation of these parameters poses significant challenges due to the inherent nonlinearity, multi-modal landscape, and high dimensionality of the optimization problem.

In practical, meta-heuristic algorithms have gained widespread traction, owing to their flexibility and robust global search capabilities \citep{yuan2023improved,wang2025dynamic}.
Notably, the Dung Beetle Optimization (DBO) algorithm, introduced by Xue and Shen~\cite{Xue2023},
has emerged as a potent tool for tackling complex real-world issues,
e.g., renewable energy system optimization~\citep{liu2024optimization}
and unmanned aerial vehicle (UAV) trajectory planning~\citep{Lyu2024}.
The algorithm's simplicity and rapid convergence rate position it as a promising contender for PV parameter extraction tasks.
However, akin to many bio-inspired algorithms,
DBO also suffers from premature convergence, which frequently leads
to search stagnation and an inability to escape from the local optimum
(see Subsection~\ref{subsec1.1} for more details).

\subsection{Research Status}  \label{subsec1.1}

To enhance the DBO algorithm, researchers have developed various modifications targeting its core mechanisms.
A synopsis of extant DBO variants is provided in Table~\ref{relatedbo},
offering insights into ongoing efforts to enhance its performance and adaptability.

Despite recent advancements in DBO and its variants, as summarized in Table~\ref{relatedbo},
several methodological and experimental limitations remain:

\begin{itemize} \setlength{\itemsep}{0pt}
\item \textbf{Q1. Lack of Long-Term Memory Integration}: Current DBO variants primarily rely on immediate states for decision-making,
    neglecting a broader historical context~\citep{Xue2023, yu2025multi,li2024enhanced}.
    For example, standard DBO algorithm~\cite{Xue2023} updated $x_i(t{+}1)$ only using $x_i(t)$ and $x_i(t{-}1)$  (see Eq.~\eqref{eq1}).
    This myopic behavior limits learning from extended search trajectories, increasing the risk of premature convergence to suboptimal solutions.

\begin{landscape}
	\begin{table}[htbp]
		\vspace{-1.5cm}
			\begin{adjustwidth}{0cm}{-0.5cm}
				\captionsetup{
						labelsep=newline,
						justification=raggedright,
						singlelinecheck=false,
						labelfont=bf,
						skip=0pt,
						margin=0pt
					}
				\caption{Related work on the improved DBO algorithms.}
				\label{relatedbo}
				\scriptsize
				\resizebox{22cm}{!}{%
					\begin{tabular}{llp{6cm}@{}p{6cm}p{5cm}}
							\toprule
							\multirow{2}{*}{Algorithm} & \multirow{2}{*}{Reference} & \multicolumn{3}{l}{Improvement Measures} \\
							\cmidrule{3-5}
							& & Population Initialization Strategies & Global Search Enhancement Strategies & Local Optimum Escape Strategies \\
							\midrule
							IDBO & Zhou et al.~\citep{zhouyazhong2023power} & - & Periodic Mutation Mechanism & - \\
							\midrule
							EDBO & Yu et al.~\citep{yu2025multi} & - & Optimal value guidance, Nonlinear dynamic adjustment, Adaptive boundary control and Enhanced foraging strategy & - \\
							\midrule
							EDBO & Li et al.~\citep{li2024enhanced}  &  -  & Elite-guided Coupled Elimination Strategy and Stochastic Elite-Guided Directional Learning  & Jacobian Escape Curve Mechanism \\
							\midrule
							MsDBO &  Chen et al.~\citep{chen2025coverage} &  -   & Optimal Boundary Control Strategy and  Stochastic Inertia Adjustment Mechanism  &  Triple-Point Fusion Search Strategy\\
							\midrule
							GODBO & Wang et al.~\citep{zilong2023multi}   & Opposition-Based Learning   &  Gbest-Guided Directional Refinement    &  -  \\
							\midrule
							MDBO &  Shen et al.~\citep{shen2023multi} & Beta Distribution & L\'{e}vy Distribution-Based Boundary Handling and Two Crossover Operator Strategy & - \\
							\midrule
							IDBO & Gao et al.~\citep{gao2025optimizing} & Chebyshev Chaotic Map & Optimized Convergence Factor Strategy and Golden Sine Strategy & - \\
							\midrule					
							ODBO &  Wang et al.~\citep{wang2024improved} & Cat Map and Opposition-Based Learning & Osprey-DBO Hybrid Search Strategy and Vertical-Horizontal Crossover Strategy & - \\
							\midrule		
							IDBO &  Fu et al.~\citep{fu2024lithium} & SPM Chaotic Map & Golden Sine Strategy & Adaptive Gaussian-Cauchy Mutation Perturbation \\
							\midrule
							IDBO &  Wang et al.~\citep{wang2024prediction} & Sine Chaotic Map & Osprey Optimization Integration & Adaptive $t$-Distribution Dynamic Selection \\
							\midrule
							IMODBO &  Tu et al.~\citep{tu2023imodbo} & Chaotic Tent Map & Adaptive Weight Factor Strategy and Variable Spiral Local Search Strategy & L\'{e}vy Flight Perturbation \\
							\midrule
							IDBO &  Lyn et al.~\citep{Lyu2024} & Cube Chaotic Map & Global Exploration Strategy and Stage-Specific Population Update Strategy & Adaptive $t$-Distribution \\
							\midrule		
							MDBO &  Ye et al.~\citep{ye2024multi} & Latin Hypercube Sampling & Mean Differential Variation Strategy & Lens Imaging Reverse Learning with Dimension-Wise Optimization \\
							\midrule
							QHDBO &  Zhu et al.~\citep{ZHUfang2024} & Good Point Set Strategy & Convergence Factor and Dynamic Spawning-Foraging Balance Strategy & Quantum $t$-Distribution Variation \\
							\midrule
							QOLDBO &  Wang et al.~\citep{wang2023quasi} & Quantum Quasi-Opposition Learning & Q-Learning and Variable Spiral Local Domain Method & Dimensional Adaptive Gaussian Variation \\
							\midrule
							MFO-DBO& Our paper  & Fractional-Order Logistic Chaotic Map & Fractional-Order Calculus Integration Strategy  & Logistic Chaotic perturbation\\
							\bottomrule
						\end{tabular}
				}
				\end{adjustwidth}
		\end{table}
\end{landscape}

\item \textbf{Q2. Inadequate Population Initialization}: Population initialization plays a critical role in determining the global search performance of metaheuristic algorithms~\citep{ZHUfang2024}.
    Although integer-order chaotic map strategies have been widely used in DBO variants,
    they often fail to ensure diversity and uniform distribution in high-dimensional spaces~\citep{gao2022review}.
    This impairs global exploration capabilities, leading to sluggish convergence rates and impaired search space navigation.

\item \textbf{Q3. Insufficient Perturbation Mechanisms}: While adaptations like $t$-distribution and adaptive Gaussian-Cauchy perturbations aim to enhance exploration,
    they frequently lack the necessary randomness to escape local optima, especially in complex or high-dimensional scenarios (see Subsection~\ref{subsubsec4.2.3} for more details).
    This undermines the algorithm's ability to explore new regions.

\item \textbf{Q4. Absence of Ablation Studies}: Hybrid DBO variants often implement multiple enhancements without isolating their individual effects~
    \citep{fu2024lithium, wang2024prediction}.
    This omission complicates the identification of effective components and hinders the strategic design of synergistic combinations.
\end{itemize}

\subsection{Our Contributions} \label{subsec1.2}

To address these issues, this paper proposes a novel DBO variant,
termed Memory Enhanced Fractional-Order Dung Beetle Optimization (MFO-DBO).
By integrating fractional-order (FO) calculus, fractional-order logistic chaotic (FOLC) mapping, and chaotic perturbation (CP),
MFO-DBO overcomes DBO's drawbacks and boosts optimization performance in benchmarks and real-world PV tasks.

First of all, for \textbf{Q1},
FO calculus is employed to incorporate long-term memory effects,
allowing the search process to utilize information from beyond the latest iterations.
By leveraging an infinite series of historical states, it improves convergence performance,
unlike integer-order derivatives that only account for finite past states.
As verified in Subsection~\ref{subsubsec4.2.2}, ten fractional-order variants surpass the original DBO, validating the effectiveness of FO calculus.
Secondly, for \textbf{Q2}, FOLC mapping optimizes population initialization by maintaining chaotic diversity, which enhances search coverage in high-dimensional spaces.
As demonstrated in Subsection~\ref{subsubsec4.2.3},
FOLC-based initialization surpasses integer-order chaotic strategies;
the latter degrades DBO performance in high-dimensional settings.
Thirdly, for \textbf{Q3}, CP applies adaptive stochastic perturbations to elite individuals,
balancing the escape from local optima and convergence speed.
As shown in Subsection~\ref{subsubsec4.2.3}, compared with adaptive $t$-distribution and Gaussian–Cauchy perturbations, CP holds better performance,
especially in high-dimensional problems.
Finally, for \textbf{Q4}, ablation studies will be carried out
to assess each enhancement's individual and combined impacts, disentangling their standalone contributions and synergistic benefits.

In summary, all the above enhancements collectively strengthen the DBO's performance (see Section~\ref{sec4})
and significantly improve its effectiveness in real-world PV parameter identification tasks
(see Section~\ref{sec5}).

This paper has two main contributions:

\begin{itemize} \setlength{\itemsep}{0pt}
\item We propose a novel algorithmic framework, termed MFO-DBO, by several targeted enhancements that markedly elevate DBO's performance, including convergence speed, precision, and robustness.
    Numerical results show that our MFO-DBO outperforms not only advanced DBO variants and CEC competition winners but also fractional-order algorithms, enhanced conventional techniques and some recent metaheuristic methods.
    %It implies that MFO-DBO exhibits a better equilibrium between exploration and exploitation capabilities than the classical DBO.

\item MFO-DBO outperforms existing methods in both accuracy and stability for photovoltaic (PV) model parameter identification.
    Compared to the classical DBO algorithm, it achieves significant reductions in root mean square error (RMSE):
    8.90\% for the Single-Diode Model (SDM), 5.17\% for the Double-Diode Model (DDM), and a substantial 21.14\% for the PV module model.
    These results demonstrate strong real-world applicability and generalization beyond synthetic benchmarks.
	
\end{itemize}

The outline of this paper is as follows.
Section~\ref{sec2} briefly reviews the DBO algorithm.
Section~\ref{sec3} presents the three enhancements and our MFO-DBO algorithm.
Section~\ref{sec4} evaluates MFO-DBO on the CEC2017 benchmark suite.
Section~\ref{sec5} applies MFO-DBO to the parameter identification problem in photovoltaic (PV) models.
Finally, some concluding remarks are presented in Section~\ref{sec6}.

\section{Preliminaries} \label{sec2}

In this section, we review the four core components of the DBO algorithm,
and its basic iterative framework reported in Algorithm~\ref{DBO}.

The Dung Beetle Optimization (DBO) algorithm simulates the foraging and reproductive behaviors of dung beetles through four agent types:
ball-rolling beetles, brood balls, small dung beetles, and thieves.
Each subgroup follows a distinct update strategy, as summarized below.

\begin{itemize} \setlength{\itemsep}{0pt}
\item \textbf{Ball-rolling beetles}: update the positions by
\begin{equation}\label{eq1}
	\left\{\begin{array}{l}
		x_i(t+1)=x_i(t)+\alpha \cdot k \cdot x_i(t-1)+ b \cdot \Delta x, \\[0.1cm]
		\Delta x=\left|x_i(t)-X^\omega\right|,
	\end{array}\right.
\end{equation}
where $x_i(t) \in \mathbb{R}^n$ is the position of dung beetle $i$ at iteration $t$;
$\alpha \in \{1, -1\}$ controls trajectory retention or deviation;
$k \in (0, 0.2]$ is the deflection coefficient (typically 0.1);
$b \in [0, 1]$ adjusts step size (usually 0.3);
$\Delta x$ is the distance to the worst solution $X^\omega \in \mathbb{R}^n$.

When encountering obstacles, a tangent-based deflection is applied by
\begin{equation}\label{eq2}
	x_i(t+1)=x_i(t)+\tan \theta \cdot \left|x_i(t)-x_i(t-1)\right|,
\end{equation}
where $\theta \in [0, \pi]$ is a deflection angle.

\item \textbf{Brood balls}: update their positions by
\begin{equation}\label{eq3}
	X_i(t+1)=X^*+B_1 \circ \left(X_i(t)-L b^*\right)+B_2 \circ \left(X_i(t)-U b^*\right),
\end{equation}
where $B_1, B_2 \sim \mathcal{U}(\bm{0}, \bm{I})$ with $\bm{0} \in \mathbb{R}^n$
and identity matrix $\bm{I} \in \mathbb{R}^{n \times n}$;
$x \circ y = (x_1 y_1, \cdots, x_n y_n)^T$ for any $x, y \in \mathbb{R}^n$;
$X^* \in \mathbb{R}^n$ denote the local best positions;
$Lb^*, Ub^* \in \mathbb{R}^n$ are the lower and upper bounds of the brood ball spawning region, respectively.

\item \textbf{Small dung beetles}: update their positions by
\begin{equation}\label{eq4}
	x_i(t+1)=x_i(t)+c_1 \cdot \left(x_i(t)-L b^b\right)+c_2 \cdot \left(x_i(t)-U b^b\right),
\end{equation}
where $c_1 \sim \mathcal{N}(0,1)$ and $c_2 \sim \mathcal{U}(0,1)$;
$Lb^b$ and $Ub^b \in \mathbb{R}^n$ are the lower and upper bounds of foraging area, respectively.

\item \textbf{Thieves}: update their positions by
\begin{equation}\label{eq5}
	x_i(t+1)=X^b+ d \cdot G \circ \left(\left|x_i(t)-X^*\right|+\left|x_i(t)-X^b\right|\right),
\end{equation}
where $G \sim \mathcal{N}(\bm{0}, \bm{I})$;
$d > 0$ is a constant; $X^b \in \mathbb{R}^n$ is the global best solution.
\end{itemize}

Therefore, the classic DBO algorithm is outlined below.
\begin{breakablealgorithm}
	\caption{The iteration of the classic DBO algorithm}
	\label{DBO}
	\begin{algorithmic}[1]
		\State \textbf{Input:} Maximum iterations $T_{\max}$; population size $N$
		\State \textbf{Output:} Best solution $X^b$ and fitness $f_b$
		\State Initialize population $\{x_i\}_{i=1}^N$
		\State Evaluate fitness and set initial best $X^b$, $f_b$
		\While{$t \leq T_{\max}$}
		\For{each individual $i = 1$ to $N$}
		\If{$i$ is ball-rolling beetle}
		\State with probability 0.9, update using Eq.~\eqref{eq1}; else use Eq.~\eqref{eq2}
		\ElsIf{$i$ is brood ball}
		\State update using Eq.~\eqref{eq3}
		\ElsIf{$i$ is small dung beetle}
		\State update using Eq.~\eqref{eq4}
		\ElsIf{$i$ is thief}
		\State update using Eq.~\eqref{eq5}
		\EndIf
		\State update $X^b$ and $f_b$ if improved
		\EndFor
		\EndWhile
		\State \Return $X^b$, $f_b$
	\end{algorithmic}
\end{breakablealgorithm}
For more details, please refer to \cite{Xue2023,gao2025optimizing,tu20243d}.

\section{The MFO-DBO Algorithm} \label{sec3}

In this section, we propose an enhanced DBO algorithm,
termed Memory Enhanced Fractional-Order Dung Beetle Optimization (MFO-DBO),
by the three major enhancements:
(1) Fractional-Order (FO) calculus to enhance historical information utilization;
(2) Fractional-Order Logistic Chaotic (FOLC) mapping to optimize initial population distribution;
(3) Chaotic Perturbation (CP) to assist the population in escaping local optima.

\subsection{Fractional-order calculus} \label{subsec3.1}

Unlike classical integer-order calculus,
Fractional-Order (FO) calculus incorporates memory and hereditary effects,
enabling systems to retain historical information over time.
These properties make FO calculus highly effective for optimization,
where past experiences can guide the search process more intelligently.
The classic DBO algorithm, which relies solely on recent states,
often suffers from premature convergence and suboptimal solutions.
By integrating FO calculus into DBO, we introduce a more adaptive and robust search mechanism,
effectively reducing stagnation and enhancing global search performance.

\begin{figure}[htp]
	\centering
	\includegraphics[width=0.72\textwidth]{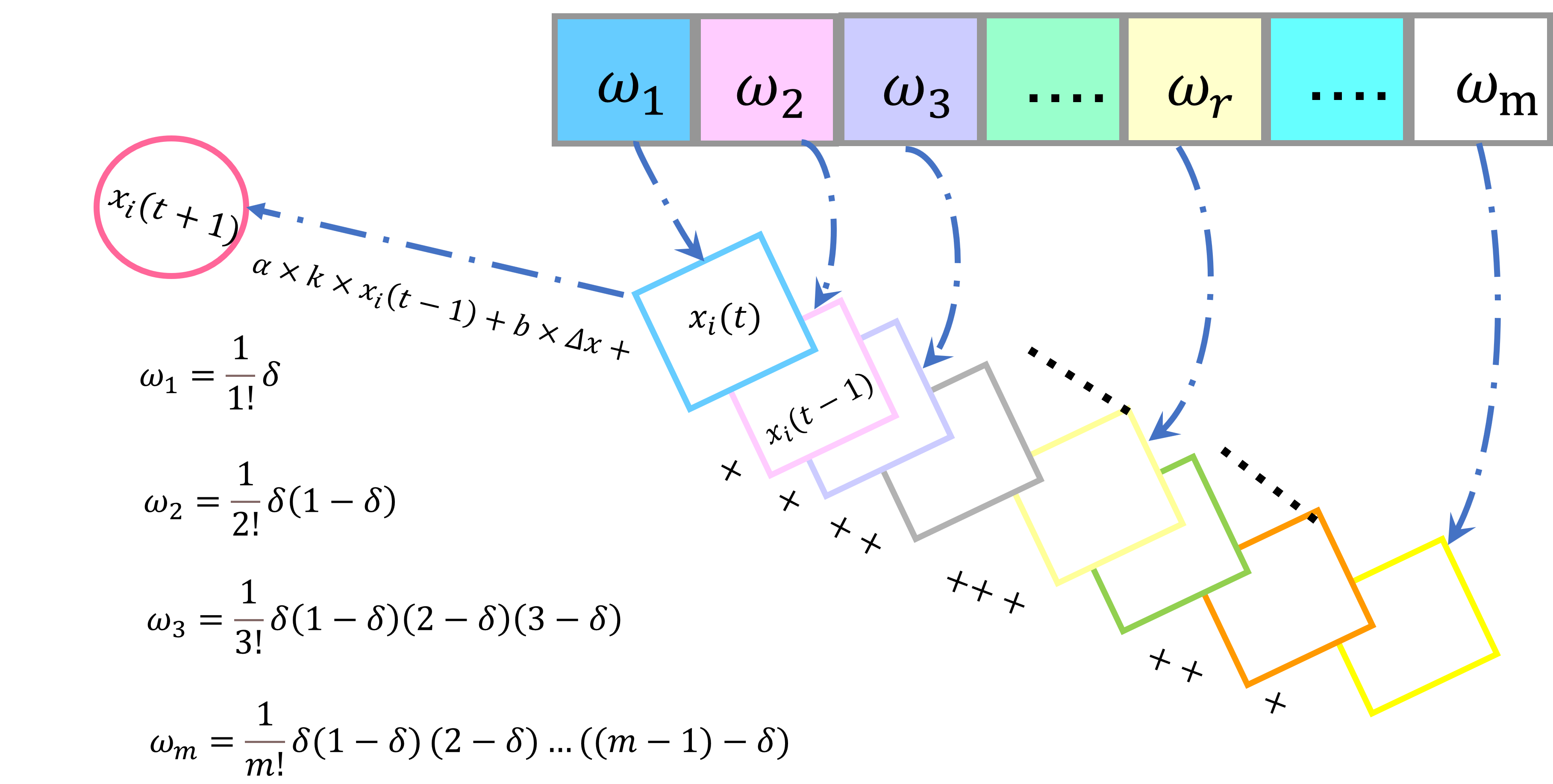}
	\caption{Memory properties of FO calculus.}
	\label{figFOcalculus}
\end{figure}

To do it, we utilize the Gr\"{u}nwald-Letnikov (GL) fractional derivative,
a method compatible with discrete applications in swarm intelligence algorithms.
The GL derivative's discrete-time formulation~\citep{YOUSRI2020} is as follows:
\begin{equation}\label{eq6}
	D^\delta(x(t))=\frac{1}{T^\delta} \sum_{k=0}^m \frac{(-1)^k \Gamma(\delta+1) x(t-k T)}{\Gamma(k+1) \Gamma(\delta-k+1)},
\end{equation}
where $\delta \in (0,1]$ is the order of the fractional derivative of $x(t)$ with respect to $t$;
$T>0$ is the sampling period;
$\Gamma(\cdot)$ denotes the gamma function;
$m \in N^{+}$ is the length of the memory terms.
Specially, when $\delta=1$, Eq.~\eqref{eq6} reduces to
\begin{equation}\label{eq7}
	D^1(x(t))=x(t+1)-x(t),
\end{equation}
where $D^1(x(t))$ represents the difference between two neighboring events.
Together with Eq.~\eqref{eq1} and Eq.~\eqref{eq7},
a special matching case of the GL derivative can be reformulated as
\begin{equation}\label{eq8}
	D^1\left(x_i(t+1)\right)=x_i(t+1)-x_i(t)=\alpha \cdot k \cdot x_i(t-1)+b \cdot \Delta x.
\end{equation}
Then, for the general case in view of the GL definition, we easily obtain
\begin{equation}\label{eq9}
	D^\delta\left(x_i(t+1)\right)=\alpha \cdot k \cdot x_i(t-1)+b \cdot \Delta x.
\end{equation}
Using the discrete form of the GL definition of Eq.~\eqref{eq6} at $T=1$,
Eq.~\eqref{eq9} can be reduced to
\begin{equation}\label{eq10}
	\begin{aligned}
		D^\delta\left(x_i(t+1)\right) & =   x_i(t+1)+\sum_{k=1}^m \frac{(-1)^k \Gamma(\delta+1) x_i(t+1-k)}{\Gamma(k+1) \Gamma(\delta-k+1)} \\
		& =\alpha \cdot k \cdot x_i(t-1)+b \cdot \Delta x.
	\end{aligned}
\end{equation}
Thereby, the general formulation of MFO-DBO solutions,
incorporating the memory perspective of FO calculus, can be reformulated as
\begin{equation} \label{eq11}
	x_i(t+1)=-\sum_{k=1}^m \frac{(-1)^k \Gamma(\delta+1) x_i(t+1-k)}{\Gamma(k+1) \Gamma(\delta-k+1)}+\alpha \cdot k \cdot x_i(t-1)+b \cdot \Delta x.
\end{equation}
This reformulation indicates that the movement of rolling dung beetles in the MFO-DBO algorithm
is influenced by their historical trajectory over a span of $m$ previous steps,
as illustrated in \autoref{figFOcalculus}.
For instance, when $m=4$, the position updates of rolling dung beetles are refined as
\begin{equation}\label{eq12}
	\begin{aligned}
		x_i(t+1) =  & \frac{1}{1!} \delta x_i(t)+\frac{1}{2!} \delta(1-\delta) x_i(t-1)+\frac{1}{3!} \delta(1-\delta)(2-\delta) x_i(t-2) \\
		& +\frac{1}{4!} \delta(1-\delta)(2-\delta)(3-\delta) x_i(t-3) +\alpha \cdot k \cdot x_i(t-1)+b \cdot \Delta x.
	\end{aligned}
\end{equation}
The choice of $m=4$ follows prior studies~\cite{Couceiro2012,yousri2022fractional}, which demonstrated that using four memory steps offers a good trade-off between historical information depth and computational efficiency in fractional-order metaheuristics.

\vspace{-0.2cm}

\subsection{Fractional-order logistic chaotic map} \label{subsec3.2}

The initial population distribution significantly affects the convergence and efficiency of DBO algorithms~\citep{ZHUfang2024, wang2024improved}.
Traditional initialization strategies, including random uniform sampling and integer-order chaotic maps,
have been widely applied in DBO variants.
However, in high-dimensional spaces, these methods often fail to provide sufficient diversity and uniform coverage of the search space, resulting in inefficient or biased searches~\citep{gao2022review}.
To mitigate it, fractional-order chaotic maps have been employed to enhance the initialization phase of swarm intelligence optimization algorithms.
Fractional-order chaotic systems leverage historical states to achieve superior ergodicity and memory-dependent dynamics \citep{PENG201996},
thereby improving adaptive exploration and reducing premature convergence.
This approach ensures comprehensive search-space coverage and enhanced population diversity.

\begin{figure}[H]
	\centering
	\includegraphics[width=0.70\textwidth]{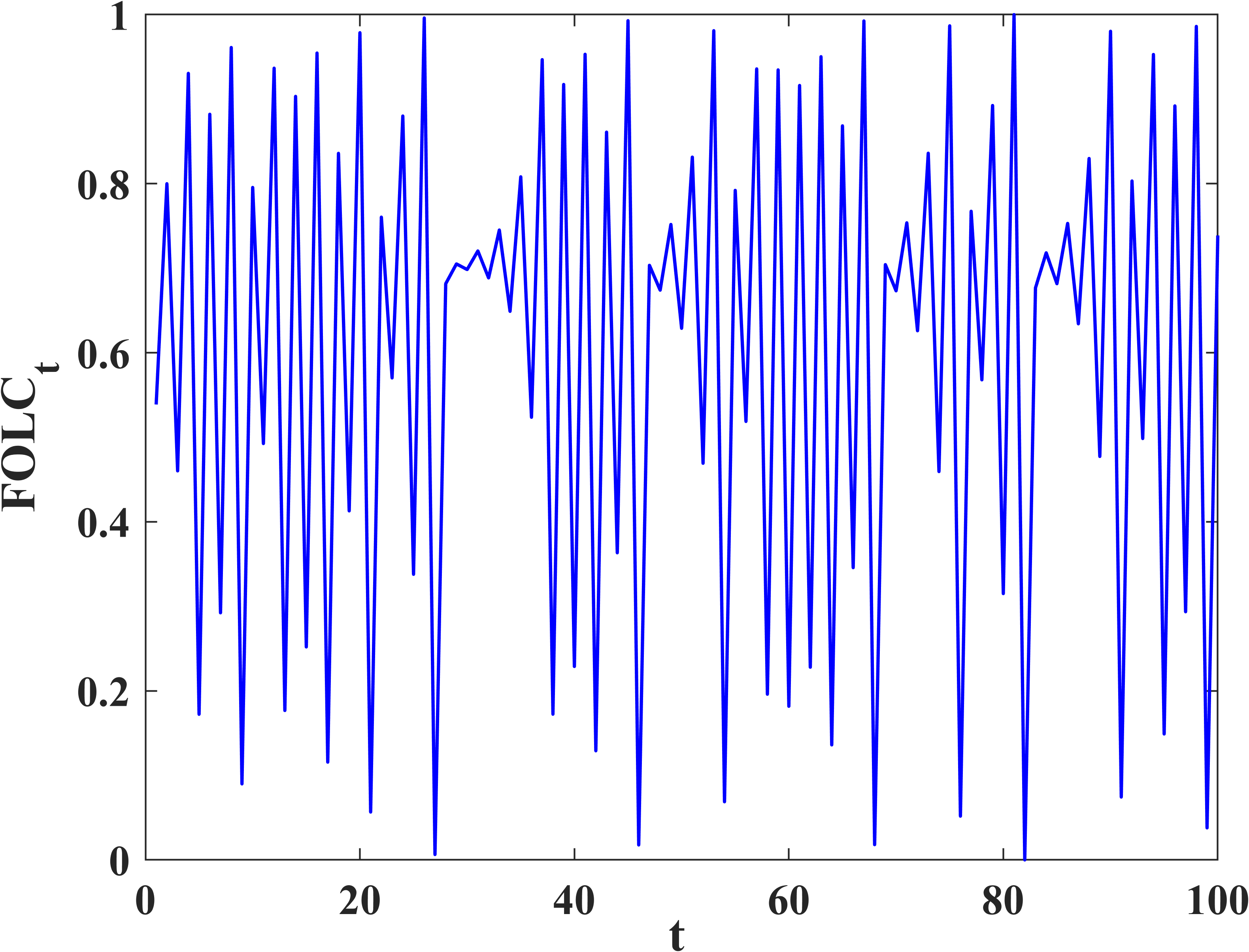}
	\caption{Fractional-order logistic chaotic map.}
	\label{figchaos}
\end{figure}

Thus, to ensure a high-quality, well-distributed initial population,
we employ a fractional-order logistic chaotic (FOLC) map~\citep{YOUSRI2020116979,wu2024data}.
\autoref{figchaos} illustrates the dynamic properties of this selected FOLC map, defined mathematically as follows:
\begin{equation}\label{eq13}
	\text{FOLC}_{t}=\text{FOLC}_{0}+\frac{\mu}{\Gamma(v)} \sum_{j=1}^t \frac{\Gamma(t-j+v)}{\Gamma(t-j+1)} \cdot \text{FOLC}_{j-1} \cdot (1-\text{FOLC}_{j-1}),
\end{equation}
where $\text{FOLC}_{t} \in (0,1)$ is the fractional-order logistic chaotic value at time $t$;
$\mu \in (0,4]$ is a control parameter;
$v \in (0,1]$ is the order of the fractional derivative.

Then, the initialization can be specifically formulated as
\begin{equation}\label{eq14}
	x_{i}= Lb + \text{FOLC} \circ \left(Ub-Lb\right),
\end{equation}
where $x_i\in \mathbb{R}^n$ is the position vector of the $i$-th dung beetle;
$\text{FOLC} = \left(\text{FOLC}_{1}, \text{FOLC}_{2}, \cdots, \text{FOLC}_{n}\right)^T \in \mathbb{R}^n$ is the fractional-order logistic chaotic vector with $\text{FOLC}_{i}$ derived by Eq.~\eqref{eq15}.

\subsection{Chaotic perturbation} \label{subsec3.3}

The DBO algorithm, while effectively balancing exploration and exploitation,
often becomes trapped in local optima, resulting in premature convergence~\citep{Xue2023}.
To overcome this limitation, various perturbation strategies—such as Gaussian, Cauchy, or adaptive distributions—have been proposed.
However, these conventional mechanisms often perform poorly in high-dimensional spaces due to their limited ability to induce diverse and globally effective search behaviors (see Section~\ref{sec4}).
To tackle this challenge more effectively, we introduce a Chaotic Perturbation (CP) mechanism.
Unlike standard random perturbations, CP utilizes the nonlinearity, ergodicity,
and sensitivity inherent in chaotic sequences to significantly improve the algorithm's exploration of new search areas and enhance solution diversity \citep{wang2017toward}.

The CP mechanism in this paper employs a logistic chaotic mapping to generate highly random and unpredictable sequences.
The resulting iterative equation is defined by
\begin{equation}\label{eq15}
	\text{chaos}_{t+1}=\mu \cdot \text{chaos}_{t} \cdot \left(1-\text{chaos}_{t}\right),
\end{equation}
where $\text{chaos}_{t} \in (0,1)$ is the logistic chaotic value at time $t$;
$\mu \in (0,4]$ is a control parameter.
By using Eq.~\eqref{eq15}, the chaotic perturbation operation can be formulated as
\begin{equation}\label{eq16}
	x_{cp}=x_b+\xi \cdot \text{chaos},	
\end{equation}
where $x_b \in \mathbb{R}^n$ is the position vector of the current best solution;
$x_{new}\in \mathbb{R}^n$ is the newly generated solution;
$\text{chaos} = \left(\text{chaos}_{1}, \text{chaos}_{2}, \cdots, \text{chaos}_{n}\right)^T \in \mathbb{R}^n$ is the fractional-order logistic chaotic vector with $\text{chaos}_{i}$ derived by Eq.~\eqref{eq15};
$\xi \in \mathbb{R}$ is a dynamically updated weighting coefficient, computed by
\begin{equation}\label{eq17}
	\xi=\log\text{sig}\left(\left( t_{\max } / 2- t\right) / K\right) \cdot \operatorname{rand}(0,1),
\end{equation}
$\text{logsig}(\cdot)$ is the log-sigmoid transfer function,
mapping values to the range (0,1);
$t_{\max}, t \in \mathbb{R}^{+}$ are the maximum and current iteration numbers, respectively;
The parameter $K > 0$ controls the perturbation intensity and is typically chosen from the interval $(0, t_{\max})$.
Following the strategy in~\cite{wang2017toward} and supported by robustness analysis, $K$ is set to 20 in this work.
This design allows stronger perturbations in the early phase to enhance global exploration,
and weaker ones in the later phase to support local refinement.

The chaotic perturbation method generates a new candidate solution during optimization.
It is accepted if it surpasses the current best solution;
Otherwise, the original solution is retained for the next iteration.
This approach balances local exploitation and global exploration, reduces premature convergence risks,
and enhances the search capability of DBO.

Finally, by incorporating FO calculus, FOLC mapping, and CP,
we propose the Memory Enhanced Fractional-Order Dung Beetle Optimization (MFO-DBO) algorithm,
which is summarized in Algorithm~\ref{alg2}.

\begin{breakablealgorithm}
	\caption{The iteration of the MFO-DBO algorithm}\label{alg2}
	\begin{algorithmic}[1]
		\Statex \textbf{Input:} Maximum iteration number $T_{\text{max}}$, population size $N$
		\Statex \textbf{Output:} Optimal position $X^b$ and its fitness value $f_b$
		\State Initialize the dung beetle population using the FOLC in Eq.~\eqref{eq14}
		\While{$t < T_{\text{max}}$}
		\For{$i= 1:N $}
		\If{$i $ == ball-rolling dung beetle}
		\State Generate a random number $\gamma = \text{rand}(1)$
		\If{$\gamma < 0.9$}
		\State update search position using FO calculus per Eq.~\eqref{eq12}
		\Else  ~~update search position using Eq.~\eqref{eq2}
		\EndIf
		\ElsIf{$i$ ==brood ball}
		\State update position using Eq.~\eqref{eq3}
		\ElsIf{$i$ == small dung beetle}
		\State update position using Eq.~\eqref{eq4}
		\ElsIf{$i$ == thief}
		\State update position using Eq.~\eqref{eq5}
		\EndIf
		\State Apply the CP to $X^b$ using to Eq.~\eqref{eq16} to generate $X_{cp}$
		\If{$f(X_{cp}) < f_b$}
		\State $X^b \gets X_{cp}$, \quad $f_b \gets f(X_{cp})$
		\EndIf
		\State update the best position $X^b$ and its fitness$f_b$
		\EndFor
		\State $t \gets t + 1$
		\EndWhile
		\State \Return $X^b$ and its fitness value $f_b$
	\end{algorithmic}
\end{breakablealgorithm}

\subsection{Computational complexity of MFO-DBO algorithm} \label{subsec3.4}

Next, we analyze the computational complexity of the MFO-DBO algorithm.
The complexity is influenced by three key factors: FOLC, FO calculus, and CP.
Thus, the overall complexity can be expressed as

\begin{equation} 	\label{eq18}
	\begin{aligned}
		O(\text{MFO-DBO}) & = O(\text{FOLC})+O(\text{FO} \ \text{calculus})+O(\text{CP})\\
		&=O(N \cdot n \cdot  t)+O(N \cdot n \cdot T)+O(1) \\
		&=O(N^2 \cdot n)+O(N \cdot n \cdot T) \\
		&=O(N^2 \cdot n + N \cdot n \cdot T) \\
		&=O(n \cdot T)
	\end{aligned}
\end{equation}
where $N$ is the finite population size;
$n$ represents the problem dimension;
$t = N$ is the number of iterations required to generate the FOLC sequence;
$T$ is the total number of iterations.

\section{Numerical Results on CEC2017} \label{sec4}

In this section, we conduct numerical experiments
to demonstrate the superior performance of the MFO-DBO algorithm on the CEC2017 benchmark suite.
All experiments were conducted in a PC (Inter Core i5-8257U, 1.4GHz, 8GB RAM) with MATLAB 2023b.

\subsection{Experimental setup} \label{subsec4.1}

For comparison, all competing algorithms, shown in~\autoref{parameter},
use an initial population size of 30 and are limited to a maximum of 500 iterations.
Each test function is executed 30 times independently to ensure statistical robustness, and the average results are reported.
The 16 selected algorithms are grouped into four categories:
\begin{itemize} \setlength{\itemsep}{0pt}
	\item DBO and its improved variants: DBO, IDBO, MDBO, QHDBO, GODBO, EDBO, MsDBO (introduced in Section~\ref{sec1});
	\item Competition-winning algorithms: LSHADE and EBOwithCMAR, which achieved top performance on the CEC2017 benchmark suite and remain among the most competitive baselines. Notably, several newer champion algorithms underperform them on this benchmark;
	\item Recently proposed metaheuristics: HEOA and HLOA, which share behavioral modeling principles with DBO-like designs; and IGWO and SCWOA, which are recent enhanced variants of classic algorithms such as GWO and WOA;
	\item Fractional-order based optimizers: FOPSO, FOSSA, and FOFPA, which integrate fractional-order calculus to improve convergence via memory-enhanced dynamics.
\end{itemize}

\begin{table}[H]
	\centering	
	\captionsetup{
		labelsep=newline,
		justification=raggedright,
		singlelinecheck=false,
		labelfont=bf,
		skip=0pt,
		margin=0pt
	}
	\caption{Parameter setting.}
	\label{parameter}
	\scriptsize
	\resizebox{\columnwidth}{!}{%
		\begin{tabular}{lllll}
			\toprule
			Algorithm & Reference &Population size & Number of iterations & Parameters \\
			\midrule
			DBO   &  \citep{Xue2023}   & 30 & 500 & $RDB = 6, BDB = 6, SDB = 7, TDB = 11$ \\
			IDBO   & \citep{wang2024prediction}   & 30 & 500 & $RDB = 6, BDB = 6, SDB = 7, TDB = 11$ \\
			MDBO  & \citep{shen2023multi}   & 30 & 500 & $RDB = 6, BDB = 6, SDB = 7, TDB = 11$ \\
			QHDBO  & \citep{ZHUfang2024}   & 30 & 500 & $RDB = 6, EFDBO = 13, TDB = 11$ \\
			GODBO   &  \citep{zilong2023multi}   & 30 & 500 & $RDB = 6, BDB = 6, SDB = 7, TDB = 11$ \\
			EDBO   &  \citep{li2024enhanced}   & 30 & 500 & $RDB = 6, BDB = 6, SDB = 7, TDB = 11$ \\
			MsDBO   &  \citep{chen2025coverage}   & 30 & 500 & $RDB = 6, BDB = 6, SDB = 7, TDB = 11$ \\
		    LSHADE   &  \citep{tanabe2014improving}   & 30 & 500 & $N^i=20*D, \widehat{N} = 4, p=0.11$ \\
		    EBOwithCMAR   &  \citep{kumar2017improving}   & 30 & 500 & $N_1^i = 18 \times D,  N_1^{\min} = 4, N_2^i = 46.8 \times D,\sigma = 0.3$ \\
		    HLOA  & \citep{Peraza2024}  & 30 & 500 & $V_0 = 1, g = 0.009807$ \\
		    HEOA  & \citep{LIAN2024122638} & 30 & 500 & $A = 0.6, LN = 0.4, FN = 0.1, EN = 0.4$ \\
		    IGWO  & \citep{wang2024modified} &30 & 500  & $\alpha=0.5, \beta=0.1, \gamma=0.5$\\
		    SCWOA & \citep{hou2025novel} & 30  &  500 & $a=2-t*(2/Max_iter), Limit =10$\\
			FOPSO   &  \citep{gao2014fractional}   & 30 & 500 & $c_1 =2, c_2 = 2, \alpha = 1/(2*exp(-0.47*f))$ \\
			FOSSA   & \citep{jiang2021fractional} & 30 & 500 & $ PD=0.7,SD=0.2, v = 1/(2*exp(-0.47*f))   $ \\
			FOFPA    & \citep{YOUSRI2020} & 30 & 500 & $a=0.6, b=0.9, \alpha=0.4, r=4$ \\
			MFO-DBO &  our paper   & 30 & 500 & $RDB = 6, BDB = 6, SDB = 7, TDB = 11$ \\
			\bottomrule
		\end{tabular}
	}
\end{table}

\subsection{Numerical simulation} \label{subsec4.2}

Next, we identify an appropriate fractional-order derivative $\delta$,
and then systematically assess the efficacy of the three proposed strategies from three aspects:
initialization quality, perturbation mechanism, and the integration of all enhancements described in Section~\ref{sec3}.
Finally, we analyze the statistical properties of MFO-DBO in comparison to 16 other algorithms using the CEC2017 test suite.

\subsubsection{Benchmark suite}\label{subsubsec4.2.1}

The CEC2017 benchmark suite at dimensions of 50 and 100 (50Dim and 100Dim) is adopted in our experiments.
This suite includes 29 single-objective test functions,
categorized as follows: unimodal (F1, F3), simple multimodal (F4–F10), hybrid (F11–F20), and composition (F21–F30) functions.
Notably, F2 is excluded due to uncontrollable experimental factors.
A summary is presented in~\autoref{cec2017_summary}.

\begin{table}[H]
	\centering
	\captionsetup{
		labelsep=newline,
		justification=raggedright,
		singlelinecheck=false,
		labelfont=bf,
		skip=0pt,
		margin=0pt
	}
	\caption{Summary of the CEC2017 benchmark functions used.}
	\label{cec2017_summary}
	\scriptsize
	\resizebox{\columnwidth}{!}{%
		\begin{tabular}{lll}
			\toprule
			Type	& NO. Range & Features \\
			\midrule
			Simple Unimodal & F1, F3 & Single global optimum; assesses exploitation. \\
			Simple Multimodal & F4–F10 & Numerous local optima; evaluates exploration. \\
			Hybrid & F11–F20 & Combination of various functions; complex. \\
			Composition & F21–F30 & Complex landscapes with multiple optima; challenging, real-world applicability. \\
			\bottomrule
		\end{tabular}
	}
\end{table}

As summarized in \autoref{cec2017_summary},
unimodal functions are designed to assess exploitation by containing a single global optimum,
whereas simple multimodal functions, with their numerous local optima,
present a greater challenge and are used to evaluate an algorithm's exploration capabilities.
Hybrid and composition functions, being more complex and challenging,
effectively reflect an algorithm's potential to address real-world optimization challenges.

\subsubsection{Determination of Fractional-Order Parameter} \label{subsubsec4.2.2}

The fractional-order parameter $\delta $, as defined in Eq.~\eqref{eq12},
notably influences the contribution of historical terms in the position update process.
By varying $\delta $, the balance between exploration and exploitation is adjusted,
thereby impacting convergence behavior and solution quality.
Specially, we test the values of $\delta $ from 0 to 1 in increments of 0.1.
Numerical results summarized in \autoref{delta} show that
the MFO-DBO algorithm consistently outperforms the standard DBO algorithm across both dimensions,
Moreover, the MFO-DBO algorithm with $\delta = 0.1$ achieves the best average ranking across all functions in the CEC2017 suite,
implying superior search performance and robustness across various problem dimensions.
Detailed results are provided in the supplementary material (Tables S1 and S2).

\begin{table}[H]
	\captionsetup{
		labelsep=newline,
		justification=raggedright,
		singlelinecheck=false,
		labelfont=bf,
		skip=0pt,
		margin=0pt 	
	}
	\caption{Ranking comparison of MFO-DBO with different $\delta$ values on CEC2017.}
	\label{delta}
	\scriptsize	
	\resizebox{\columnwidth}{!}{%
		\begin{tabular}{l@{\hskip 9pt}l@{\hskip 9pt}l@{\hskip 9pt}l@{\hskip 9pt}l@{\hskip 9pt}l@{\hskip 9pt}l@{\hskip 9pt}l@{\hskip 9pt}l@{\hskip 9pt}l@{\hskip 9pt}l@{\hskip 9pt}l@{\hskip 9pt}l}
			\toprule
			&  &\multirow{2}{*}{DBO} & \multicolumn{10}{c}{MFO-DBO} \\
			\cmidrule{4-13}
			& & & $\delta=0.1$ & $\delta=0.2$ & $\delta=0.3$ & $\delta=0.4$ & $\delta=0.5$ & $\delta=0.6$ & $\delta=0.7$ & $\delta=0.8$ & $\delta=0.9$ & $\delta=1.0$ \\
			\midrule			
			50Dim & Rank &9.17 	& \textbf{2.34} & 3.90 &	3.69 &	5.10 &	5.83 &	6.38 &	7.41 &	6.93 &	7.66 &	7.41\\	
			100Dim & Rank & 9.28  & \textbf{2.52}  & 3.38  & 4.72	 & 4.38  & 6.03  & 6.48 	& 6.72  & 6.93  & 7.69  & 7.86 \\
			\bottomrule
		\end{tabular}
	}
\end{table}

\begin{figure}[H]
	\centering
	\begin{minipage}{0.48\textwidth}
		\centering
		\includegraphics[width=\textwidth, height=4.5cm, keepaspectratio]{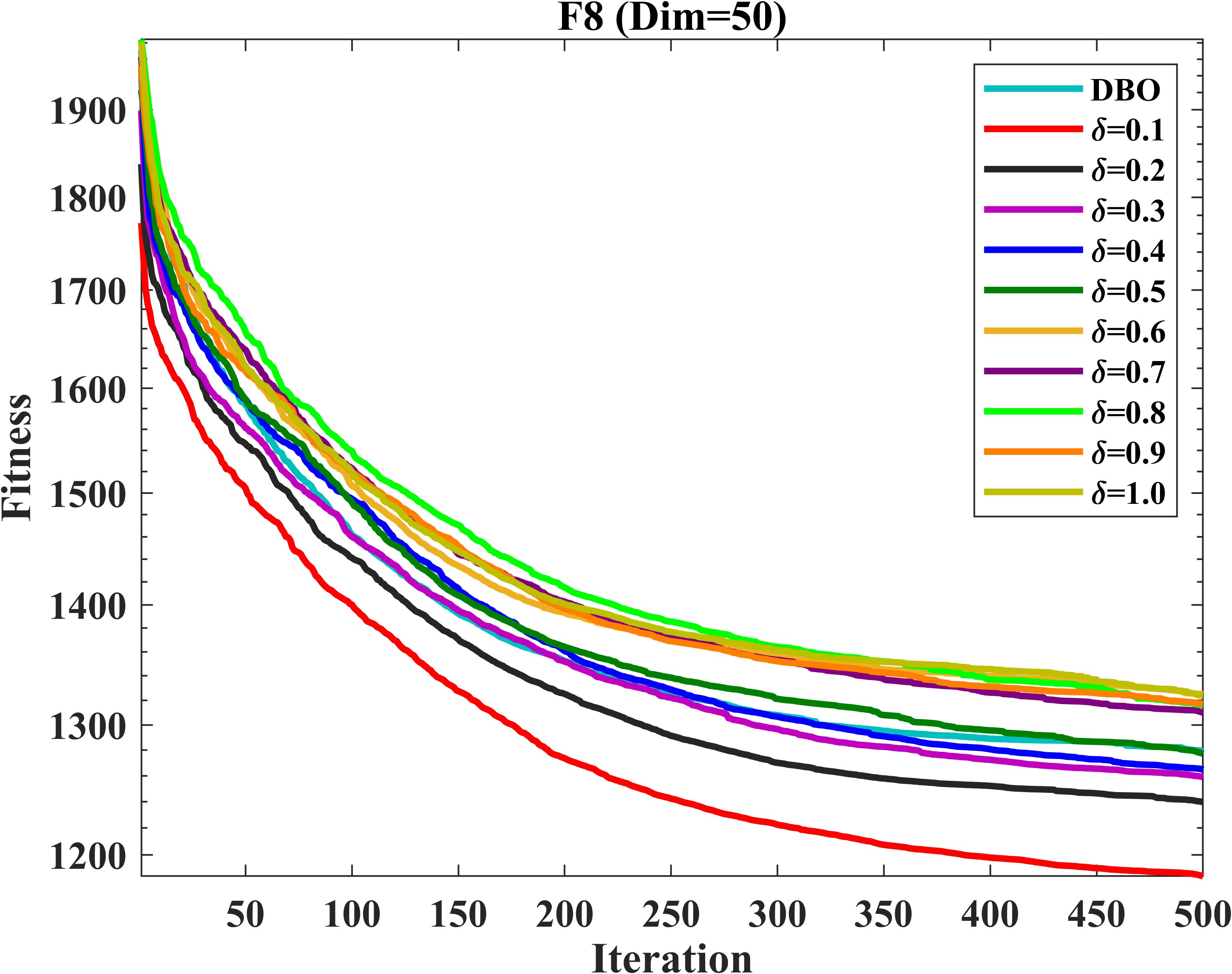}
		\subcaption{F8 (50Dim)}
	\end{minipage}
	\hfill
	\begin{minipage}{0.48\textwidth}
		\centering
		\includegraphics[width=\textwidth, height=4.5cm, keepaspectratio]{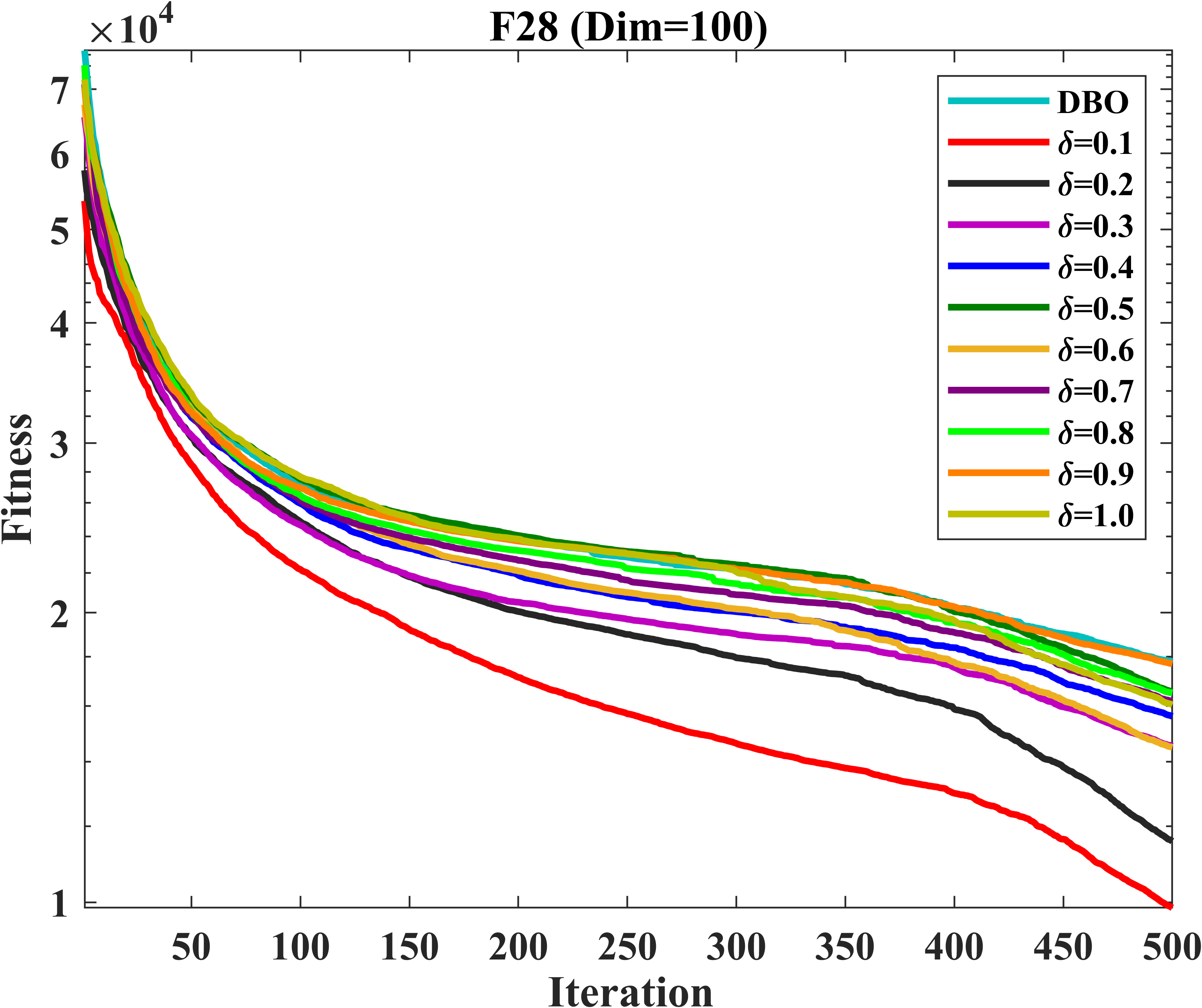}
		\subcaption{F28 (100Dim)}
	\end{minipage}
	\caption{Performance of MFO-DBO under different $\delta$ values on CEC2017.}
	\label{DETA}
\end{figure}

\autoref{DETA} presents the convergence curves of MFO-DBO under different $\delta$ values.
Two representative functions are selected for illustration: F8 (50Dim) and F28 (100Dim).
The plots clearly indicate that $\delta = 0.1$ achieves the best convergence in both cases,
corroborating the earlier numerical results.
These results, derived from both quantitative rankings and convergence trends,
confirm that $\delta = 0.1$ is the optimal parameter setting for MFO-DBO.

\subsubsection{Effectiveness Analysis of Improvement Strategies} \label{subsubsec4.2.3}

To demonstrate the effectiveness of the proposed enhancement strategies for the DBO algorithm,
we define three variants: DBO1 (DBO + FOLC), DBO2 (DBO + FO calculus), and DBO3 (DBO + CP).
The comprehensive algorithm that integrates all three enhancements is denoted as MFO-DBO.
We conduct systematic experiments to evaluate the impact of each improvement in terms of initialization methods,
different perturbation mechanisms, and the independent and synergistic effects of our core improvement strategies.
In the following subsections,
$B|S|W$ statistics indicate whether the algorithm in the row performs Better (B), Similar (S),
or Worse (W) than the one in the column, based on mean and standard deviation.

\textbf{Note:} Only summary-level comparisons are presented;
detailed results for all benchmark functions are provided in the supplementary material (Tables S3-S6).

\subsubsection*{a) Effectiveness of Initialization Strategies}

This section evaluates the impact of different initialization strategies on the DBO algorithm.
We specifically validate the effectiveness of fractional-order Logistic chaotic map (FOLC) initialization
and classic Logistic chaotic map initialization.
The algorithms involved in this comparison are the standard DBO, DBO1 (DBO + FOLC), and DBOC (DBO + Chaotic map).

\begin{table}[H]
	\centering
	\captionsetup{
		labelsep=newline,
		justification=raggedright,
		singlelinecheck=false,
		labelfont=bf,
		skip=0pt,
		margin=3pt
	}
	\caption{Mean and Standard Deviation: DBO vs. Initialization Strategies on CEC2017.}
	\label{vschaos}
	\scriptsize
	\begin{tabular}{lp{4cm}p{2cm}p{2cm}p{2cm}}
		\toprule
		\multirow{2}{*}{Dimension} &  & \multicolumn{3}{l}{Algorithms} \\
		\cmidrule(r){3-5}
		& & DBO & DBOC & DBO1 \\
		\midrule
		\multirow{2}{*}{50Dim}
		& $B|S|W$ (DBO vs. others) & NA & $20|3|6$ & $4|2|23$  \\
		& $B|S|W$ (DBO1 vs. others) & $23|2|4$ & $25|0|4$  & NA \\
		\midrule
		\multirow{2}{*}{100Dim}
		& $B|S|W$ (DBO vs. others) & NA & $17|1|11$ & $8|0|21$  \\
		& $B|S|W$ (DBO1 vs. others) & $21|0|8$ & $20|1|8$ & NA \\
		\bottomrule	
	\end{tabular}	
\end{table}

\begin{figure}[H]
	\centering
	\includegraphics[width=0.65\textwidth]{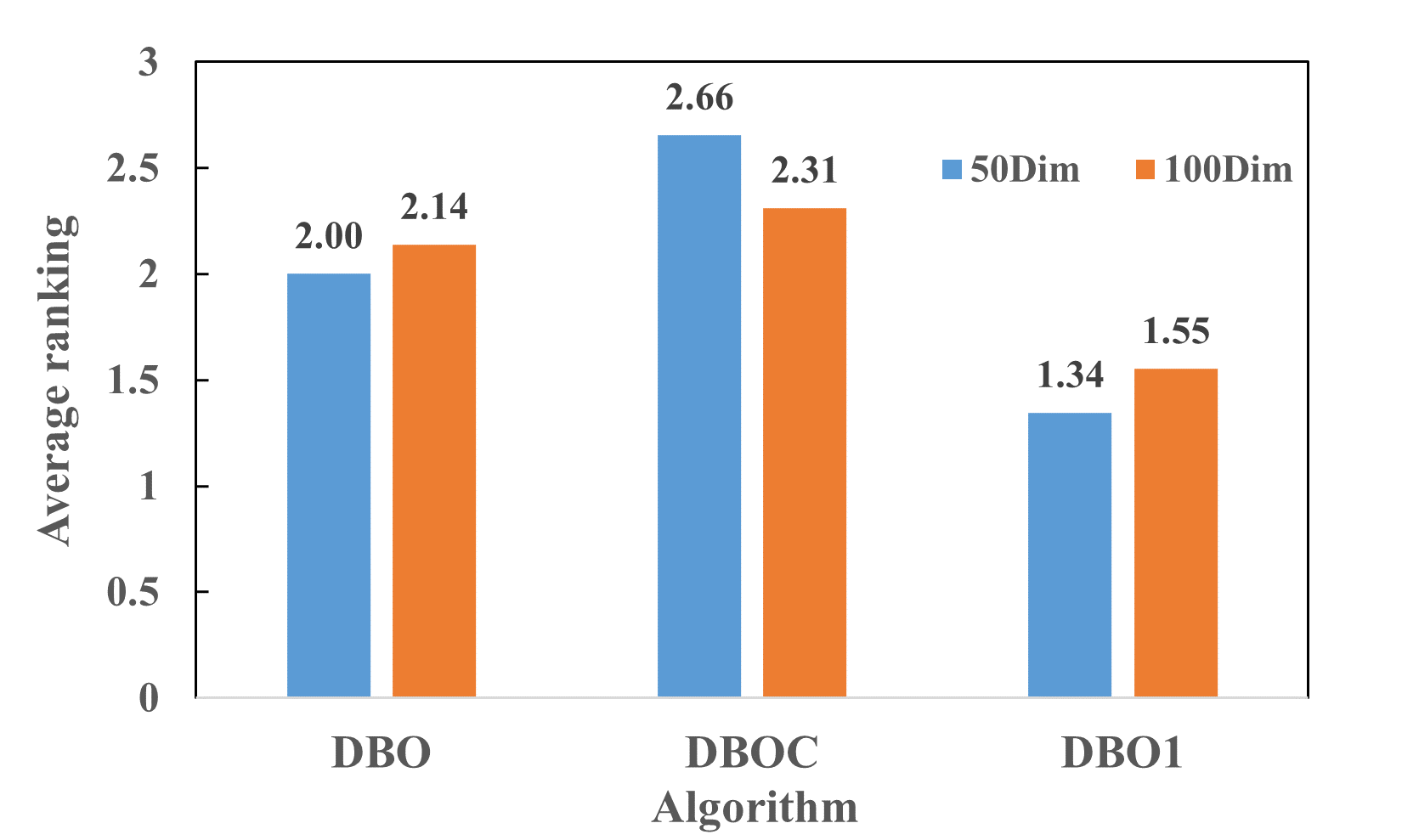}
	\caption{Average rankings of DBO vs. Initialization Strategies on CEC2017.}
	\label{VS1}
\end{figure}

As shown in \autoref{vschaos} and \autoref{VS1}, DBO1 outperforms the standard DBO on 23 (50Dim) and 21 (100Dim) functions,
and achieves the best average rank of 1.34 and 1.55, respectively.
In contrast, DBOC only surpasses DBO on 6 and 11 functions, and ranks lowest among the three.
These results confirm that FOLC-based initialization substantially enhances global search capability,
while classical chaotic mapping offers limited or even negative effect.

\subsubsection*{b) Effectiveness of Perturbation Mechanisms}

This section evaluates the impact of different perturbation mechanisms on the DBO algorithm.
We specifically validate the effectiveness of adaptive $t$-distribution perturbation,
adaptive Gaussian-Cauchy perturbation, and our proposed chaotic perturbation.
The algorithms involved in this comparison are the standard DBO, DBOAT (DBO + adaptive $t$-distribution perturbation),
DBOGC (DBO + adaptive Gaussian-Cauchy perturbation), and DBO3 (DBO + chaotic perturbation).

\begin{table}[H]
	\centering
	\captionsetup{
		labelsep=newline,
		justification=raggedright,
		singlelinecheck=false,
		labelfont=bf,
		skip=0pt,
		margin=0pt
	}
	\caption{Mean and Standard Deviation: DBO vs. Perturbation Strategies on CEC2017.}
	\label{RD}
	\scriptsize
	\begin{tabular}{lp{3.6cm}p{1.8cm}p{1.8cm}p{1.8cm}l}
		\toprule
		\multirow{2}{*}{Dimension} &  & \multicolumn{4}{l}{Algorithms} \\
		\cmidrule(r){3-6}
		& & DBO & DBOAT & DBOGC & DBO3\\
		\midrule
		\multirow{2}{*}{50Dim}
		& $B|S|W$ (DBO vs. others) & NA & $14|4|11$ & $17|2|10$ & $2|1|26$\\
		& $B|S|W$ (DBO3 vs. others) & $26|1|2$ & $26|0|3$ & $26|0|3$ & NA \\
		\midrule
		\multirow{2}{*}{100Dim}
		& $B|S|W$ (DBO vs. others) & NA & $20|0|9$ & $18|2|9$ & $2|1|26$ \\
		& $B|S|W$ (DBO3 vs. others) & $26|1|2$ & $26|1|2$ & $27|1|1$ & NA \\
		\bottomrule	
	\end{tabular}	
\end{table}

\begin{figure}[htp]
	\centering
	\includegraphics[width=0.65\textwidth]{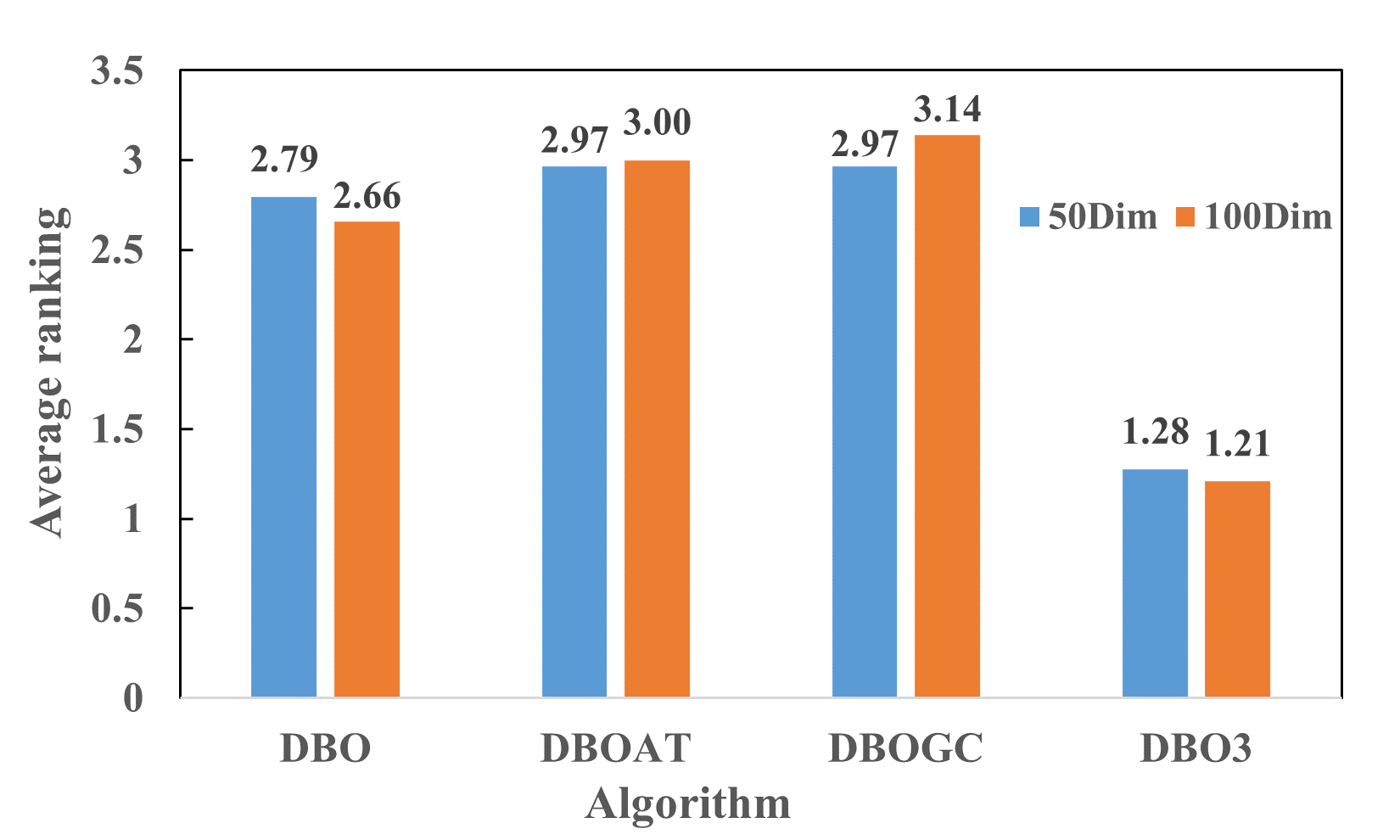}
	\caption{Average rankings of DBO vs. Perturbation Strategies on CEC2017.}
	\label{VS2}
\end{figure}

According to \autoref{RD} and \autoref{VS2}, DBO3 outperforms the standard DBO on 26 functions in both 50Dim and 100Dim cases,
and achieves the best average ranks of 1.28 and 1.21, respectively.
In contrast, DBOAT and DBOGC perform worse than DBO on 14 and 17 functions in 50Dim,
and their performance further deteriorates in the 100Dim case.
These results highlight the superiority of the chaotic perturbation mechanism and the ineffectiveness of the adaptive $t$-distribution
and Gaussian–Cauchy strategies in high-dimensional scenarios.

\subsubsection*{c) Effectiveness of Proposed Strategies}

This section provides a detailed evaluation of the effectiveness of the three core improvement strategies
by examining their individual contributions as well as their combined impact.
Specifically, we compare the standard DBO with three enhanced variants: DBO1 (DBO + FOLC map), DBO2 (DBO + FO calculus), and DBO3 (DBO + CP),
in addition to the comprehensive version, MFO-DBO, which integrates all three enhancements.

\begin{table}[H]
	\centering
	\captionsetup{
		labelsep=newline,
		justification=raggedright,
		singlelinecheck=false,
		labelfont=bf,
		skip=0pt,
		margin=0pt
	}
	\caption{Mean and standard deviation: DBO vs. Proposed Improvement Strategies on CEC2017.}
	\label{three enhancements}
	\scriptsize
	\resizebox{\columnwidth}{!}{%
		\begin{tabular}{ll lllll}
			\toprule
			\multirow{2}{*}{Dimension} &  & \multicolumn{5}{l}{Algorithms} \\
			\cmidrule(r){3-7}
			& & DBO & DBO1 & DBO2 & DBO3 & MFO-DBO \\
			\midrule
			\multirow{2}{*}{50Dim}
			& $B|S|W$ (DBO vs. others) & NA & $4|2|23$ & $2|1|26$ & $2|1|26$ & $0|1|28$ \\
			& $B|S|W$ (MFO-DBO vs. others) & $28|1|0$ & $28|0|1$ & $27|2|0$ & $23|0|6$ & NA \\
			\midrule
			\multirow{2}{*}{100Dim}
			& $B|S|W$ (DBO vs. others) & NA & $8|0|21$ & $3|0|26$ & $2|1|26$ & $0|0|29$ \\
			& $B|S|W$ (MFO-DBO vs. others) & $29|0|0$ & $28|0|1$ & $28|1|0$ & $25|0|4$ & NA \\
			\bottomrule	
		\end{tabular}	
	}
\end{table}

\begin{figure}[htp]
	\centering
	\includegraphics[width=0.65\textwidth]{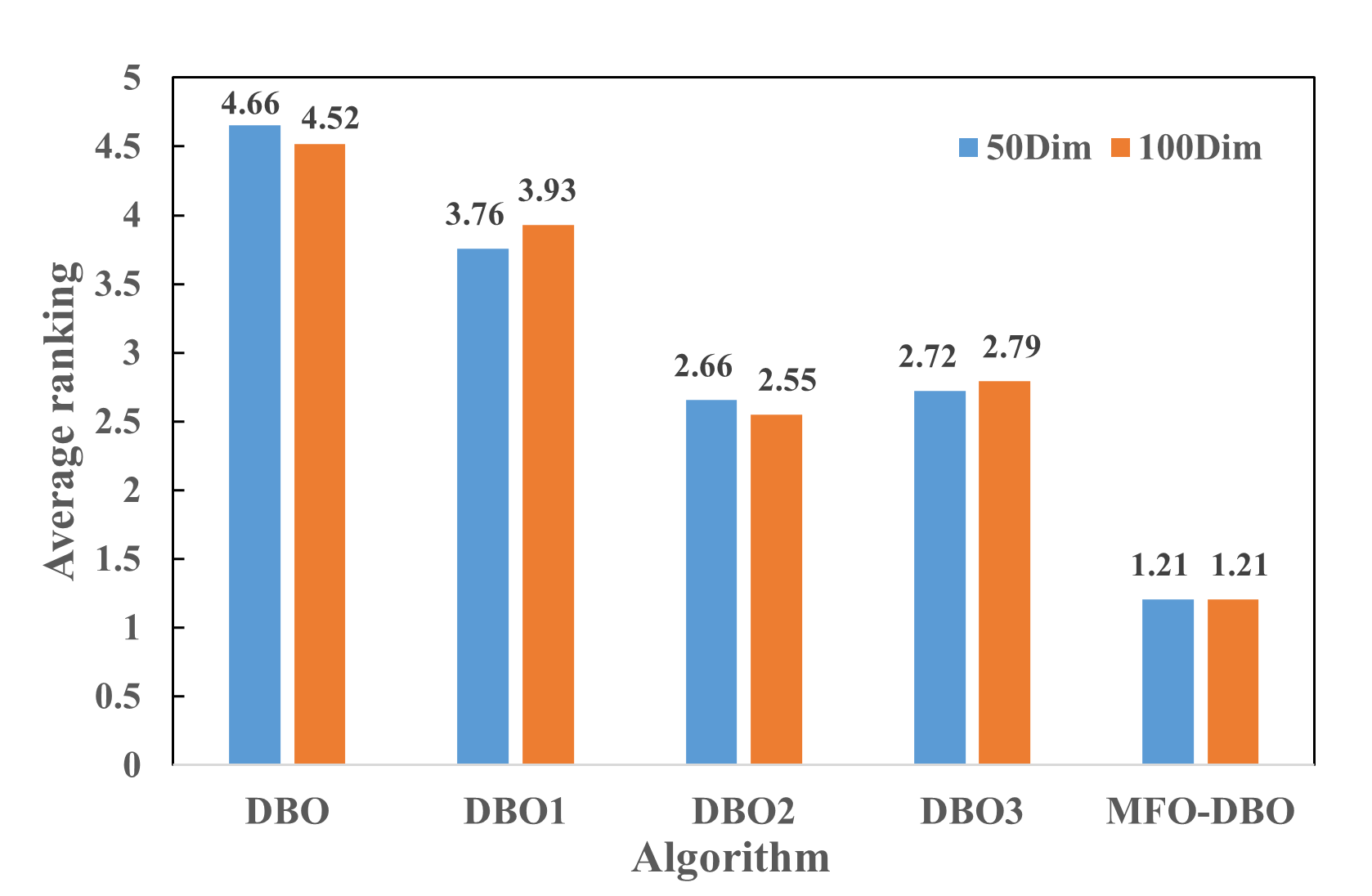}
	\caption{Average rankings of DBO vs. proposed improvement strategies on CEC2017.}
	\label{VS3}
\end{figure}

As illustrated in \autoref{three enhancements} and \autoref{VS3},
all enhanced variants—DBO1, DBO2, and DBO3—significantly outperform the baseline DBO, with wins on 23, 26, and 26 functions, respectively, in the 50Dim case.
MFO-DBO further improves performance by outperforming DBO1, DBO2, and DBO3 on 28, 27, and 23 functions, respectively.
It consistently achieves the best average rank of 1.21 in both 50Dim and 100Dim, confirming the effectiveness of combining initialization, memory, and perturbation strategies into a unified optimization framework.

\subsubsection{Comparisons with Several DBO Variants} \label{subsubsec4.2.4}

This assessment examines the stability, convergence behavior,
and overall effectiveness of our proposed MFO-DBO algorithm compared with six other improved DBO variants on CEC2017 test suite.
The competing algorithms include IDBO, MDBO, QHDBO, GODBO, EDBO, and MsDBO.
The parameter settings used in the experiments are detailed in~\autoref{parameter}.
The evaluation involves five aspects of statistical analysis:
\begin{itemize}[labelsep=0.5em, leftmargin=3em, itemsep=0pt, topsep=0pt, parsep=0pt, partopsep=0pt]
	\item[a)] Mean and standard deviation analysis, presented in~\autoref{VS_VDBO} and~\autoref{VS4};
	\item[b)] Wilcoxon rank-sum test, reported in~\autoref{wilcoxon_VSDBO};
	\item[c)] Friedman test rankings, shown in~\autoref{friedman_VSDBO};
	\item[d)] Convergence curve analysis, shown in~\autoref{CO_VDBO};
	\item[e)] Stability analysis, shown in~\autoref{SA_VDBO}.
\end{itemize}

\noindent
This section presents summary-level results of MFO-DBO compared with DBO variants,
including performance comparisons based on mean and standard deviation, Wilcoxon $p$-values, and Friedman ranks.
Full numerical results for all benchmark functions are provided in the supplementary material (Tables S7–S10).

\subsubsection*{a) Mean and Standard Deviation Analysis}
\addcontentsline{toc}{subsubsection}{A. Mean and Standard Deviation Analysis}
\label{avg_var}

\begin{figure}[htp]
	\centering
	\includegraphics[width=0.62\textwidth]{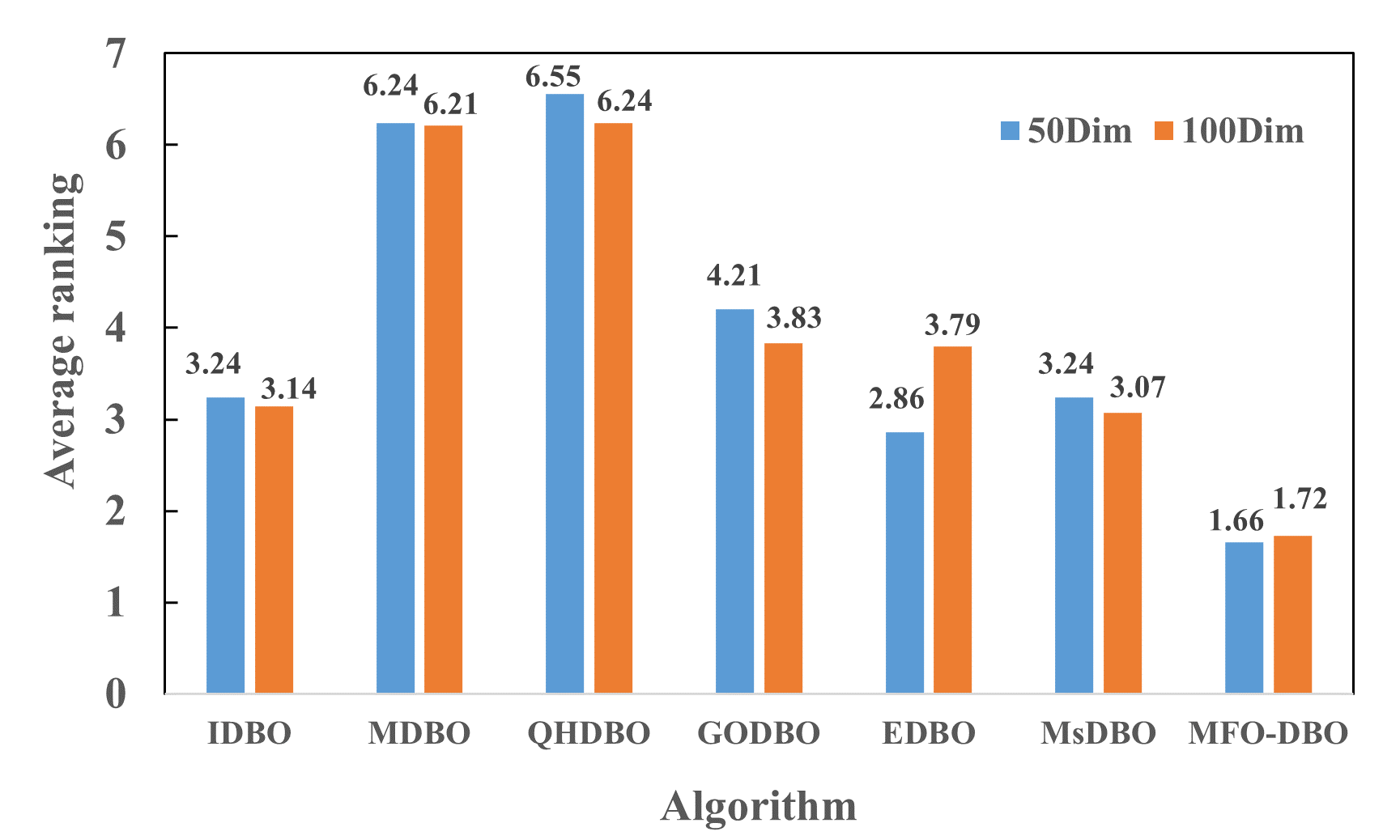}
	\caption{Average rankings of MFO-DBO vs. DBO Variants on CEC2017.}
	\label{VS4}
\end{figure}

As shown in~\autoref{VS_VDBO}, MFO-DBO achieves dominant performance over all six DBO variants.
In the 50Dim case, it outperforms IDBO, MDBO, QHDBO, GODBO, EDBO, and MsDBO on 23, 29, 29, 29, 21, and 26 functions, respectively.
When aggregated across both dimensions,
it achieves win ratios of 75.9\%, 98.3\%, 98.3\%, 96.6\%, 77.6\%, and 87.9\% of all functions,
highlight MFO-DBO’s strong advantages in convergence accuracy and robustness.
This superiority is further confirmed by the average ranking bar chart in~\autoref{VS4},
where MFO-DBO consistently ranks first across both 50Dim and 100Dim, clearly surpassing the other variants.

\begin{table}[H]
	\centering
	\captionsetup{
		labelsep=newline,
		justification=raggedright,
		singlelinecheck=false,
		labelfont=bf,
		skip=0pt,
		margin=0pt
	}
	\caption{Mean and Standard Deviation: MFO-DBO vs. DBO Variants on CEC2017.}
	\label{VS_VDBO}
	\scriptsize
	\resizebox{\columnwidth}{!}{%
		\begin{tabular}{lllllllll}
			\toprule
			\multirow{2}{*}{Dimension} &  & \multicolumn{6}{l}{Algorithms} & \multirow{2}{*}{Overall}\\
			\cmidrule(r){3-8}
			& & IDBO & MDBO & QHDBO & GODBO & EDBO & MsDBO & \\
			\midrule
			50Dim & $B|S|W$  &$23|0|6$ & $29|0|0$ &$ 29|0|0$ & $29|0|0$ &$ 21|0|8$ & $26|1|2$ & $157|1|16$\\
			100Dim	& $B|S|W$ & $21|0|8$ & $28|0|1$ & $28|0|1$ & $27|0|2$ & $24|0|5$ & $25|0|4$ & $153|0|21$ \\
			\midrule
			Overall  &  $B|S|W$  & $44|0|14$   & $57|0|1$  & $57|0|1$ & $56|0|2$  & $45|0|13 $ &  $51|1|6$  & $310|1|37$ \\
			\bottomrule	
		\end{tabular}	
	}
\end{table}

\subsubsection*{b) Wilcoxon Rank-Sum Test}
\addcontentsline{toc}{subsubsection}{B. Wilcoxon Rank-Sum Test}
\label{wilcoxon}

The Wilcoxon rank-sum test is adopted to assess statistical significance between paired algorithms,
where the symbols $+$, $\approx$, and $-$ represent significantly better, similar, or worse performance, respectively.
The significance level is set at 0.05 (i.e., $p < 0.05$).
As shown in~\autoref{wilcoxon_VSDBO},
MFO-DBO significantly outperforms IDBO, MDBO, QHDBO, GODBO, EDBO, and MsDBO in 21, 29, 29, 27, 19, and 20 functions, respectively.
Overall, MFO-DBO achieves statistically significant improvements in 283 out of 348 comparisons (81.3\%), and is rarely inferior, with only 26 losses. These results further confirm the robustness and consistent superiority of MFO-DBO over the other DBO variants.

\begin{table}[H]
	\centering
	\captionsetup{
		labelsep=newline,
		justification=raggedright,
		singlelinecheck=false,
		labelfont=bf,
		skip=0pt,
		margin=0pt
	}
	\caption{Wilcoxon rank sum comparisons: MFO-DBO vs. DBO Variants on CEC2017.}
	\label{wilcoxon_VSDBO}
	\scriptsize
	\resizebox{\columnwidth}{!}{%
		\begin{tabular}{lllllllll}
			\toprule
			\multirow{2}{*}{Dimension} &  & \multicolumn{6}{l}{Algorithms} & \multirow{2}{*}{Overall}\\
			\cmidrule(r){3-8}
			& & IDBO & MDBO & QHDBO & GODBO & EDBO & MsDBO &\\
			\midrule
			50Dim & $+/\approx/-$  &21/4/4 & 29/0/0 & 29/0/0 & 27/2/0 & 19/4/6 & 20/9/0 & 145/19/10\\
			100Dim & $+/\approx/-$ & 20/2/7 & 27/1/1 & 28/0/1 & 24/4/1 & 19/6/4 & 20/7/2  & 138/20/16 \\
			\midrule
			Overall  &  $+/\approx/-$  & 41/6/11   & 56/1/1  & 57/0/1 & 51/6/1   & 38/10/10 &  40/16/2  & 283/39/26   \\
			\bottomrule	
		\end{tabular}%	
	}
\end{table}

\subsubsection*{c) Friedman test rankings}
\addcontentsline{toc}{subsubsection}{C. Friedman test rankings}
\label{Cfriedman}

The Friedman test is used to statistically rank algorithms based on performance across all benchmark functions.
As shown in \autoref{friedman_VSDBO}, MFO-DBO consistently achieves the lowest average rank in both 50Dim and 100Dim settings,
with scores of 2.00 and 1.86, respectively.
The result confirms MFO-DBO’s superior overall performance relative to all six DBO variants.

\begin{table}[H]
	\centering
	\captionsetup{
		labelsep=newline,
		justification=raggedright,
		singlelinecheck=false,
		labelfont=bf,
		skip=0pt,
		margin=0pt
	}
	\caption{Friedman test rankings: MFO-DBO vs. DBO Variants on CEC2017.}
	\label{friedman_VSDBO}
	\scriptsize
	\renewcommand{\arraystretch}{1.2}
	\resizebox{\columnwidth}{!}{%
		\begin{tabular}{lllllllll}
			\toprule
			\multicolumn{1}{l}{Dimension} & \multicolumn{8}{l}{Algorithm and Frideman} \\
			\midrule
			\multirow{2}{*}{50Dim} & Algorithm & MFO-DBO & EDBO & IDBO & MsDBO & GODBO   & MDBO & QHDBO    \\
			& Friedman  & 2.00 & 2.62  & 3.17 & 3.45  & 3.97  & 6.31 & 6.48   \\
			\multirow{2}{*}{100Dim} & Algorithm & MFO-DBO & IDBO & MsDBO & EDBO  & GODBO & MDBO &  QHDBO    \\
			& Friedman   & 1.86  & 3.21 & 3.21  & 3.52  & 3.83 & 6.17 & 6.21         \\
			\bottomrule
		\end{tabular}
	}
\end{table}

\subsubsection*{d) Convergence analysis}
\addcontentsline{toc}{subsubsection}{D. Convergence analysis}
\label{convergence}

\begin{figure}[H]
	\centering
	\begin{minipage}{0.48\textwidth}
		\centering
		\includegraphics[width=\textwidth, height=4.5cm, keepaspectratio]{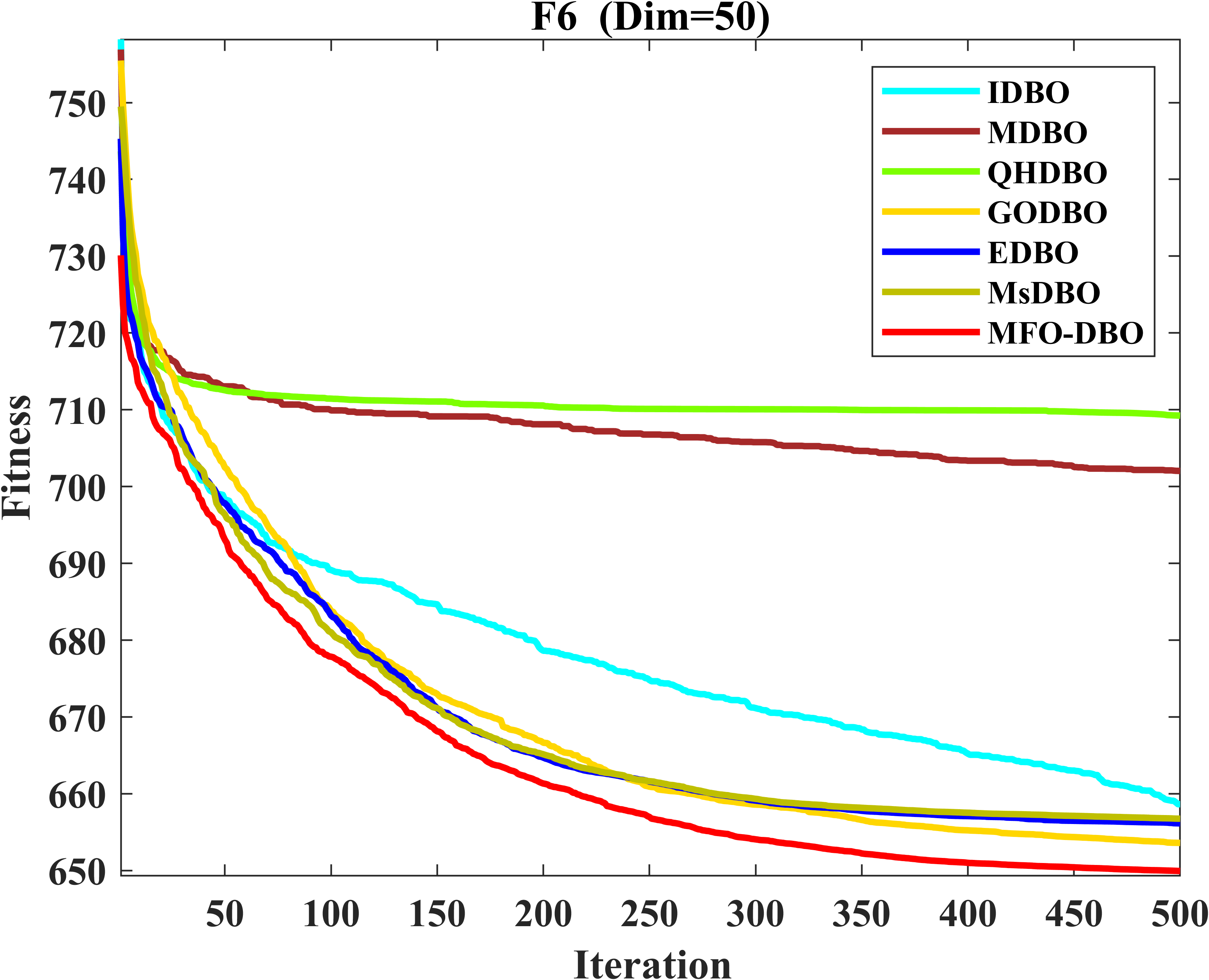}
		\subcaption{F6}
	\end{minipage}
	\hfill
	\begin{minipage}{0.48\textwidth}
		\centering
		\includegraphics[width=\textwidth, height=4.5cm, keepaspectratio]{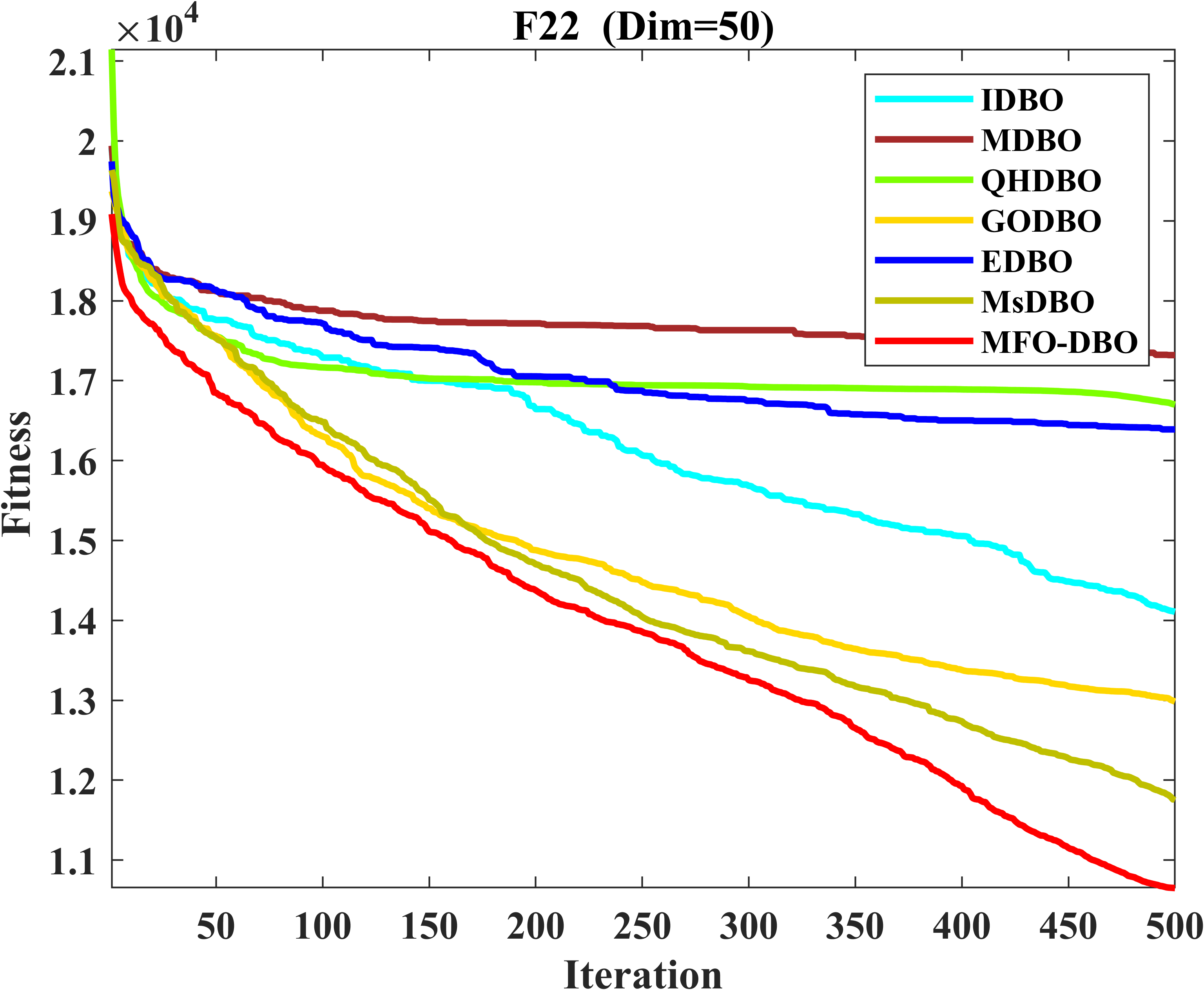}
		\subcaption{F22}
	\end{minipage}
	
	\vspace{0.5cm}
	
	\begin{minipage}{0.48\textwidth}
		\centering
		\includegraphics[width=\textwidth, height=4.5cm, keepaspectratio]{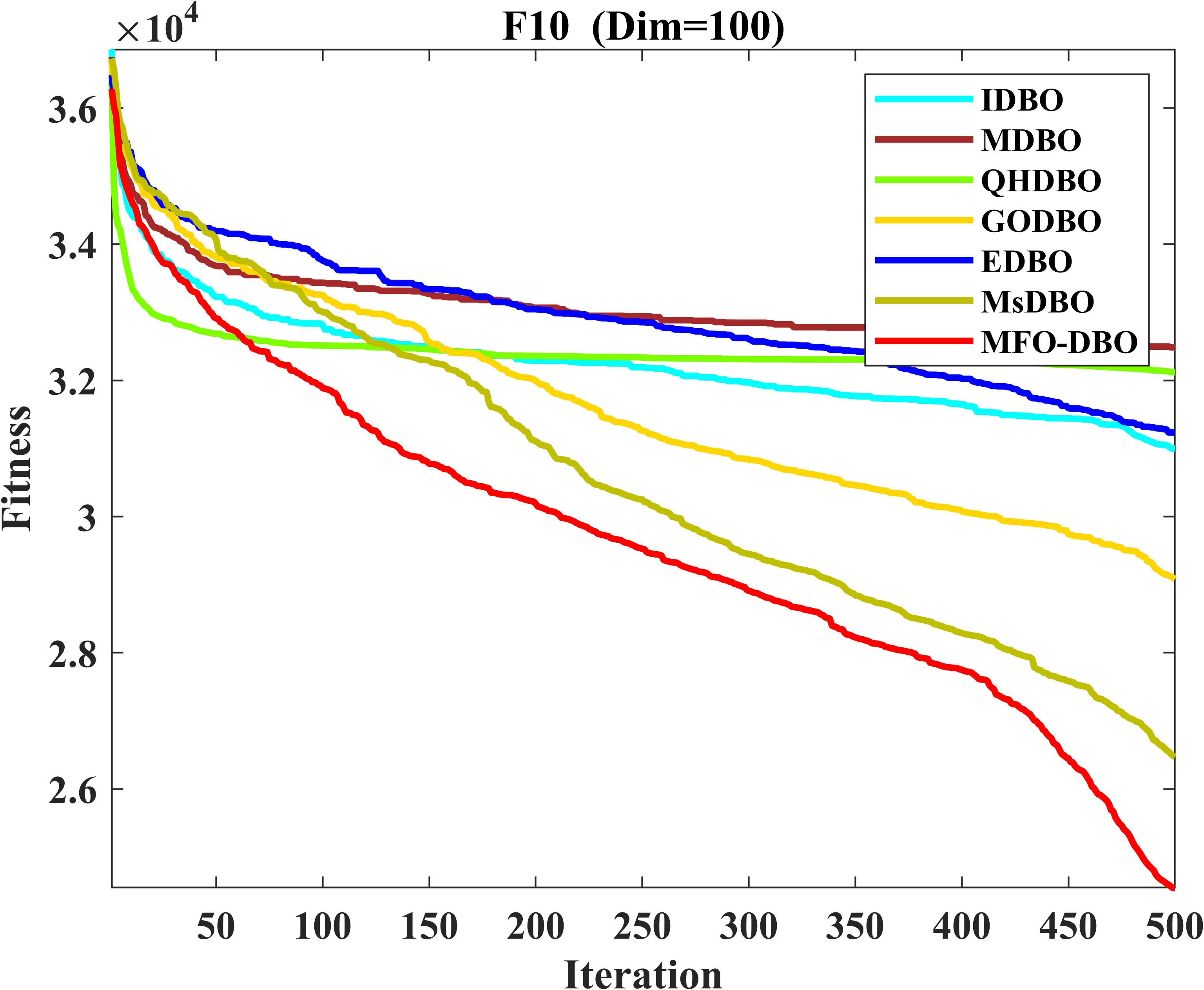}
		\subcaption{F10}
	\end{minipage}
	\hfill
	\begin{minipage}{0.48\textwidth}
		\centering
		\includegraphics[width=\textwidth, height=4.5cm, keepaspectratio]{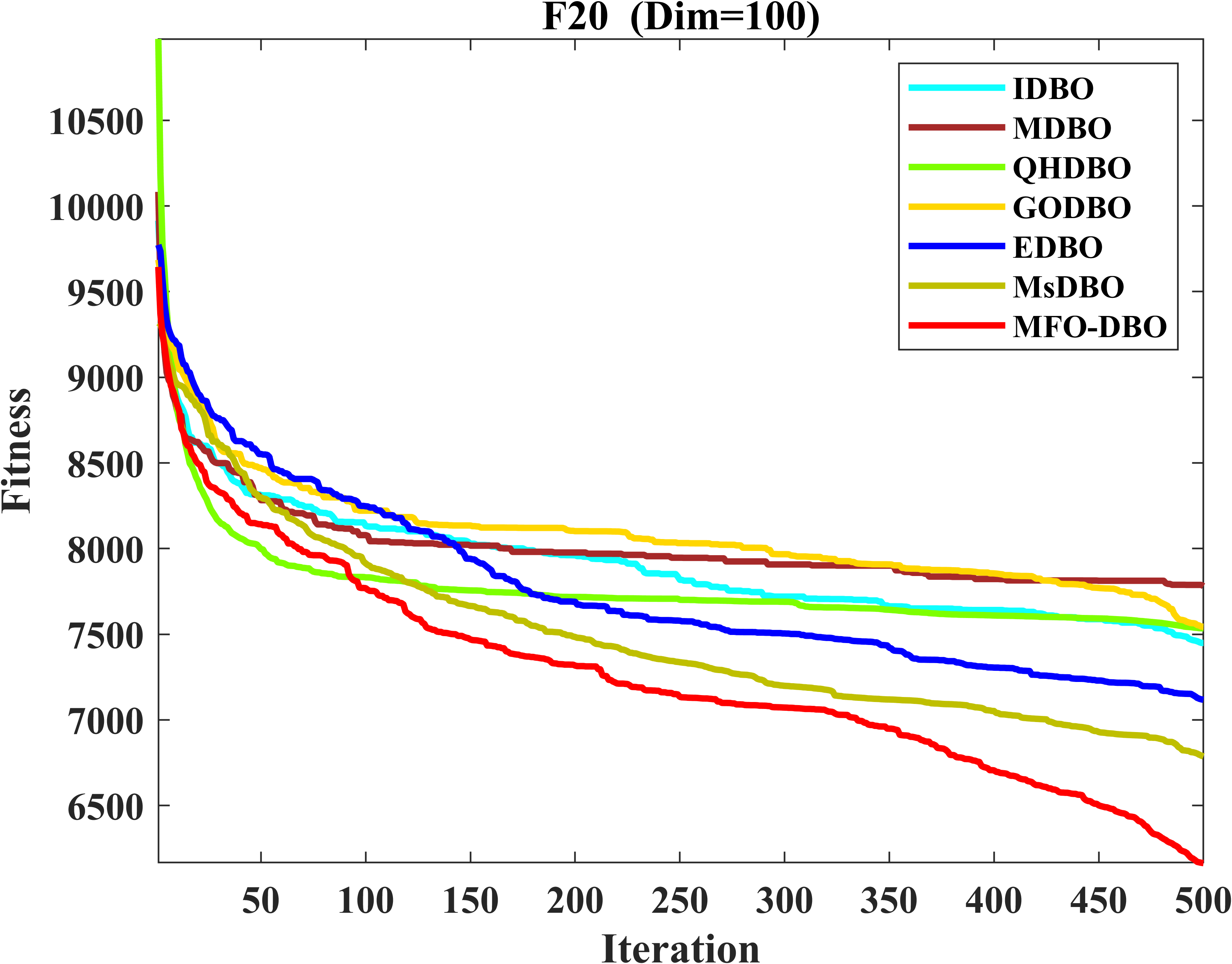}
		\subcaption{F20}
	\end{minipage}
	\caption{Convergence curves: MFO-DBO vs. DBO Variants on CEC2017.}
	\label{CO_VDBO}
\end{figure}

\autoref{CO_VDBO} shows the convergence behavior of MFO-DBO versus six DBO variants on selected representative functions: F6 and F22 (50Dim), and F10 and F20 (100Dim). In these cases, MFO-DBO consistently converges faster and achieves lower objective values than its competitors. Notably, on hybrid functions like F20 and composite functions like F22, MFO-DBO demonstrates significant advantages in convergence and final solution quality. These results confirm its superior search efficiency and convergence stability in high-dimensional scenarios.

\subsubsection*{e) Stability analysis}
\addcontentsline{toc}{subsubsection}{E. Stability analysis}
\label{Stability}

\begin{figure}[H]
	\centering
	\begin{minipage}{0.48\textwidth}
		\centering
		\includegraphics[width=\textwidth, height=4.5cm, keepaspectratio]{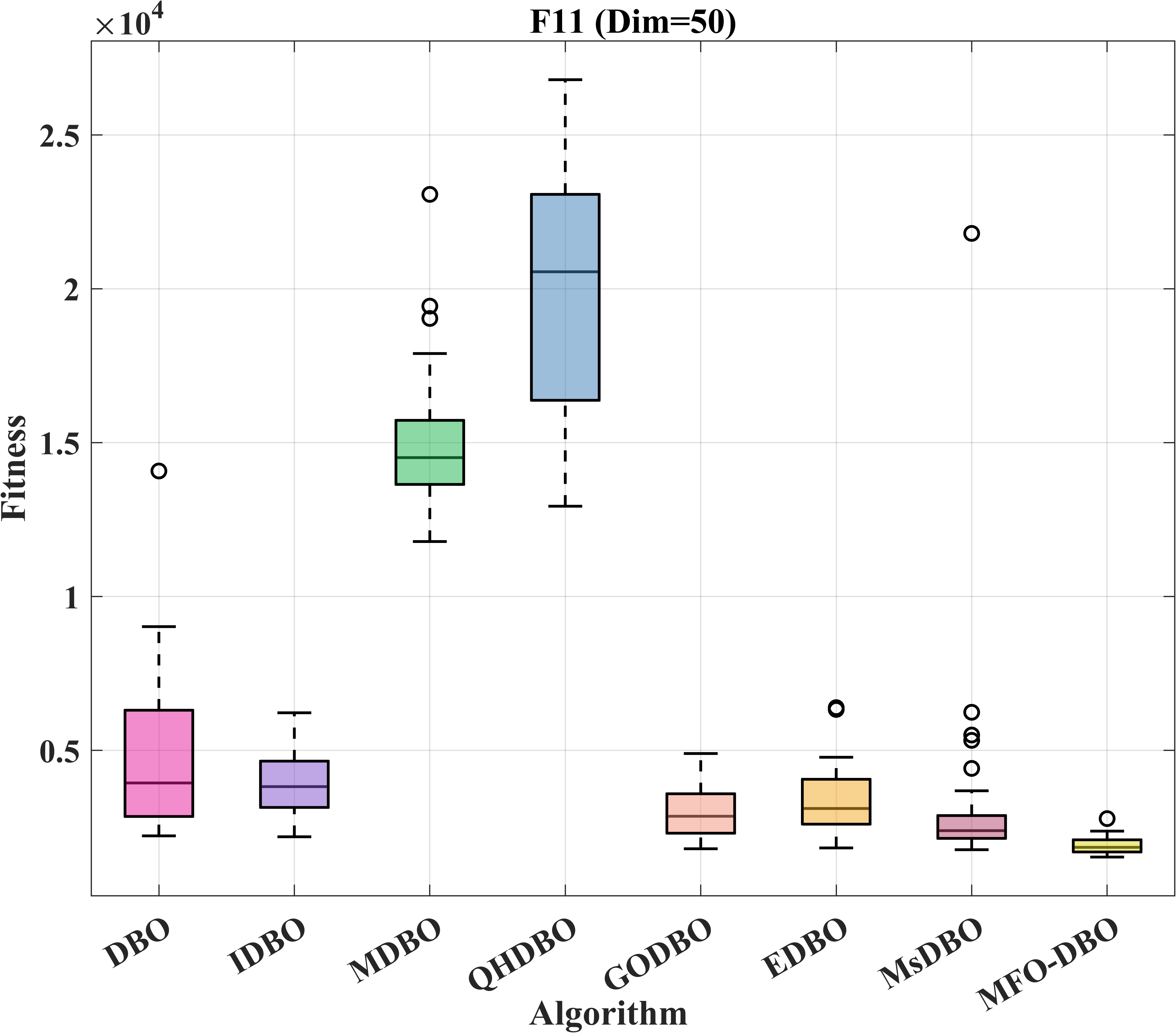}
		\subcaption{F11}
	\end{minipage}
	\hfill
	\begin{minipage}{0.48\textwidth}
		\centering
		\includegraphics[width=\textwidth, height=4.5cm, keepaspectratio]{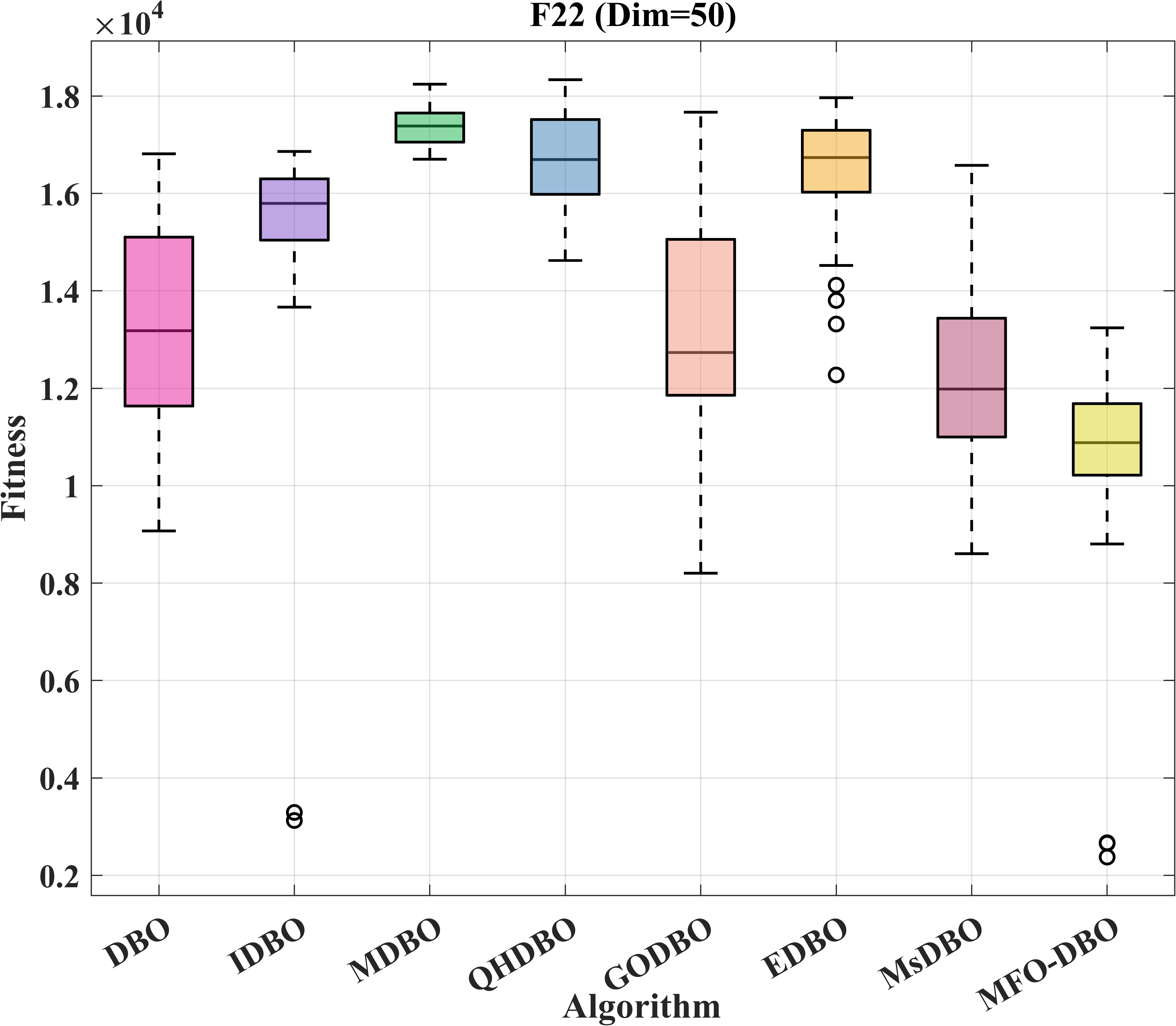}
		\subcaption{F22}
	\end{minipage}
	
	\vspace{0.5cm}
	
	\begin{minipage}{0.48\textwidth}
		\centering
		\includegraphics[width=\textwidth, height=4.5cm, keepaspectratio]{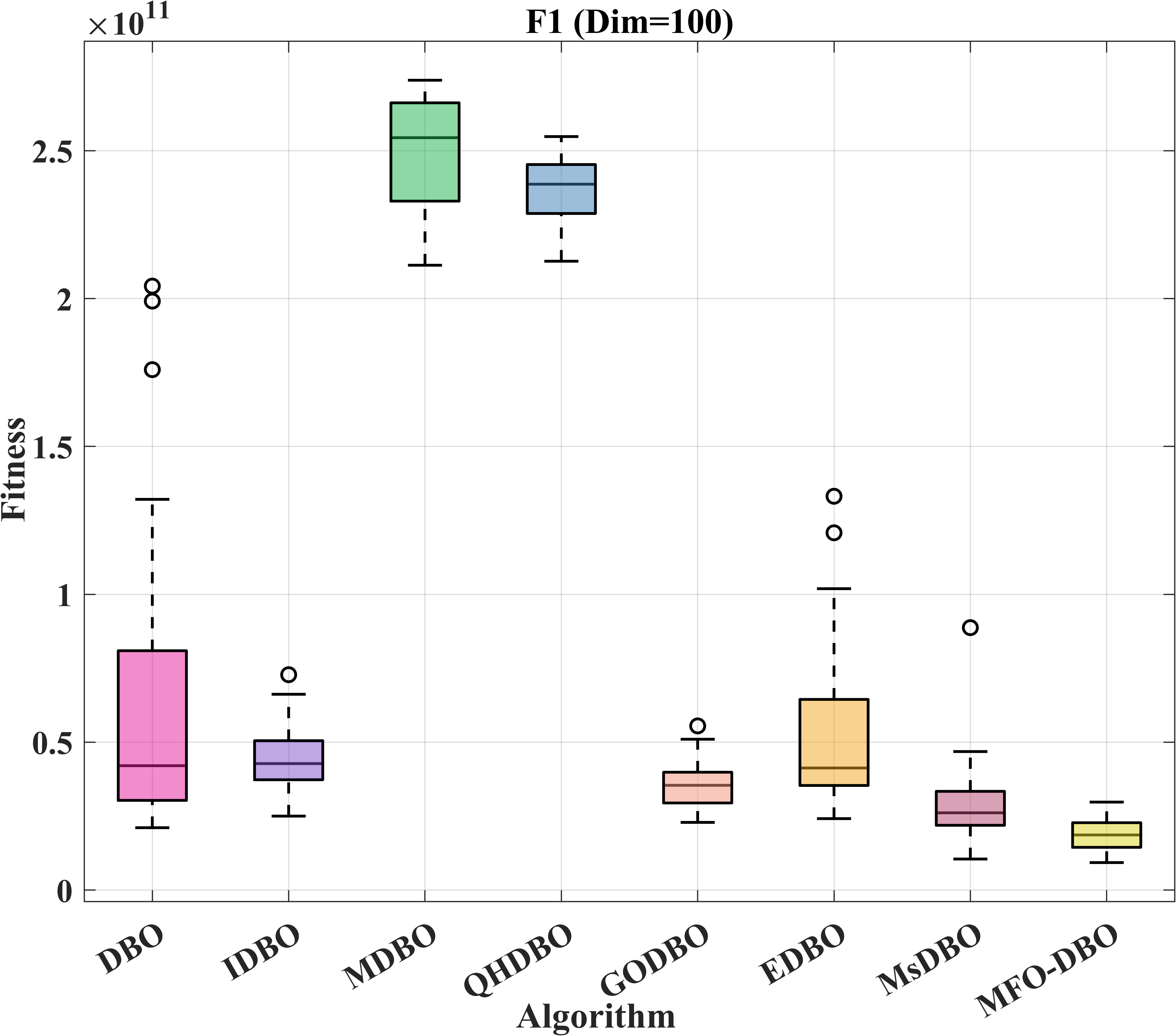}
		\subcaption{F1}
	\end{minipage}
	\hfill
	\begin{minipage}{0.48\textwidth}
		\centering
		\includegraphics[width=\textwidth, height=4.5cm, keepaspectratio]{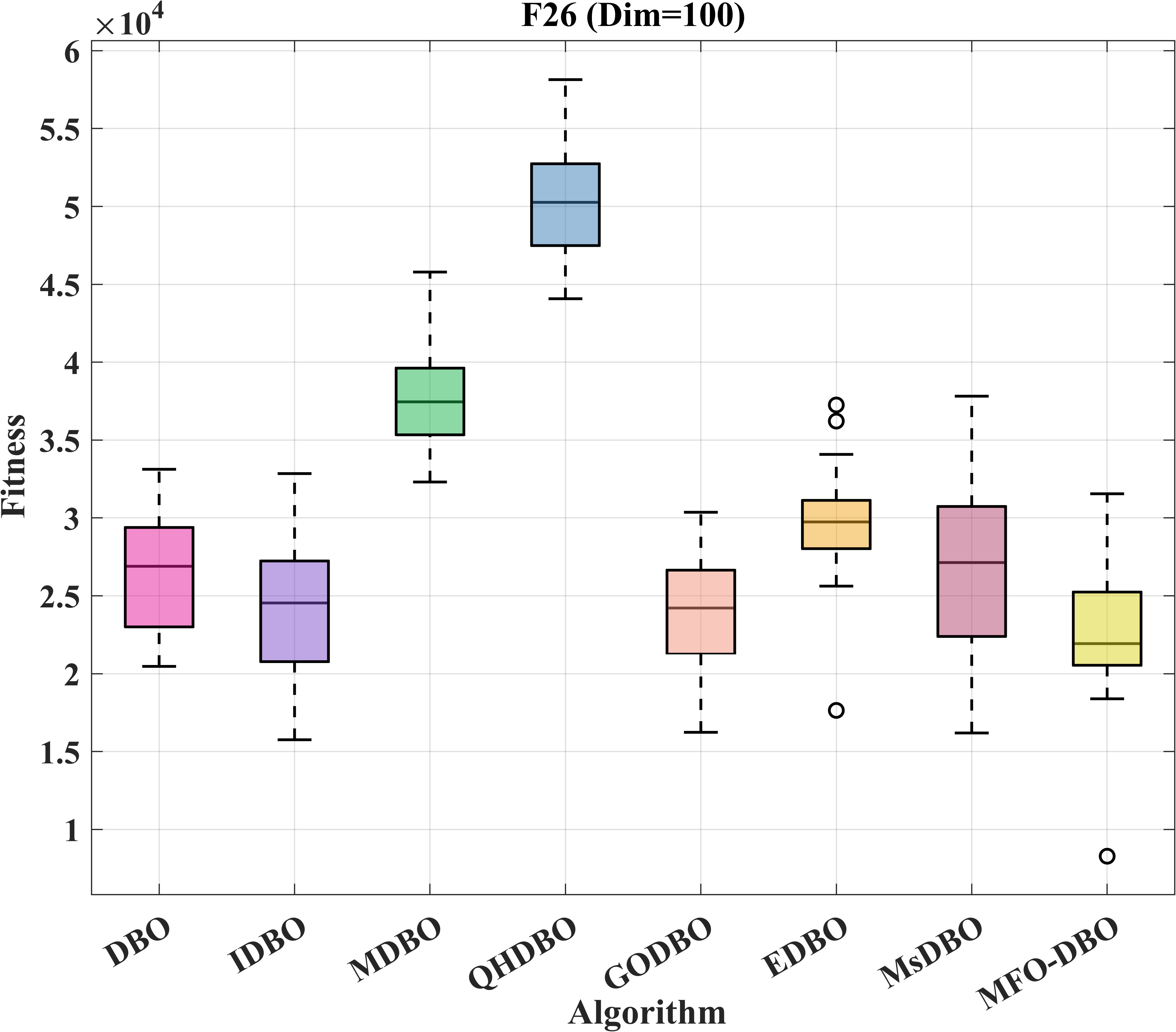}
		\subcaption{F26}
	\end{minipage}
	\caption{Box plots: MFO-DBO vs. DBO Variants on CEC2017.}
	\label{SA_VDBO}
\end{figure}

\autoref{SA_VDBO} presents the stability analysis of MFO-DBO using boxplots on representative CEC2017 functions,
including F11 and F22 (50Dim), and F1 and F26 (100Dim).
The boxplots compare MFO-DBO with several DBO variants.
As shown, MFO-DBO exhibits narrower interquartile ranges and lower medians,
indicating both high stability and superior solution quality across multiple runs.
Algorithms such as MDBO and EDBO also demonstrate relatively stable behavior on certain functions (e.g., \textcolor{blue}{Fig.~\ref{SA_VDBO}(b)}),
but their higher medians indicate suboptimal solution quality.
These results complement the prior statistical analyses and further confirm the robustness and reliability of MFO-DBO in complex optimization tasks.

\subsubsection{Comparisons with Other State-of-the-Art Algorithms} \label{subsubsec4.2.5}

In this section,
a comprehensive performance comparison is conducted to evaluate the competitiveness
and generalization ability of the proposed MFO-DBO algorithm against nine advanced algorithms.
These include two CEC competition-winning algorithms (LSHADE and EBOwithCMAR),
four recently proposed metaheuristics (HEOA, HLOA, IGWO, SCWOA),
and three fractional-order calculus-based optimizers (FOPSO, FOSSA, and FOFPA).
All algorithms are tested under identical experimental settings (see \autoref{parameter} for details).
We assess the performance from five perspectives:
\begin{itemize} [labelsep=0.5em, leftmargin=3em, itemsep=0pt, topsep=0pt, parsep=0pt, partopsep=0pt]
	\item[a)] Mean and standard deviation analysis, presented in~\autoref{VS_CHAMPION} and~\autoref{VS5};
	\item[b)] Wilcoxon rank-sum test, reported in~\autoref{Wilcoxon_VSCHAMPION};
	\item[c)] Friedman test rankings, listed in~\autoref{friedman_VSCHAMPION};
	\item[d)] Convergence analysis, illustrated in~\autoref{CO_ADBO};
	\item[e)] Stability analysis, shown in~\autoref{SA_ADBO}.
\end{itemize}

\noindent
This section presents summary-level  comparisons based on mean and standard deviation,
Wilcoxon $p$-values, and Friedman rankings. Detailed results are provided in the supplementary material (Tables S11–S14).

\subsubsection*{a) Mean and Standard Deviation Analysis}
\addcontentsline{toc}{subsubsection}{A. Mean and Standard Deviation Analysis}
\label{avg_var1}

\begin{figure}[htp]
	\centering
	\includegraphics[width=0.75\textwidth]{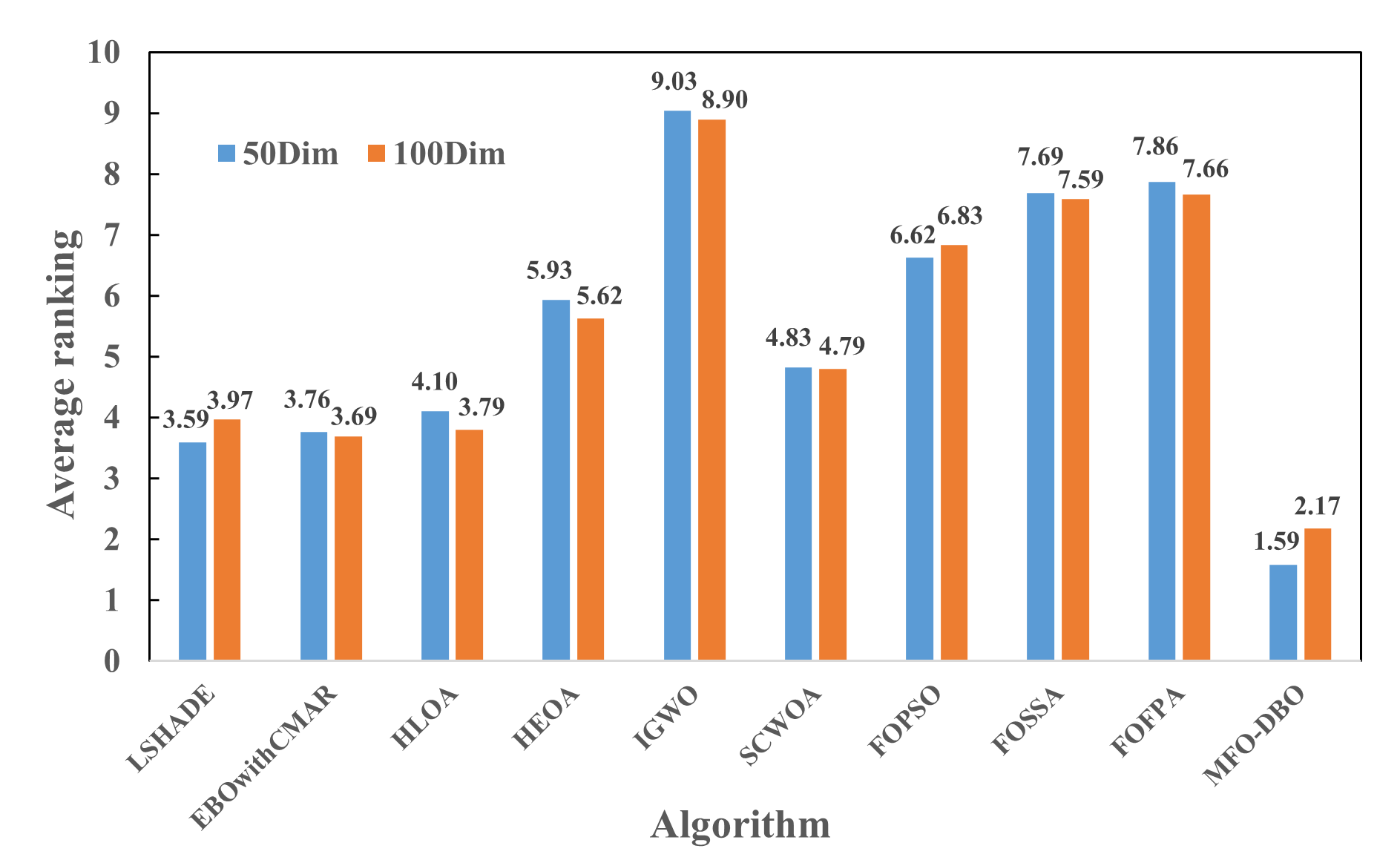}
	\caption{Average rankings of MFO-DBO vs. Advanced Algorithms on CEC2017.}
	\label{VS5}
\end{figure}

According to \autoref{VS_CHAMPION}, MFO-DBO exhibits clear performance advantages over all nine advanced algorithms.
In the 50Dim case,
it outperforms LSHADE, EBOwithCMAR, HEOA, HLOA, IGWO, SCWOA, FOPSO, FOSSA, and FOFPA
on 21, 21, 28, 29, 29, 29, 29, 28, and 29 functions, respectively.
When aggregated over both dimensions, MFO-DBO yields superior results in 90.0\% (470 out of 522) of all comparisons,
including consistent dominance over all three fractional-order algorithms.
This trend is visually reinforced in \autoref{VS5},
where MFO-DBO secures the lowest average rank in both dimensions,
demonstrating its strong generalization ability and competitiveness across a wide range of state-of-the-art optimizers.

\begin{table}[H]
	\centering
	\captionsetup{
		labelsep=newline,
		justification=raggedright,
		singlelinecheck=false,
		labelfont=bf,
		skip=0pt,
		margin=0pt
	}
	\caption{Mean and Standard Deviation: MFO-DBO vs. Advanced Algorithms on CEC2017.}
	\label{VS_CHAMPION}
	\scriptsize
	\resizebox{\columnwidth}{!}{%
		\begin{tabular}{ll llllllllll}
			\toprule
			\multirow{2}{*}{Dimension} &  & \multicolumn{9}{l}{Algorithms} & \multirow{2}{*}{Overall}\\
			\cmidrule(r){3-11}
			& & LSHADE & EBOwithCMAR & HLOA & HEOA & IGWO & SCWOA & FOPSO & FOSSA & FOFPA  & \\
			\midrule
			50Dim & $B|S|W$  &$21|0|8$ & $21|1|7$ & $28|0|1$ & $29|0|0$ & $29|0|0$ & $29|0|0$ & $29|0|0$ & $28|0|1$ & $29|0|0$ &$243|1|17$\\
			%\midrule
			100Dim	& $B|S|W$ & $20|0|9$ & $17|0|12$ & $20|0|9$ & $28|0|1$ & $29|0|0$ & $28|0|1$ & $29|0|0$ & $27|0|2$ & $29|0|0$ & $227|0|34$ \\
			\midrule
			Overall  &  $B|S|W$  & $41|0|17$   & $38|1|19$  & $48|0|10$ & $57|0|1$  & $58|0|0 $ &  $57|0|1$  & $58|0|0 $  & $55|0|3 $ & $58|0|0 $& $470|1|51$\\
			\bottomrule	
		\end{tabular}%	
	}
\end{table}

\subsubsection*{b) Wilcoxon Rank-Sum Test}
\addcontentsline{toc}{subsubsection}{B. Wilcoxon Rank-Sum Test}
\label{wilcoxon1}

The Wilcoxon test results comparing MFO-DBO with nine advanced algorithms are presented in~\autoref{Wilcoxon_VSCHAMPION}.
Compared to 4.2.4, the statistical advantage of MFO-DBO becomes even more pronounced.
In the 50Dim case, it achieves significant wins on 21, 20, 24, 29, 29, 29, 29, 28, and 29 functions over
LSHADE, EBOwithCMAR, HLOA, HEOA, IGWO, SCWOA, FOPSO, FOSSA, and FOFPA, respectively.
Across both dimensions, MFO-DBO secures 461 significant wins out of 522 pairwise comparisons (88.3\%),
demonstrating broader and more consistent superiority.
Notably, MFO-DBO maintains dominance across both competition-winning algorithms and fractional-order optimizers,
and is outperformed in only 37 cases.
These results underscore the algorithm’s strong generalization ability and statistical robustness against state-of-the-art competitors.

\begin{table}[H]
	\centering
	\captionsetup{
		labelsep=newline,
		justification=raggedright,
		singlelinecheck=false,
		labelfont=bf,
		skip=0pt,
		margin=0pt
	}
	\caption{Wilcoxon rank sum test results: MFO-DBO vs. Advanced Algorithms on CEC2017.}
	\label{Wilcoxon_VSCHAMPION}
	\scriptsize
	\resizebox{\columnwidth}{!}{%
		\begin{tabular}{ll llllllllll}
			\toprule
			\multirow{2}{*}{Dimension} &  & \multicolumn{9}{l}{Algorithms}& \multirow{2}{*}{Overall} \\
			\cmidrule(r){3-11}
			& & LSHADE & EBOwithCMAR & HLOA & HEOA & IGWO & SCWOA & FOPSO & FOSSA & FOFPA & \\
			\midrule
			50Dim & $+/\approx/-$  &21/1/7 & 20/2/7 & 24/5/0 & 29/0/0 & 29/0/0 & 29/0/0 & 29/0/0 & 28/1/0& 29/0/0& 238/9/14\\
			100Dim	& $+/\approx/-$ & 20/0/9 & 17/0/12 & 20/4/5 & 27/1/1 & 29/0/0 & 28/1/0 & 29/0/0 & 25/2/2 & 28/1/0 & 223/9/29 \\
			\midrule
			Overall  & $+/\approx/-$  & 41/1/16   & 37/2/19  & 44/9/5 & 56/1/1  & 58/0/0 &  57/1/0  & 58/0/0  & 53/3/2 & 57/1/0  & 461/18/43\\
			\bottomrule	
		\end{tabular}%	
	}
\end{table}

\subsubsection*{c) Friedman test rankings}
\addcontentsline{toc}{subsubsection}{C. Friedman test rankings}
\label{Cfriedman1}

The Friedman test is conducted to assess and rank the performance of MFO-DBO against the advanced algorithms,
with results shown in~\autoref{friedman_VSCHAMPION}. MFO-DBO attains the lowest average rank across both 50Dim and 100Dim tasks, underscoring its consistent superiority and robustness.
\begin{table}[H]
	\centering
	\captionsetup{
		labelsep=newline,
		justification=raggedright,
		singlelinecheck=false,
		labelfont=bf,
		skip=0pt,
		margin=0pt
	}
	\caption{Friedman test rankings: MFO-DBO vs. Advanced Algorithms on CEC2017.}
	\label{friedman_VSCHAMPION}
	\scriptsize
	\renewcommand{\arraystretch}{1.2}
	\resizebox{\columnwidth}{!}{%
		\begin{tabular}{llllllllllll}
			\toprule
			\multicolumn{1}{l}{Dimension} & \multicolumn{11}{l}{Algorithm and Frideman} \\
			\midrule
			\multirow{2}{*}{50Dim} & Algorithm & MFO-DBO & LSHADE & EBOwithCMAR & HLOA & SCWOA & HEOA & FOPSO  & FOSSA & FOFPA  & IGWO  \\
			& Friedman  & 1.83 & 3.38 & 3.41 & 4.41 & 4.93 & 5.76 & 6.62  & 7.59 & 7.79 & 9.28\\
			\multirow{2}{*}{100Dim} & Algorithm & MFO-DBO & EBOwithCMAR & LSHADE  & HLOA  & SCWOA & HEOA   & FOPSO & FOSSA & FOFPA & IGWO  \\
			& Friedman   & 2.38  & 3.45 & 3.72    & 3.93  & 4.86  &  5.69  & 6.83 & 7.55 & 7.59 & 9.00\\
			\bottomrule
		\end{tabular}
	}
\end{table}

\subsubsection*{d) Convergence analysis}
\addcontentsline{toc}{subsubsection}{D. Convergence analysis}
\label{convergence1}

\begin{figure}[H]
	\centering
	\begin{minipage}{0.48\textwidth}
		\centering
		\includegraphics[width=\textwidth, height=4.cm, keepaspectratio]{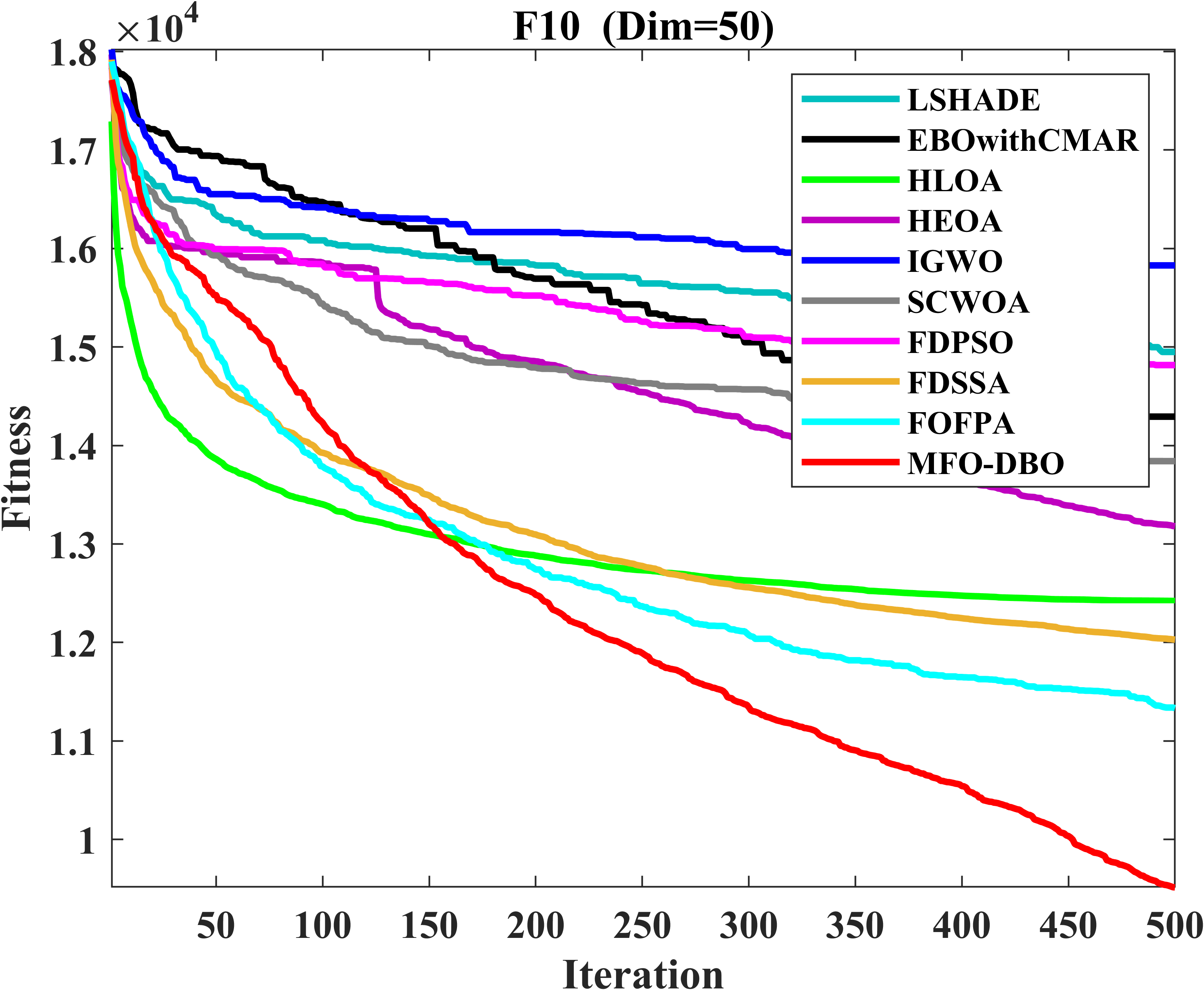}
		\subcaption{F10}
	\end{minipage}
	\hfill
	\begin{minipage}{0.48\textwidth}
		\centering
		\includegraphics[width=\textwidth, height=4.5cm, keepaspectratio]{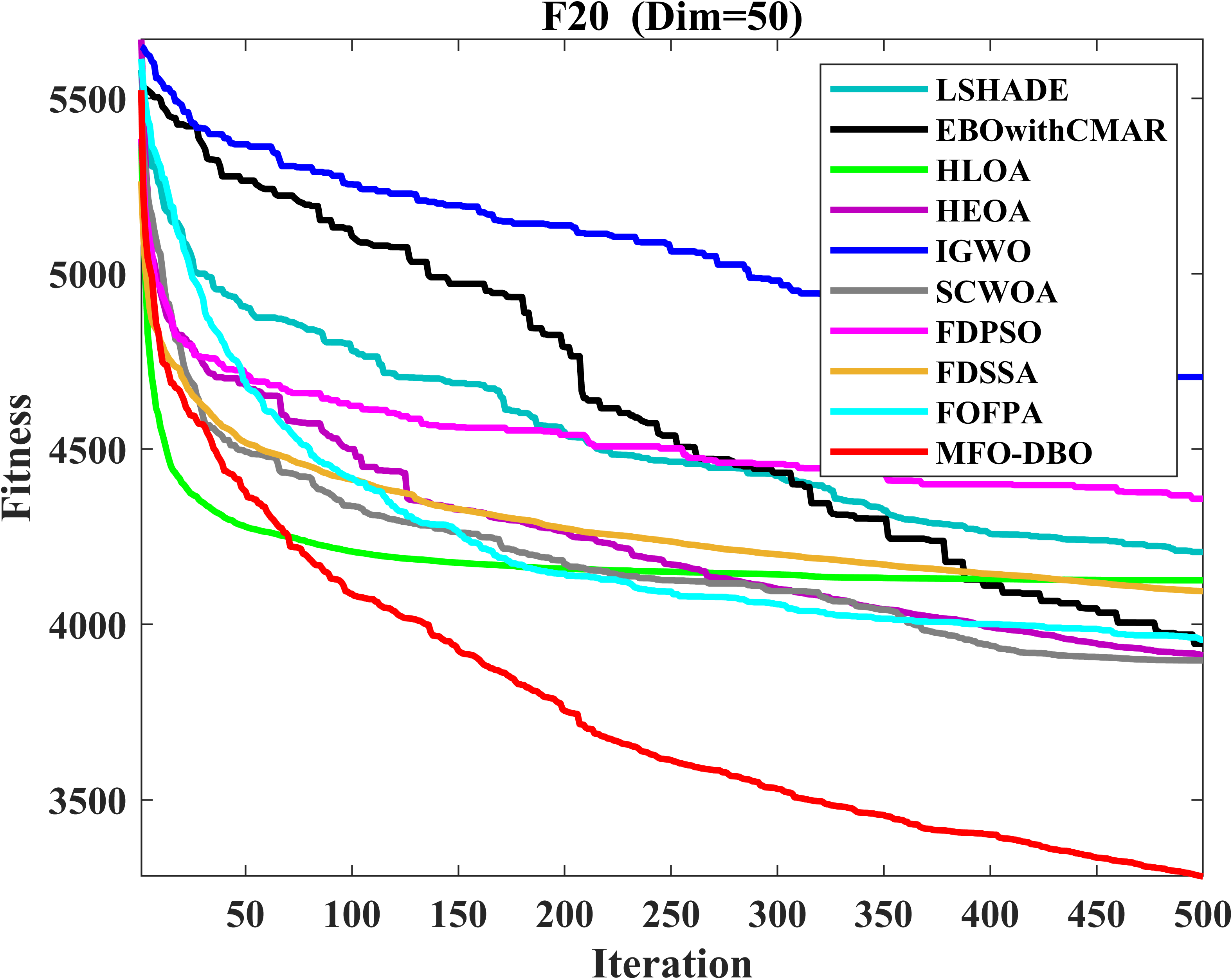}
		\subcaption{F20}
	\end{minipage}
	
	\vspace{0.5cm}
	
	\begin{minipage}{0.48\textwidth}
		\centering
		\includegraphics[width=\textwidth, height=4.5cm, keepaspectratio]{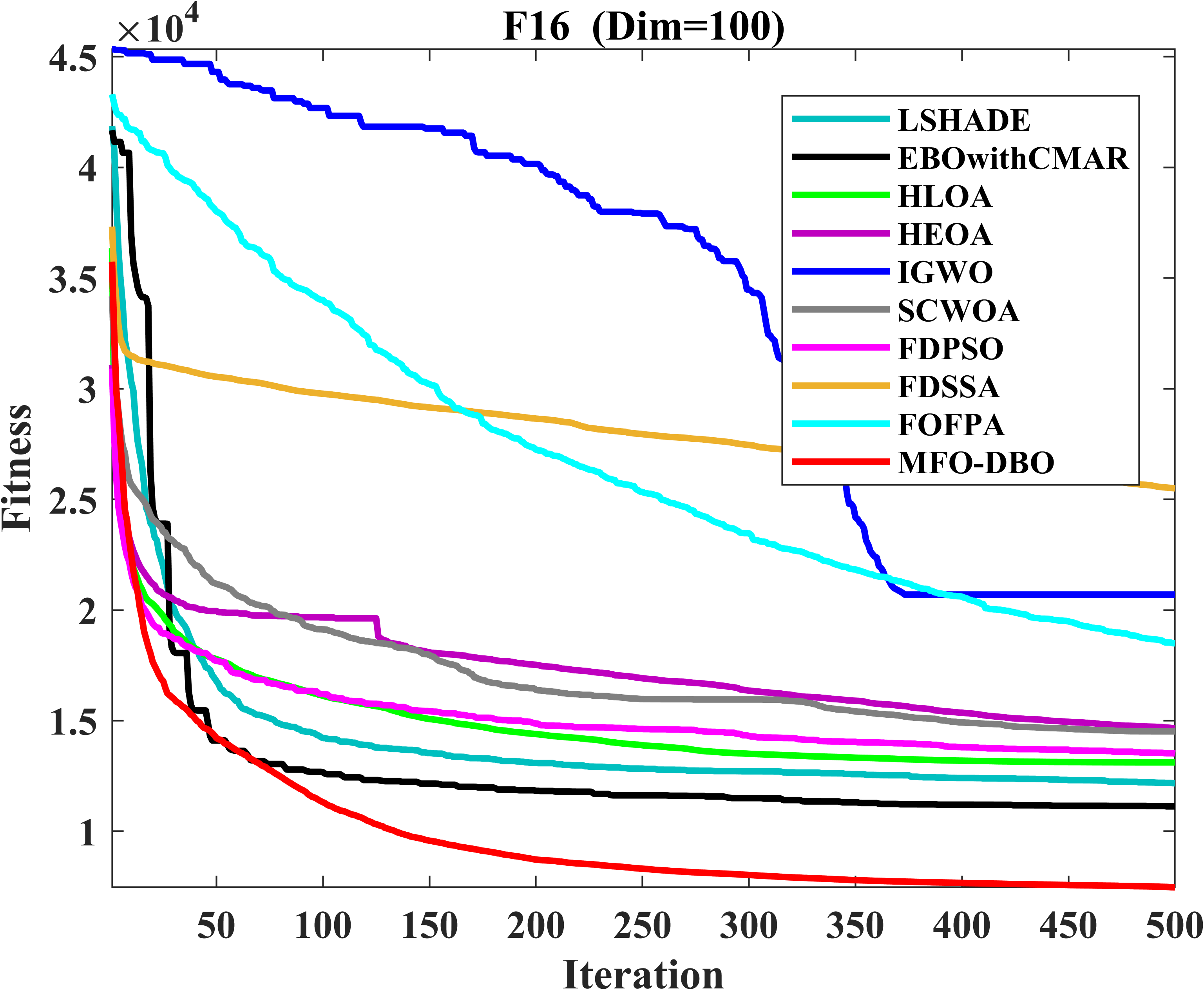}
		\subcaption{F16}
	\end{minipage}
	\hfill
	\begin{minipage}{0.48\textwidth}
		\centering
		\includegraphics[width=\textwidth, height=4.5cm, keepaspectratio]{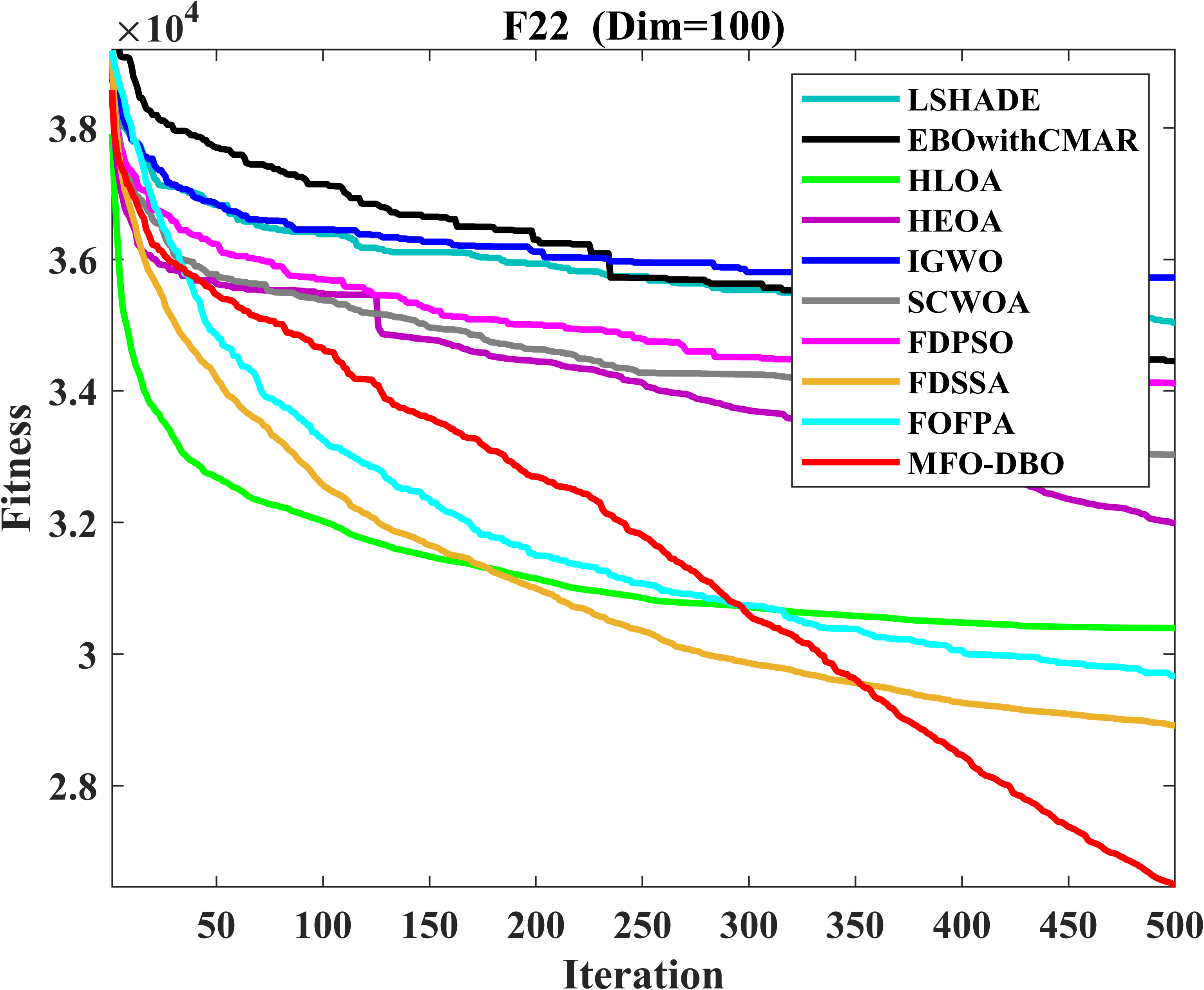}
		\subcaption{F22}
	\end{minipage}
	\caption{Convergence curves: MFO-DBO vs. Advanced Algorithms on CEC2017.}
	\label{CO_ADBO}
\end{figure}

\autoref{CO_ADBO} illustrates the convergence behavior of MFO-DBO compared with advanced algorithms on the CEC2017 test suite.
Functions F10 and F20 (50Dim), and F16 and F22 (100Dim), are selected as representatives.
As shown in the figure,
MFO-DBO consistently achieves faster convergence and better solution quality across both dimensional settings.
These results highlight the algorithm’s superior convergence speed and search efficiency relative to state-of-the-art competitors.

\subsubsection*{e) Stability analysis}
\addcontentsline{toc}{subsubsection}{E. Stability analysis}
\label{Stability1}

\begin{figure}[H]
	\centering
	\begin{minipage}{0.48\textwidth}
		\centering
		\includegraphics[width=\textwidth, height=4.5cm, keepaspectratio]{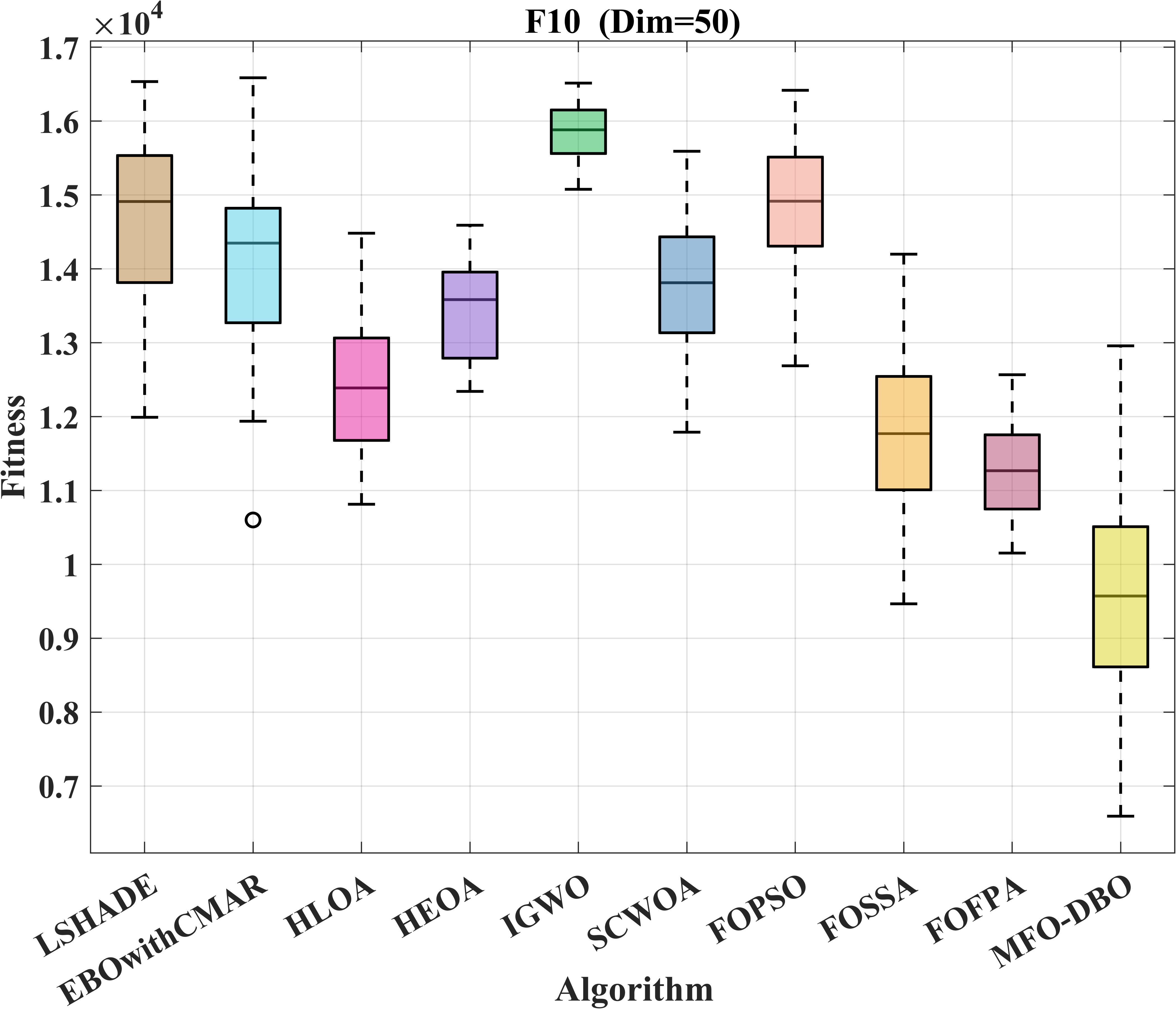}
		\subcaption{F10}
	\end{minipage}
	\hfill
	\begin{minipage}{0.48\textwidth}
		\centering
		\includegraphics[width=\textwidth, height=4.5cm, keepaspectratio]{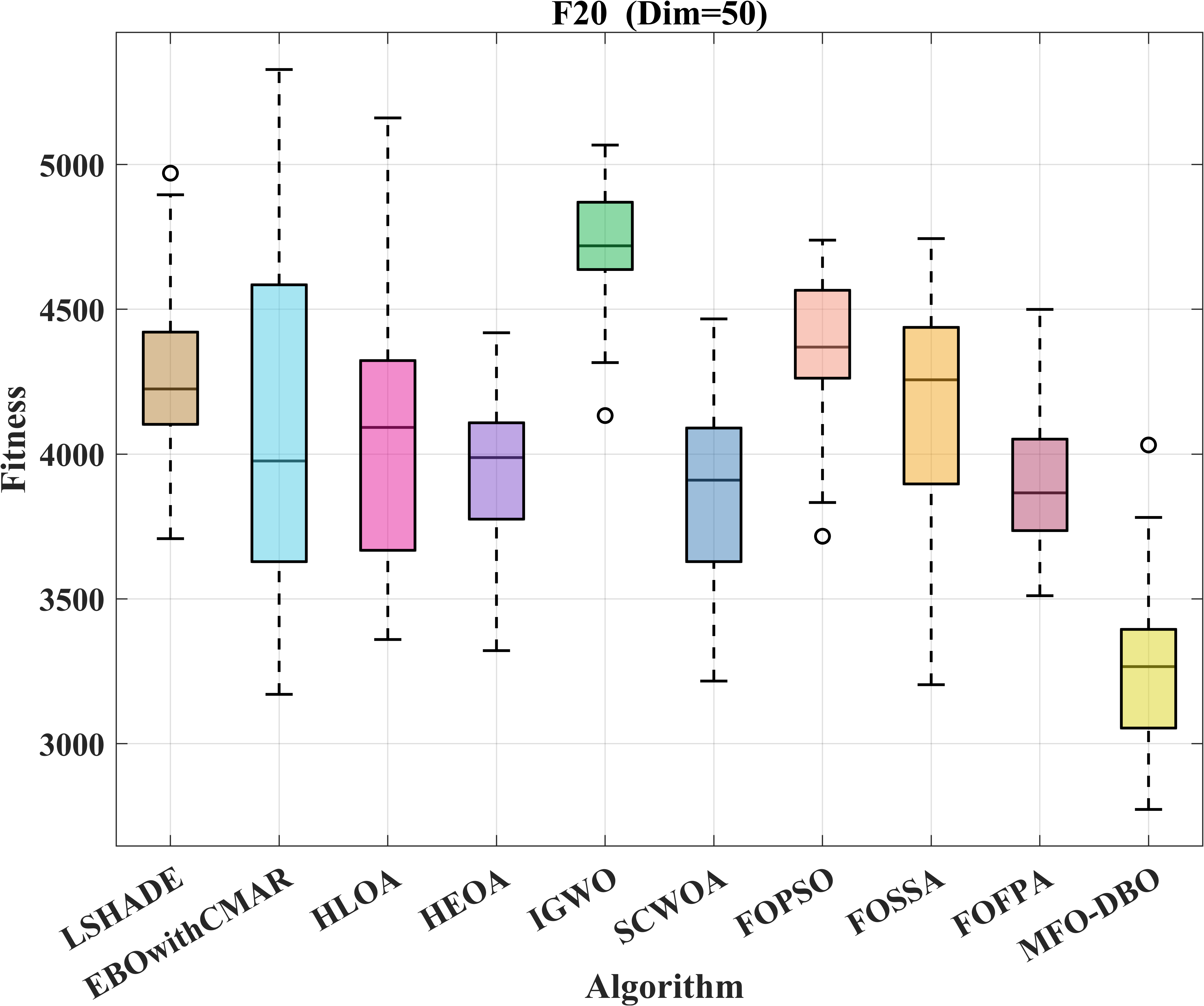}
		\subcaption{F20}
	\end{minipage}
	
	\vspace{0.5cm}
	
	\begin{minipage}{0.48\textwidth}
		\centering
		\includegraphics[width=\textwidth, height=4.5cm, keepaspectratio]{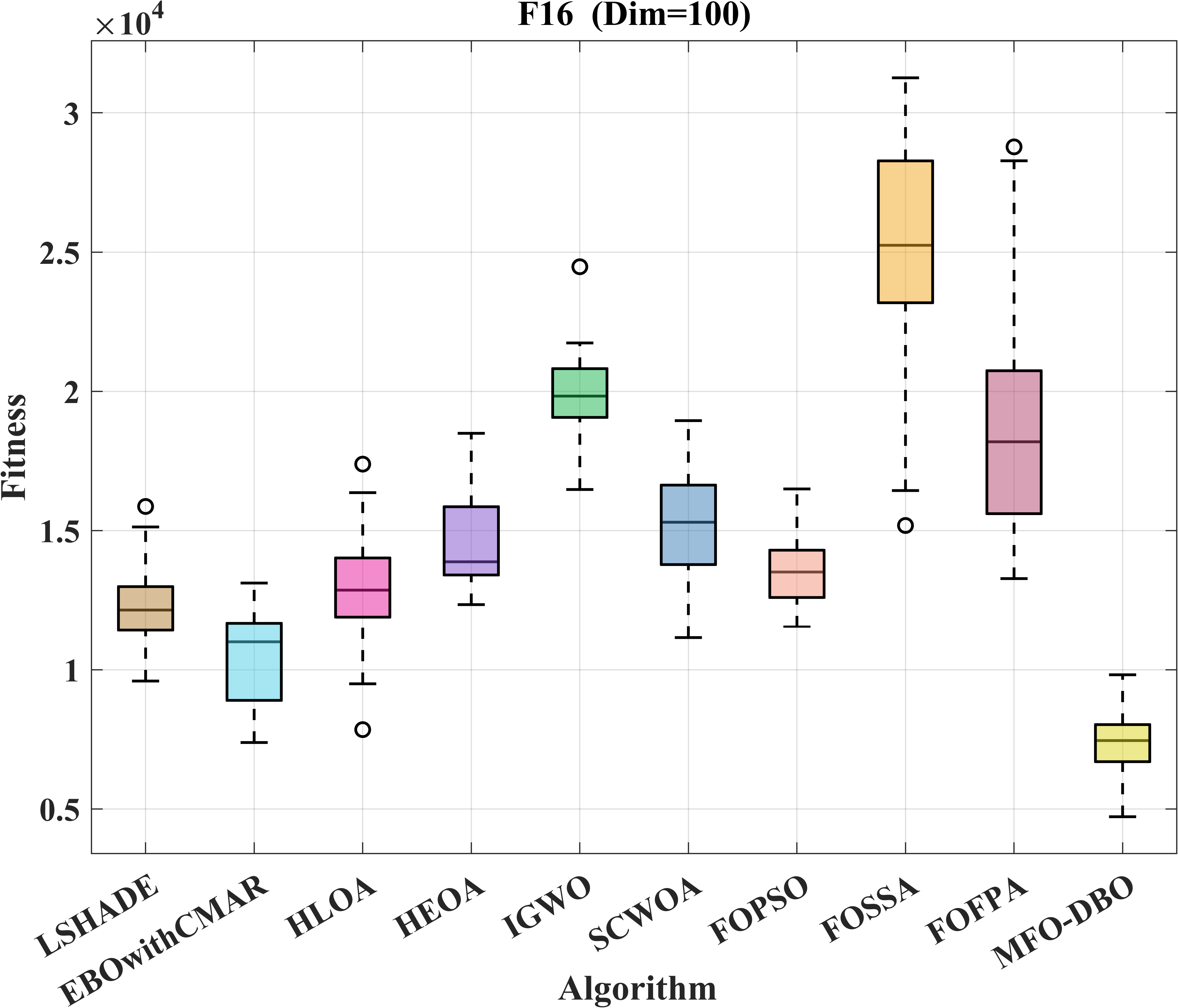}
		\subcaption{F16}
	\end{minipage}
	\hfill
	\begin{minipage}{0.48\textwidth}
		\centering
		\includegraphics[width=\textwidth, height=4.5cm, keepaspectratio]{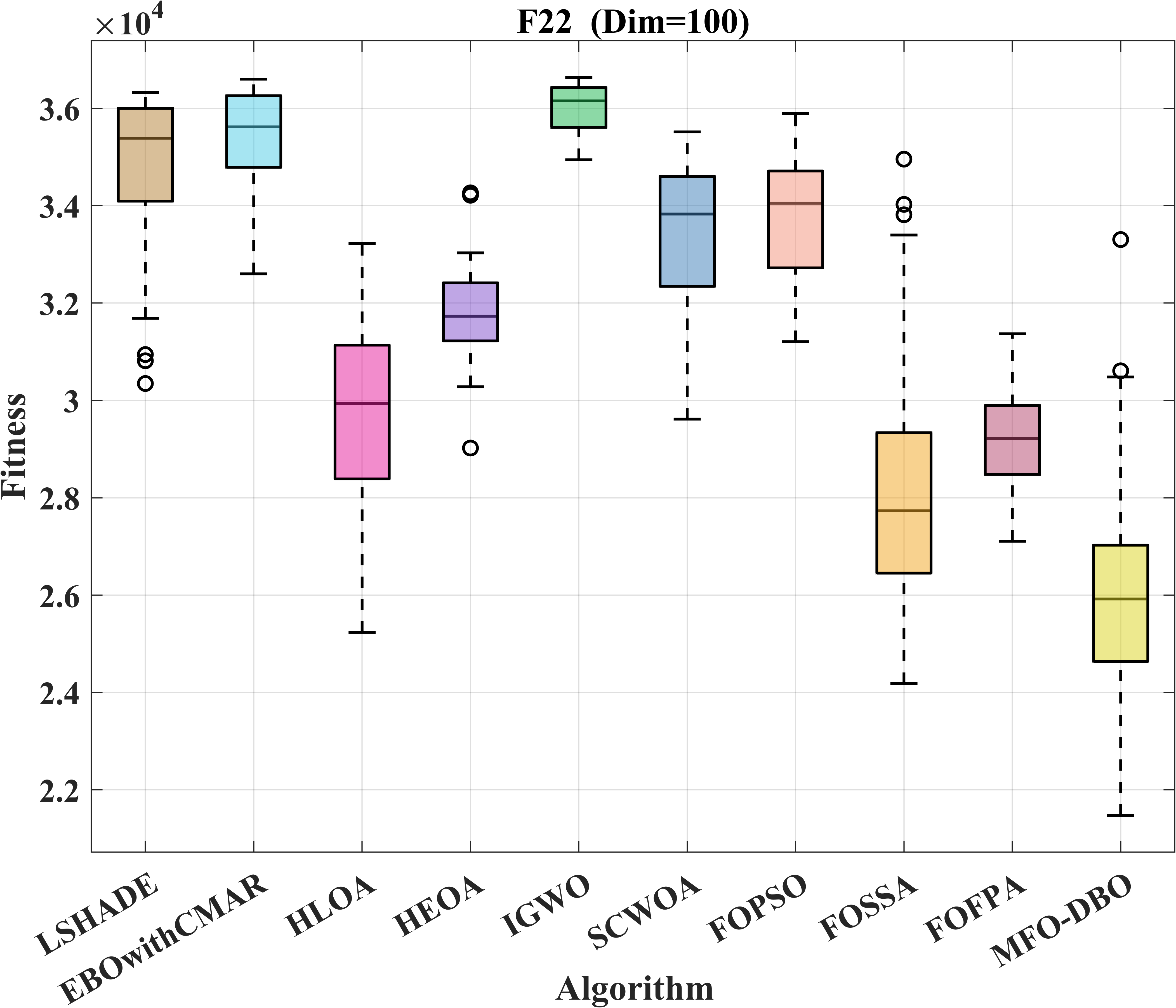}
		\subcaption{F22}
	\end{minipage}
	\caption{Box plots: MFO-DBO vs. Advanced Algorithms on CEC2017.}
	\label{SA_ADBO}
\end{figure}

\autoref{SA_ADBO} presents boxplot comparisons between MFO-DBO and the competing algorithms
on several representative functions from the CEC2017 test suite, including F10 and F20 (50Dim) and F16 and F22 (100Dim).
As shown in the figure, MFO-DBO exhibits more concentrated data distributions,
reflecting superior stability and consistently high-quality solutions.
For instance, in \textcolor{blue}{Fig.~\ref{SA_ADBO}(d)},
while EBOwithCMAR, HEOA, IGWO, and FOFPA demonstrate relatively stable performance,
their boxplots are generally positioned higher, indicating suboptimal solution quality.
Compared to these advanced methods, MFO-DBO achieves better stability and accuracy,
especially in high-dimensional problems.
This further underscoring its performance advantage.

\subsection{Exploration and Exploitation in MFO-DBO} \label{subsec4.3}

\begin{figure}[H]
	\centering
	\begin{minipage}{0.48\textwidth}
		\centering
		\includegraphics[width=\textwidth, height=4.5cm, keepaspectratio]{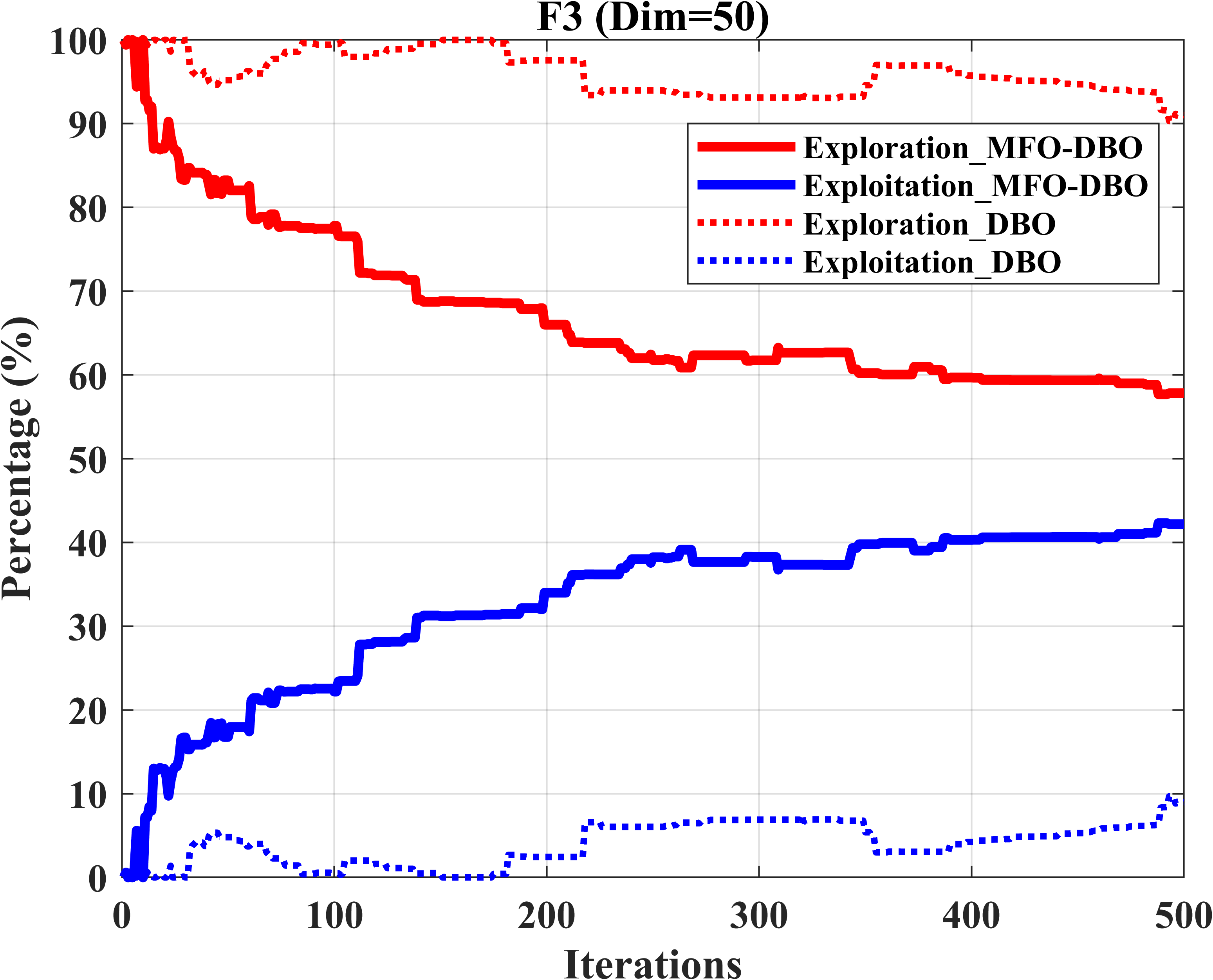}
		\subcaption{F3}
	\end{minipage}
	\hfill
	\begin{minipage}{0.48\textwidth}
		\centering
		\includegraphics[width=\textwidth, height=4.5cm, keepaspectratio]{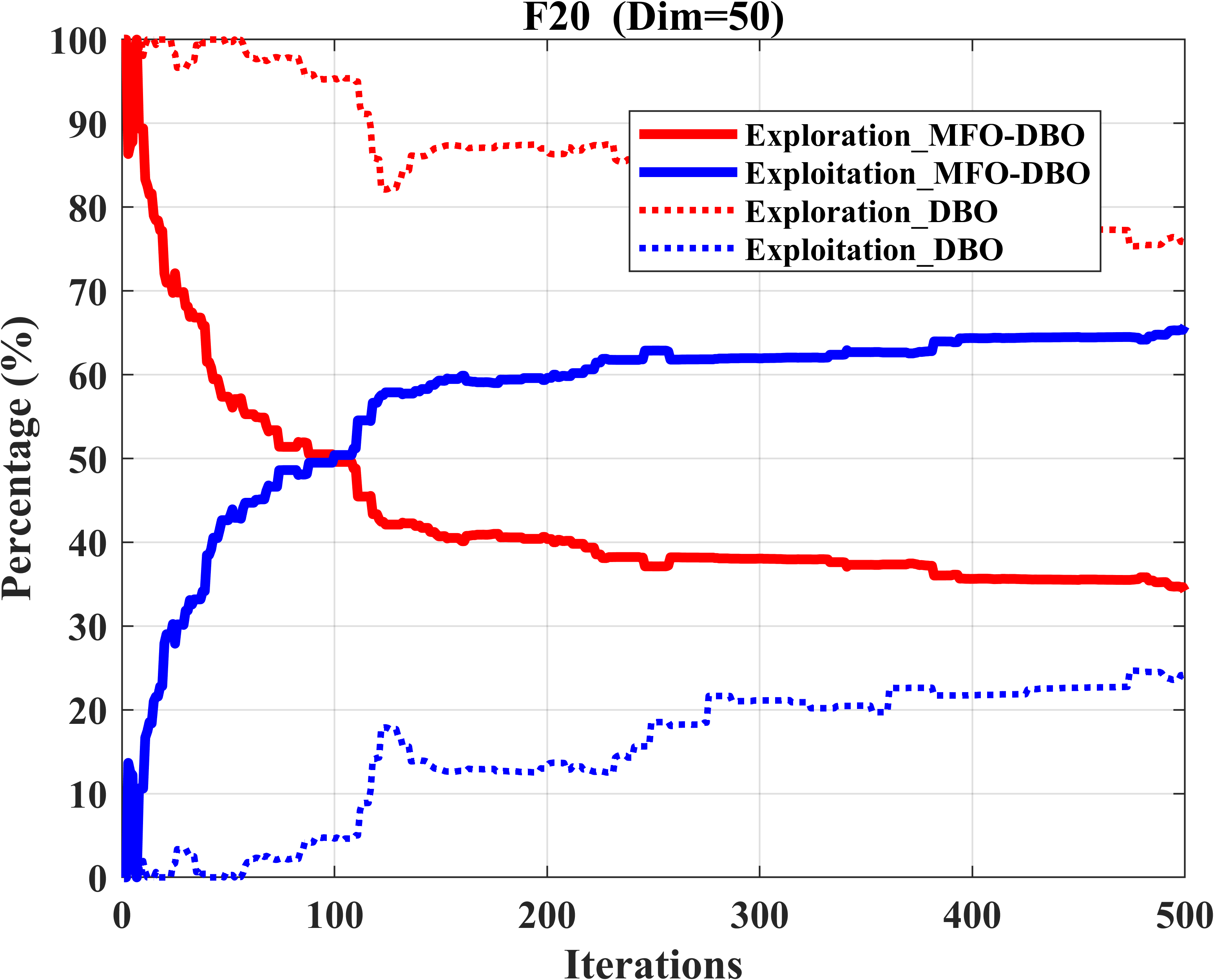}
		\subcaption{F20}
	\end{minipage}
	
	\begin{minipage}{0.48\textwidth}
		\centering
		\includegraphics[width=\textwidth, height=4.5cm, keepaspectratio]{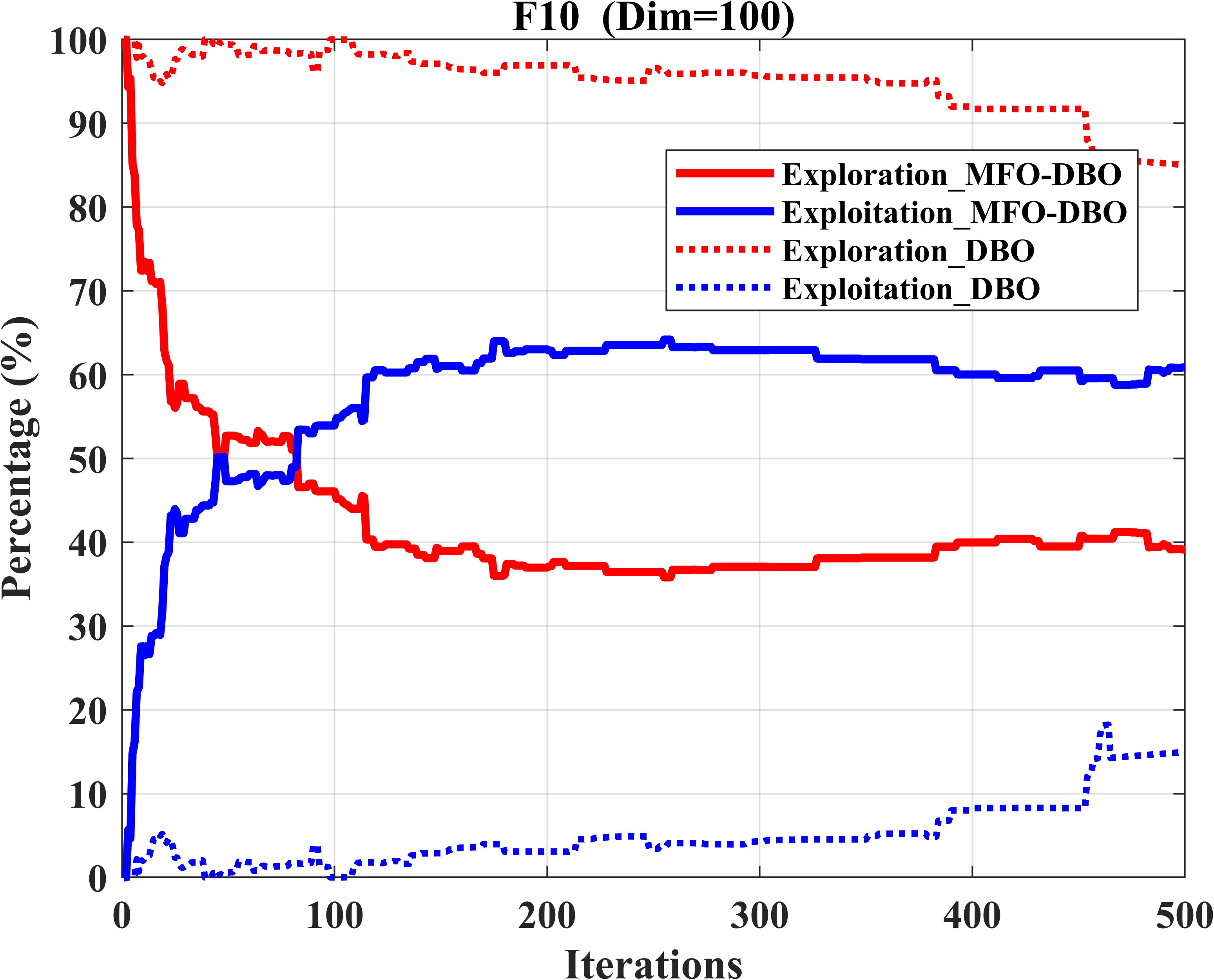}
		\subcaption{F10}
	\end{minipage}
	\hfill
	\begin{minipage}{0.48\textwidth}
		\centering
		\includegraphics[width=\textwidth, height=4.5cm, keepaspectratio]{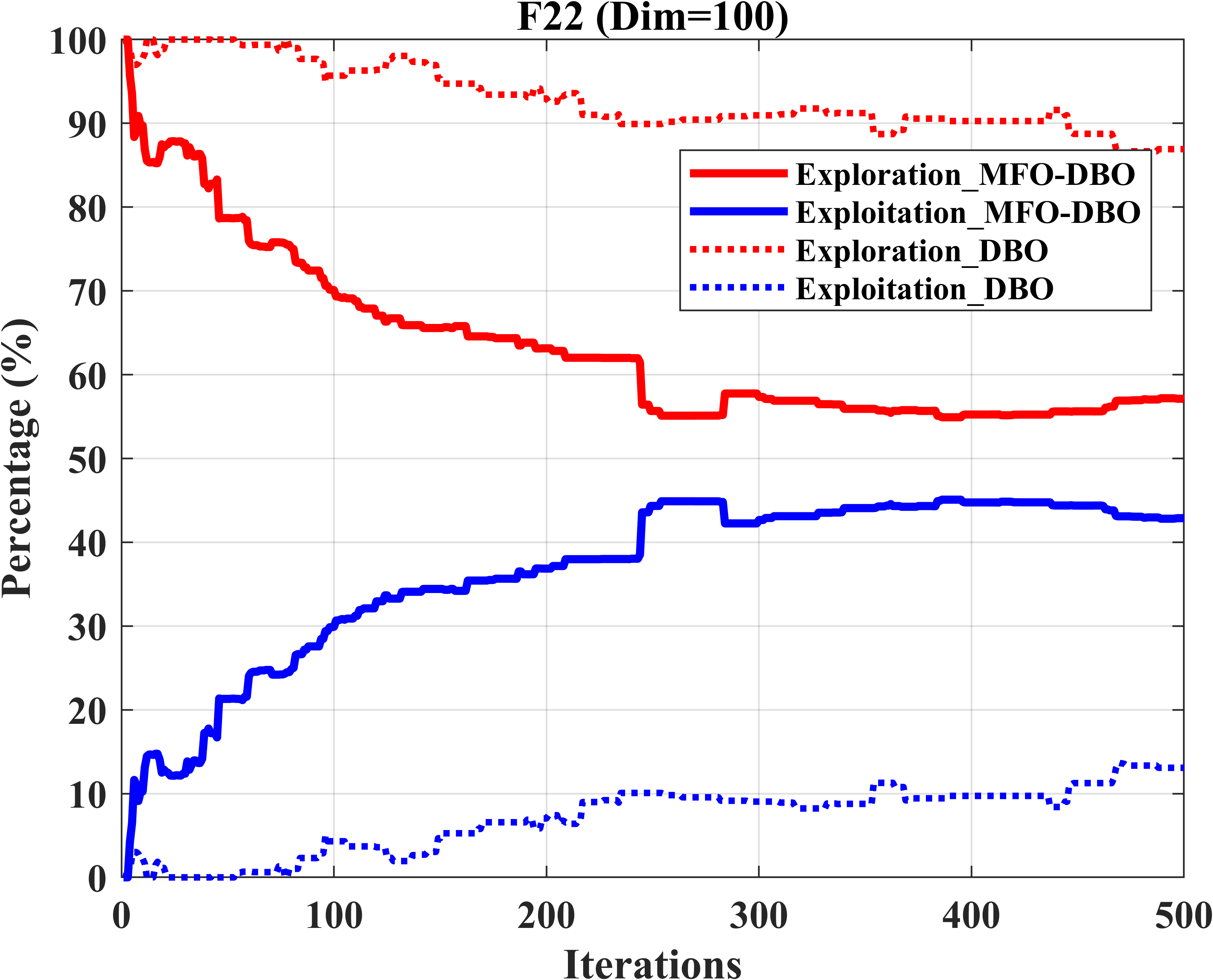}
		\subcaption{F22}
	\end{minipage}
	\caption{Box plots: MFO-DBO vs. Advanced Algorithms on CEC2017.}
	\label{EE_DBO}
\end{figure}

To further understand the search dynamics of the proposed MFO-DBO algorithm,
we analyze its balance between exploration and exploitation—two fundamental components
that jointly determine global search ability and convergence efficiency.
Adequate exploration helps discover diverse promising regions,
while timely exploitation improves solution refinement~\citep{yu2025multi}.
To evaluate this balance,
we conducted comparative experiments with the standard DBO algorithm on selected functions from the CEC2017 benchmark suite,
including F3 and F20 (50Dim), and F10 and F22 (100Dim).
The corresponding exploration and exploitation trends are shown in~\autoref{EE_DBO}.

As illustrated in~\autoref{EE_DBO},
MFO-DBO demonstrates an efficient transition from exploration to exploitation. Its exploration rate decreases rapidly in early iterations,
enabling a wide search,
while the exploitation rate rises steadily to refine solutions during later stages.
In contrast, DBO maintains a persistently high exploration level with limited exploitation growth,
which weakens its convergence capability.
These results indicate that MFO-DBO achieves a more balanced search behavior,
effectively mitigating premature convergence and enhancing optimization performance.

\section{Parameter Identification in Photovoltaic Systems} \label{sec5}

In this section, we apply the proposed MFO-DBO algorithm to the parameter identification in three typical photovoltaic (PV) models,
including the Single Diode Model (SDM), Double Diode Model (DDM), and PV Module Model.
Numerical experiments are conducted using benchmark current–voltage (I–V) data to evaluate the algorithm's accuracy and robustness.

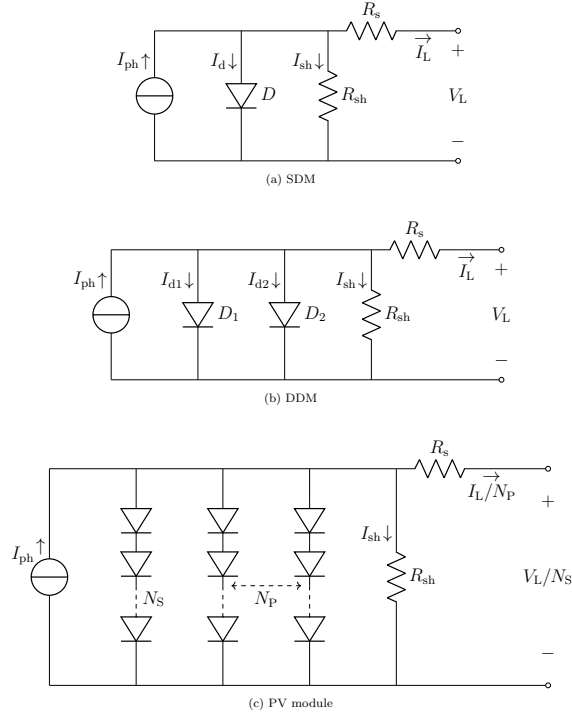
\begin{figure}[H]
	\centering
	\resizebox{0.6\textwidth}{!}{%
		\begin{tabular}{c}
			\begin{subfigure}{\textwidth}
				\centering
				\begin{circuitikz}[american voltages]
					\draw (0,0) to[I, l] (0,3);
					\draw (2,3) to[D, l=$D$] (2,0);
					\draw (4,3) to[R, l=$R_{\mathrm{sh}}$] (4,0);
					\draw (4,3) to[R, l=$R_{\mathrm{s}}$] (6,3);
					\draw (0,3) to[short, -] (4,3);
					\draw (6,3) to[short, -o] (7,3);
					\draw (0,0) to[short, -o] (7,0);
					\draw (7,3) to[open, v=$V_{\mathrm{L}}$] (7,0);
					
					\node at (-0.2,2.3) {$\uparrow$};
					\node at (-0.2,2.3) [left] {$I_\mathrm{ph}$};
					\node at (1.8,2.3) {$\downarrow$};
					\node at (1.8,2.3) [left] {$I_\mathrm{d}$};
					\node at (3.8,2.3) {$\downarrow$};
					\node at (3.8,2.3) [left] {$I_\mathrm{sh}$};
					\node at (6.2,2.8) {$\rightarrow$};
					\node at (6.2,2.8) [below] {$I_\mathrm{L}$};
				\end{circuitikz}
				\subcaption{SDM}
			\end{subfigure}
			\\[0.5cm]
			
			\begin{subfigure}{\textwidth}
				\centering
				\begin{circuitikz}[american voltages]
					\draw (0,0) to[I, l] (0,3);
					\draw (2,3) to[D, l=$D_{1}$] (2,0);
					\draw (4,3) to[D, l=$D_{2}$] (4,0);
					\draw (6,3) to[R, l=$R_{\mathrm{sh}}$] (6,0);
					\draw (6,3) to[R, l=$R_{\mathrm{s}}$] (8,3);
					\draw (0,3) to[short, -] (6,3);
					\draw (8,3) to[short, -o] (9,3);
					\draw (0,0) to[short, -o] (9,0);
					\draw (9,3) to[open, v=$V_{\mathrm{L}}$] (9,0);
					
					\node at (-0.2,2.3) {$\uparrow$};
					\node at (-0.2,2.3) [left] {$I_\mathrm{ph}$};
					\node at (1.8,2.3) {$\downarrow$};
					\node at (1.8,2.3) [left] {$I_\mathrm{d1}$};
					\node at (3.8,2.3) {$\downarrow$};
					\node at (3.8,2.3) [left] {$I_\mathrm{d2}$};
					\node at (5.8,2.3) {$\downarrow$};
					\node at (5.8,2.3) [left] {$I_\mathrm{sh}$};
					\node at (8.2,2.8) {$\rightarrow$};
					\node at (8.2,2.8) [below] {$I_\mathrm{L}$};
				\end{circuitikz}
				\subcaption{DDM}
			\end{subfigure}
			\\[0.5cm]
			
			\begin{subfigure}{\textwidth}
				\centering
				\begin{circuitikz}[american voltages]
					\draw (0,0) to[I, l] (0,5);
					\draw (2,5) to[short, -] (2,4.3);
					\draw (2,4.3) to[D] (2,3.3);
					\draw (2,3.3) to[D] (2,2.3);
					\draw[dashed] (2,2.3) -- (2,1.5);
					\draw (2,1.6) to[D] (2,1);
					\draw (2,1) to[short, -] (2,0);
					\draw (4,5) to[short, -] (4,4.3);
					\draw (4,4.3) to[D] (4,3.3);
					\draw (4,3.3) to[D] (4,2.3);
					\draw[dashed] (4,2.3) -- (4,1.5);
					\draw (4,1.6) to[D] (4,1);
					\draw (4,1) to[short, -] (4,0);
					\draw (6,5) to[short, -] (6,4.3);
					\draw (6,4.3) to[D] (6,3.3);
					\draw (6,3.3) to[D] (6,2.3);
					\draw[dashed] (6,2.3) -- (6,1.5);
					\draw (6,1.6) to[D] (6,1);
					\draw (6,1) to[short, -] (6,0);
					\draw (8,5) to[R, l=$R_{\mathrm{sh}}$] (8,0);
					\draw (8,5) to[R, l=$R_{\mathrm{s}}$] (10,5);
					\draw (0,5) to[short, -] (8,5);
					\draw (10,5) to[short, -o] (11.5,5);
					\draw (0,0) to[short, -o] (11.5,0);
					\draw (11.5,5) to[open, v=$V_{\mathrm{L}}/N_\mathrm{S}$] (11.5,0);
					\draw[dashed,<->] (4.2,2.3) -- (5.8,2.3) node[midway, below] {$N_\mathrm{P}$};
					\node at (-0.2,3.2) {$\uparrow$};
					\node at (-0.2,3) [left] {$I_\mathrm{ph}$};
					\node at (2,2) [right] {$N_\mathrm{S}$};
					\node at (7.8,3.5) {$\downarrow$};
					\node at (7.8,3.5) [left] {$I_\mathrm{sh}$};
					\node at (10.2,4.8) {$\rightarrow$};
					\node at (10.2,4.8) [below] {$I_\mathrm{L}/N_\mathrm{P}$};
				\end{circuitikz}
				\subcaption{PV module}
			\end{subfigure}
			
		\end{tabular}
	}
	\caption{Equivalent circuit diagrams for photovoltaic cells.}
	\label{fig5}
\end{figure}

\subsection{Problem statement} \label{subsec5.1}

Parameter identification in PV system involves using optimization techniques to estimate key circuit parameters,
thereby minimizing the disparity between model-predicted values and experimental measurements.
\autoref{fig5} illustrates the equivalent circuit diagrams of the three models considered in this study,
namely SDM, DDM, and PV Module Model.

\subsubsection*{(1) Single Diode Model }
\addcontentsline{toc}{subsubsection}{(1). Single Diode Model}
\label{A_SDM}

The SDM is widely used due to its balance between modeling accuracy and simplicity.
It consists of five key parameters: photogenerated current (${I_{ph}}$),
reverse saturation current ($I_{sd}$), series resistance ($R_s$), shunt resistance ($R_{sh}$),
and diode ideality factor ($n$).
The governing mathematical equations for SDM are:

\begin{equation}
	I_L = I_{ph} - I_{sh} - I_d,
	\label{eq19}
\end{equation}

\begin{equation}
	I_d = I_{sd} \cdot \left[\exp \left( \frac{(I_L \cdot R_s + V_L) \cdot q}{T \cdot n \cdot k} \right) - 1 \right],
	\label{eq20}
\end{equation}

\begin{equation}
	I_{sh} = \frac{I_L \cdot R_s + V_L}{R_{sh}},
	\label{eq21}
\end{equation}
where $I_L$ indicates the output current;
$I_d$ is the diode current;
$V_L$ denotes the output voltage;
$T$ is the temperature in Kelvin;
$q$ represents the electronic charge \((1.60217646 \times 10^{-19})\);
$k$ indicates the Boltzmann’s constant $(1.3806503 \times 10^{-23} \text{ J/k})$.
The final expression for the output current in SDM is:

\begin{equation}
	I_L = I_{ph} - \frac{I_L \cdot R_s + V_L}{R_{sh}} - I_{sd} \cdot \left[\exp\left(\frac{(I_L \cdot R_s + V_L) \cdot q}{T \cdot n \cdot k}\right) - 1\right],
	\label{eq22}
\end{equation}

\subsubsection*{(2) Double Diode Model }
\addcontentsline{toc}{subsubsection}{(2). Double Diode Model }
\label{B_DDM}

Compared to SDM, DDM accounts for recombination losses in the depletion region,
providing a more precise representation of PV system behavior.
It includes seven parameters to be identified: photogenerated current (${I_{ph}}$),
diffusion current ($I_{sd1}$), saturation current ($I_{sd2}$), series resistance ($R_s$), shunt resistance ($R_{sh}$),
and deality factors of the two diodes ($n_{1},n_{2}$).
The governing equation of DDM is given as:

\begin{equation}
	\begin{aligned}
		I_L & =  I_{ph} - I_{sh} - I_{d1} - I_{d2} \\
		& = I_{ph} - \frac{I_L \cdot R_s + V_L}{R_{sh}} - I_{sd1} \cdot \left[ \exp\left(\frac{(I_L \cdot R_s + V_L) \cdot q}{T \cdot n_1 \cdot k}\right) - 1 \right] \\
		&\quad - I_{sd2} \cdot \left[ \exp\left(\frac{(I_L \cdot R_s + V_L) \cdot q}{T \cdot n_2 \cdot k}\right) - 1 \right],
	\end{aligned}
	\label{eq23}
\end{equation}

\subsubsection*{(3) PV module Model }
\addcontentsline{toc}{subsubsection}{(2). PV module Model }
\label{C_PV}

In practical applications,
PV cells are connected in series ($N_S$) and parallel ($N_p$) configurations to form PV modules,
enabling higher voltage and current outputs. The mathematical expression of the PV system is given by:

\begin{equation}
	\begin{aligned}
		\frac{I_L}{N_p} = I_{ph} - I_{sd} \cdot \left[ \exp\left( \frac{ \left( R_S \cdot \frac{I_L}{N_p}+ \frac{V_L}{N_S} \right) \cdot q }{T \cdot n \cdot k} \right) - 1 \right] - \frac{ R_S \cdot \frac{I_L}{N_p} + \frac{V_L}{N_S} }{R_{sh}},\\
	\end{aligned}
	\label{eq24}
\end{equation}
where $V_L$ and $I_L$ are the output voltage and current of the PV module, respectively.
The bounds of different parameters in three PV models are reported in \autoref{Parameter range}.

\begin{table}[htbp]
	\captionsetup{
		labelsep=newline,
		justification=raggedright,
		singlelinecheck=false,
		labelfont=bf,
		skip=0pt,
		margin=0pt
	}
	\caption{Parameter range for the single and double diode models, and PV module model.}
	\label{Parameter range}
    \footnotesize
	\begin{tabular}{p{3.5cm}p{2cm}p{2cm}p{2cm}p{2cm}}
		\toprule
		\multirow{2}{*}{Parameter} & \multicolumn{2}{c}{SDM/DDM} & \multicolumn{2}{c}{PV module} \\
		\cmidrule(lr){2-3} \cmidrule(lr){4-5}
		& LB & UB & LB & UB \\
		\midrule
		$I_{ph}$ (A)            & 0   & 1     & 0   & 2    \\
		$I_{sd}, I_{sd1}, I_{sd2}$ ($\mu$A) & 0   & 1     & 0   & 50   \\
		$R_S$ ($\Omega$)         & 0   & 0.5   & 0   & 2    \\
		$R_{sh}$ ($\Omega$)      & 0   & 100   & 0   & 2000 \\
		$n, n_1, n_2$            & 1   & 2     & 1   & 50   \\
		\bottomrule
	\end{tabular}
	
\end{table}

In our tests, the Root Mean Square Error (RMSE) is employed as the fitness function to evaluate the accuracy of the identified parameters,
formulated as

\begin{equation}
	\text{RMSE}(X) = \sqrt{\frac{1}{N} \sum_{i=1}^N f_i^2(X, I_L, V_L)},
	\label{eq25}
\end{equation}
where $ N $ is the total number of benchmark measurement points;
$X$ is the vector of unknown parameters to be identified.
The aim of parameter identification is to determine the optimal parameter set $X$ that minimizes RMSE.
In short, the parameter identification problems can be expressed by
\begin{equation}
	\text{SDM:} \begin{cases}
		\begin{aligned}
			f_i(X, I_L, V_L) & = I_{ph} - I_{sd} \cdot \left[ \exp\left(\frac{(I_L \cdot R_s + V_L) \cdot q}{T \cdot n \cdot k}\right) - 1 \right] \\
			&\quad - \frac{I_L \cdot R_s + V_L}{R_{sh}} - I_L,\\
			X & = \{I_{ph}, I_{sd}, R_s, R_{sh}, n\}.
		\end{aligned}
		\label{eq26}
	\end{cases}
\end{equation}

\begin{equation}
	\text{DDM:}
	\begin{cases}
		\begin{aligned}
			f_i(X, I_L, V_L) &=  I_{ph} - I_{sd1} \cdot \left[ \exp\left(\frac{(I_L \cdot R_s + V_L) \cdot q}{T \cdot n_1 \cdot k}\right) - 1 \right] \\
			&\quad - I_{sd2} \cdot \left[  \exp\left(\frac{(I_L \cdot R_s + V_L) \cdot q}{T \cdot n_2 \cdot k}\right) - 1 \right] \\
			&\quad - \frac{I_L \cdot R_s + V_L}{R_{sh}} - I_L, \\
			X &= \{I_{ph}, I_{sd1}, R_s, R_{sh}, n_1, I_{sd2}, n_2\}.
		\end{aligned}
		\label{eq27}
	\end{cases}
\end{equation}

\begin{equation}
	\text{PV Module:}
	\begin{cases}
		\begin{aligned}
			f_i(X, I_L, V_L) &=  I_{ph} \cdot N_p - \frac{ \left( R_S \cdot \frac{I_L}{N_p} + \frac{V_L}{N_S} \right)}{ R_{sh} / N_p } - I_L\\
			& - I_{sd} \cdot  N_p \cdot \left[ \exp\left( \frac{ \left( R_S \cdot \frac{I_L}{N_p} + \frac{V_L}{N_S}   \right) \cdot q}{T \cdot n \cdot k} \right) - 1 \right] ,\\	
			X & = \{I_{ph}, I_{sd}, R_s, R_{sh}, n\}.
		\end{aligned}
		\label{eq28}
	\end{cases}
\end{equation}

\subsection{Result analysis} \label{subsec5.2}

The I-V data is from a 57 mm diameter RTC France silicon solar cell under standard test conditions,
with an irradiance of 1000 W/m$^2$ and a temperature of 33$^\circ$C,
and has been extensively utilized in PV parameter identification research~\citep{yu2017parameters}.
The optimization process is conducted by employing 30 search agents over 500 iterations,
with performance comparisons against 16 competing algorithms.
Numerical results are based on 30 independent runs,
with the best outcomes summarized in~\textcolor{blue}{\Cref{SDM,DDM,PV}} for SDM, DDM, and PV module model, respectively.

\begin{table}[H]
	\captionsetup{
		labelsep=newline,
		justification=raggedright,
		singlelinecheck=false,
		labelfont=bf,
		skip=0pt,
		margin=0pt
	}
	\caption{Comparison of outcomes achieved by 17 methods on SDM.}
	\label{SDM}
	\small
	\resizebox{\columnwidth}{!}{%
		\begin{tabular}{p{7em}p{5em}p{5em}p{5em}p{5em}p{5em}l}
			%\begin{tabular}{lllllll}
			\toprule
			Algorithm & $I_{ph}$ (A) & $I_{sd}$ ($\mu$A) & $R_s$ ($\Omega$) & $R_{sh}$ ($\Omega$) & n & RMSE \\
			\midrule
			MFO-DBO & 0.76078  & 0.32302  & 0.03638  & 53.71795  & 1.48118  & \textbf{9.86022E-04} \\
			EBOwithCMAR & 0.76078  & 0.32302  & 0.03638  & 53.71853  & 1.48118  & 9.86022E-04 \\
			GODBO & 0.76074  & 0.33940  & 0.03617  & 55.21647  & 1.48618  & 9.90636E-04 \\
			MsDBO & 0.76081  & 0.30721  & 0.03658  & 51.97400  & 1.47616  & 9.91727E-04 \\
			EDBO & 0.76074  & 0.36172  & 0.03591  & 56.94046  & 1.49267  & 1.00979E-03 \\
			IDBO & 0.76085  & 0.26030  & 0.03731  & 49.15435  & 1.45974  & 1.07642E-03 \\
			QHDBO & 0.76087  & 0.38138  & 0.03568  & 56.18353  & 1.49813  & 1.04262E-03 \\
			DBO & 0.76092  & 0.25534  & 0.03729  & 47.69290  & 1.45789  & 1.08234E-03 \\
			FOSSA & 0.76099  & 0.28611  & 0.03698  & 48.51853  & 1.46915  & 1.08340E-03 \\
			LSHADE & 0.76062  & 0.41932  & 0.03527  & 61.73187  & 1.50793  & 1.10892E-03 \\
			HLOA & 0.76002  & 0.37807  & 0.03595  & 74.35342  & 1.49697  & 1.17681E-03 \\
			HEOA & 0.76164  & 0.47177  & 0.03456  & 52.16631  & 1.52053  & 1.40002E-03 \\
			SCWOA	&0.76220 & 0.59537 & 0.03353  & 59.66722  &1.54571 & 1.95507E-03\\	
			FOPSO & 0.76117  & 0.99826  & 0.02931  & 64.02412  & 1.60475  & 4.05537E-03 \\
			FOFPA & 0.75271  & 0.60337  & 0.03118  & 48.19111  & 1.54946  & 7.70500E-03 \\
			MDBO & 0.31659  & 0.01171  & 0.04984  & 27.94674  & 1.20889  & 2.28084E-02 \\
			IGWO & 0.77445  & 0.26603  & 0.05128    & 49.52725  & 1.45234 & 4.07820E-02	\\
					
			\bottomrule
		\end{tabular}
	}
\end{table}

\begin{table}[H]
	\centering
	\captionsetup{
		labelsep=newline,
		justification=raggedright,
		singlelinecheck=false,
		labelfont=bf,
		skip=0pt,
		margin=0pt
	}
	\caption{Comparison of outcomes achieved by 17 methods on DDM.}
	\label{DDM}
	\small
	\resizebox{\columnwidth}{!}{%
		\begin{tabular}{lllllllll}
			\toprule
			Algorithm & $I_{ph}$ (A) & $I_{sd1}$ ($\mu$A) & $R_s$ ($\Omega$) & $R_{sh}$ ($\Omega$) & $n_1$ & $I_{sd2}$ ($\mu$A) & $n_2$ & RMSE \\
			\midrule
			MFO-DBO & 0.76077  & 0.80672  & 0.03673  & 55.95082  & 2.00000  & 0.22164  & 1.44953  & \textbf{9.82673E-04} \\
			EBOwithCMAR & 0.76078  & 0.22780  & 0.03667  & 55.19751  & 1.45251  & 0.53862  & 1.91967  & 9.83443E-04 \\
			MsDBO & 0.76079  & 0.24676  & 0.03670  & 53.50088  & 1.71796  & 0.21067  & 1.44947  & 9.87022E-04 \\
			GODBO & 0.76084  & 0.27779  & 0.03667  & 52.01491  & 1.46808  & 0.10059  & 1.82577  & 9.91893E-04 \\
			LSHADE & 0.76063  & 0.21236  & 0.03633  & 57.77256  & 1.98844  & 0.30967  & 1.47845  & 9.99838E-04 \\
			IDBO & 0.76060  & 0.34720  & 0.03607  & 57.06767  & 1.48857  & 0.01149  & 1.97826  & 1.00001E-03 \\
			EDBO & 0.76059  & 0.21579  & 0.03639  & 61.13749  & 1.45107  & 0.57667  & 1.85040  & 1.01061E-03 \\
			DBO & 0.76082  & 0.01679  & 0.03837  & 57.12710  & 1.27320  & 0.95690  & 1.70525  & 1.03624E-03 \\
			FOSSA & 0.76016  & 0.22109  & 0.03770  & 56.26268  & 1.44826  & 0.02625  & 1.54154  & 1.25163E-03 \\
			HLOA & 0.76056  & 0.48651  & 0.03495  & 72.66145  & 1.52737  & 0.00295  & 1.35613  & 1.26730E-03 \\
			QHDBO & 0.76140  & 0.26800  & 0.03794  & 40.86966  & 1.83409  & 0.16748  & 1.42267  & 1.29323E-03 \\
			SCWOA	& 0.75826 	& 0.27158 	& 0.03695 	& 99.15510 	& 1.46451 	& 0.14480 	& 1.99427 & 1.89312E-03\\
			HEOA & 0.76156  & 0.88650  & 0.03016  & 92.34183  & 1.90220  & 0.88368  & 1.60333  & 3.08185E-03 \\
			FOPSO & 0.76441  & 0.02352  & 0.02874  & 37.12485  & 2.00000  & 0.99866  & 1.60596  & 4.96723E-03 \\
			MDBO & 0.37123  & 0.23582  & 0.02272  & 37.64344  & 1.49092  & 0.14937  & 1.43788  & 1.78308E-02 \\
			FOFPA & 0.78613  & 0.40198  & 0.03145  & 13.06438  & 1.64247  & 0.54256  & 1.57580  & 1.42955E-02 \\
			IGWO & 0.73561 	& 0.01417 	& 0.04175 	& 58.54045 	& 1.29970  & 0.59287   & 1.61960 & 4.36645E-02 \\
					
			\bottomrule
		\end{tabular}
	}
\end{table}

\begin{table}[H]
	\captionsetup{
		labelsep=newline,
		justification=raggedright,
		singlelinecheck=false,
		labelfont=bf,
		skip=0pt,
		margin=0pt
	}
	\caption{Comparison of outcomes achieved by 17 methods on PV module model.}
	\label{PV}
	\small
	\resizebox{\columnwidth}{!}{%
		\begin{tabular}{p{7em}p{5em}p{5em}p{5em}p{5em}p{5em}l}
			%\begin{tabular}{lllllll}
			\toprule
			Algorithm & $I_{ph}$ (A) & $I_{sd}$ ($\mu$A) & $R_s$ ($\Omega$) & $R_{sh}$ ($\Omega$) & n & RMSE \\
			\midrule
			MFO-DBO & 0.20611  & 0.70133  & 2.00000  & 1626.26441  & 16.22338  & \textbf{2.42563E-03} \\
			EBOwithCMAR & 0.20607  & 0.70925  & 1.99888  & 1680.61871  & 16.23745  & 2.42576E-03 \\
			IDBO & 0.20626  & 0.88735  & 1.96135  & 2000.00000  & 16.52733 & 2.73405E-03 \\
			DBO & 0.20666  & 0.63609  & 2.00000  & 1305.44073  & 16.09759  & 3.07627E-03 \\
			HLOA & 0.20643  & 0.81711  & 1.94232  & 1190.92709  & 16.43104  & 3.21675E-03 \\
			GODBO & 0.20637  & 0.84350  & 2.00000  & 1994.47362  & 16.45862  & 3.25876E-03 \\
			MsDBO & 0.20643 & 1.49156  & 1.84561  & 1997.11001  & 17.25620  & 3.72045E-03 \\
			LSHADE & 0.20666  & 2.43352  & 1.71418  & 2000.00000  & 18.00430  & 5.49085E-03 \\
			EDBO & 0.20698  & 3.03864  & 1.66300  & 2000.00000  & 18.36111  & 6.31868E-03 \\
			HEOA & 0.20813  & 2.42853  & 1.78856  & 1257.85575  & 18.00048  & 7.97667E-03 \\
			SCWOA	& 0.20860  & 6.36877   & 1.42760    & 1273.78762 	& 19.67595 & 1.03551E-02  \\
			QHDBO & 0.20931  & 3.89023  & 1.51020  & 715.64766  & 18.80691  & 1.03616E-02 \\
			FOSSA & 0.20775  & 4.42514  & 1.41024  & 976.87289  & 19.03004  & 1.12864E-02 \\
			MDBO & 0.20889  & 9.37119  & 0.60831  & 888.79466  & 20.44320  & 1.46626E-02 \\
			FOPSO & 0.21191  & 22.99654  & 0.79706  & 1130.69213  & 22.41777  & 2.28079E-02 \\
			FOFPA & 0.20472  & 29.14231  & 0.80037  & 940.66454  & 23.15895  & 2.77964E-02 \\
			IGWO	&0.19472 	&48.75407 	&0.28936 	&1970.28066 & 24.44968 & 7.63825E-02	\\
			\bottomrule
		\end{tabular}
	}
\end{table}

%\begin{itemize} [labelsep=0.5em, leftmargin=2.2em, itemsep=0pt, topsep=0pt, parsep=0pt, partopsep=0pt]
%	\setlength{\itemsep}{0pt}
%	
%\end{itemize}

%In short, for all three models (SDM, DDM, and PV),
%MFO-DBO shows strong competitiveness.
%Additionally, when compared with the algorithms in~\citep{yuan2023improved,wang2025dynamic},
%MFO-DBO still exhibits strong competitiveness,
%further validating the effectiveness of the proposed method.
From \textcolor{blue}{\Cref{SDM,DDM,PV}}, we have the following observations.
\begin{itemize} [labelsep=0.5em, leftmargin=2.2em, itemsep=0pt, topsep=0pt, parsep=0pt, partopsep=0pt]
	\setlength{\itemsep}{0pt}
	\item For SDM, MFO-DBO and EBOwithCMAR both achieve the lowest RMSE,
	with MFO-DBO reaching $9.86022 \times 10^{-4}$—an 8.90\% reduction compared to DBO.
	This result highlights the effectiveness of the proposed improvements in enhancing parameter identification accuracy.
	\item For DDM, MFO-DBO achieves the best result among all algorithms,
	obtaining an RMSE of $9.82673 \times 10^{-4}$ and reducing the error by 5.17\% relative to DBO.
	As model complexity increases, DBO tends to get trapped in local optima, whereas MFO-DBO maintains robust performance.
	\item For the PV model, MFO-DBO achieves the lowest RMSE of $2.42563 \times 10^{-3}$,
	significantly outperforming DBO by 21.14\%. This confirms the strong adaptability and practical value of the proposed enhancements in real-world applications.
\end{itemize}

\begin{figure}[H]
	\centering
	\begin{minipage}{0.45\textwidth}
		\centering
		\includegraphics[width=\textwidth]{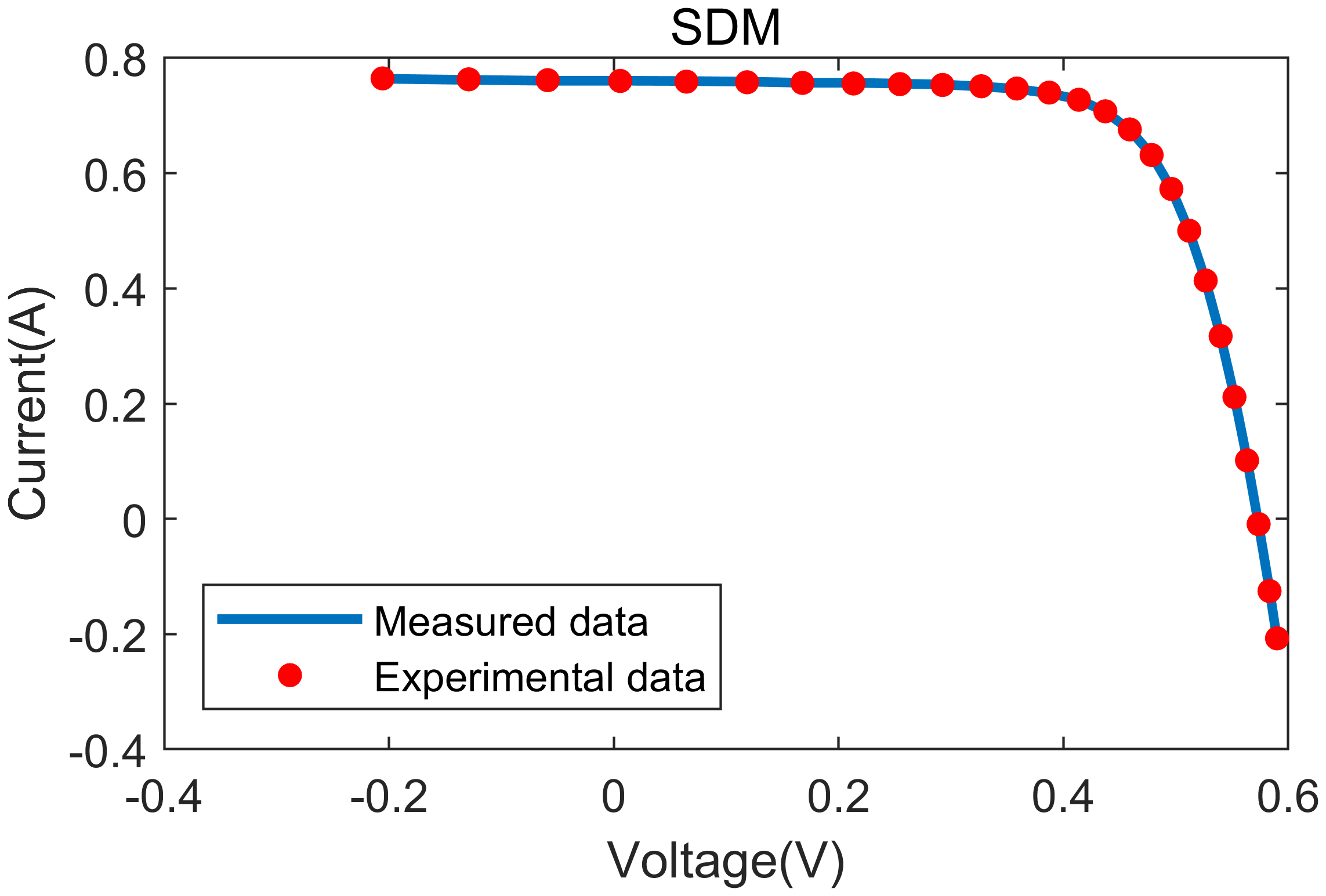}
		\subcaption{\textit{I-V} characteristics}
	\end{minipage}%
	\begin{minipage}{0.45\textwidth}
		\centering
		\includegraphics[width=\textwidth]{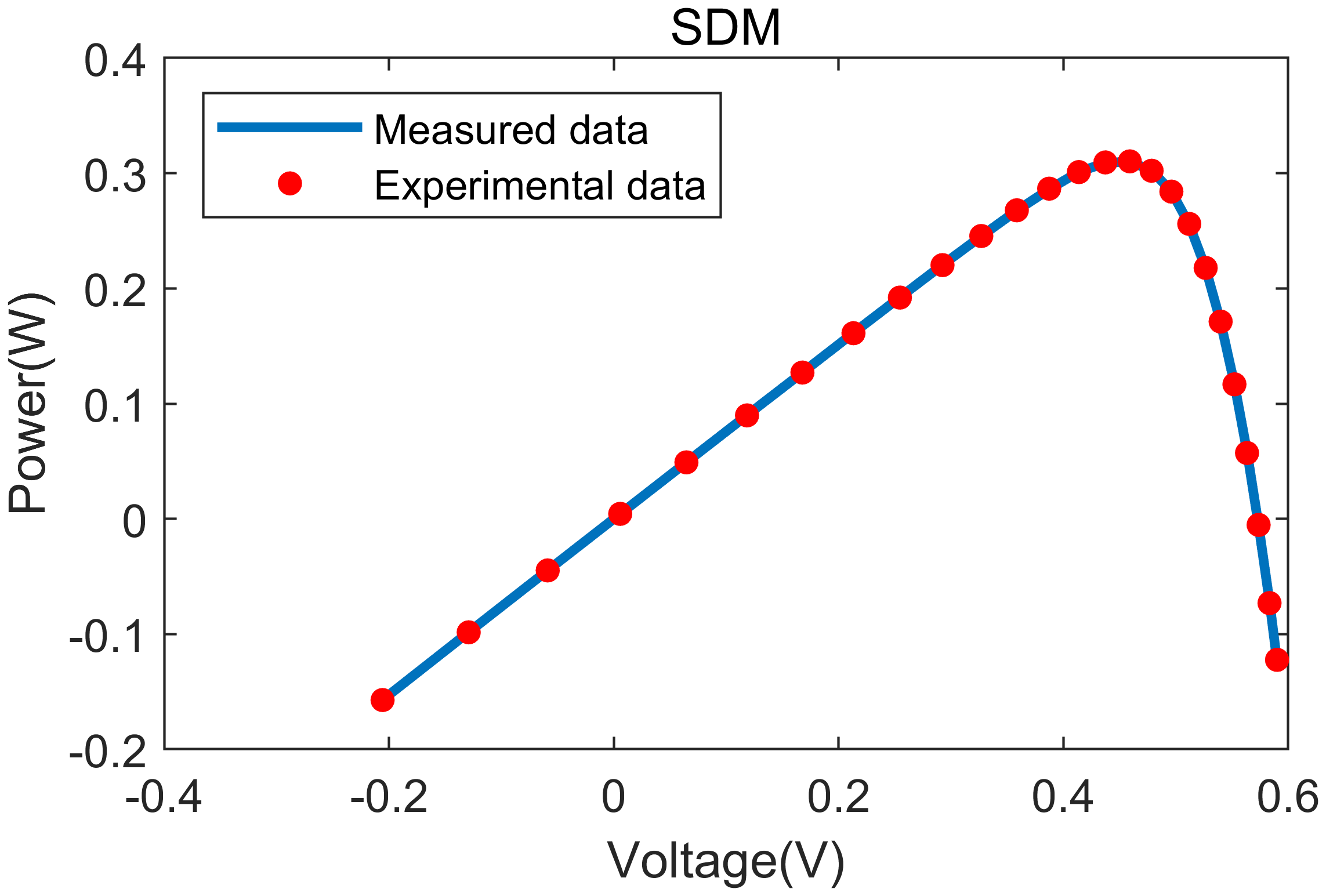}
		\subcaption{\textit{P-V} characteristics}
	\end{minipage}
	
	\caption{Measured and MFO-DBO simulated data for SDM.}
	\label{figSDM}
\end{figure}

\begin{figure}[H]
	\centering
	\begin{minipage}{0.45\textwidth}
		\centering
		\includegraphics[width=\textwidth]{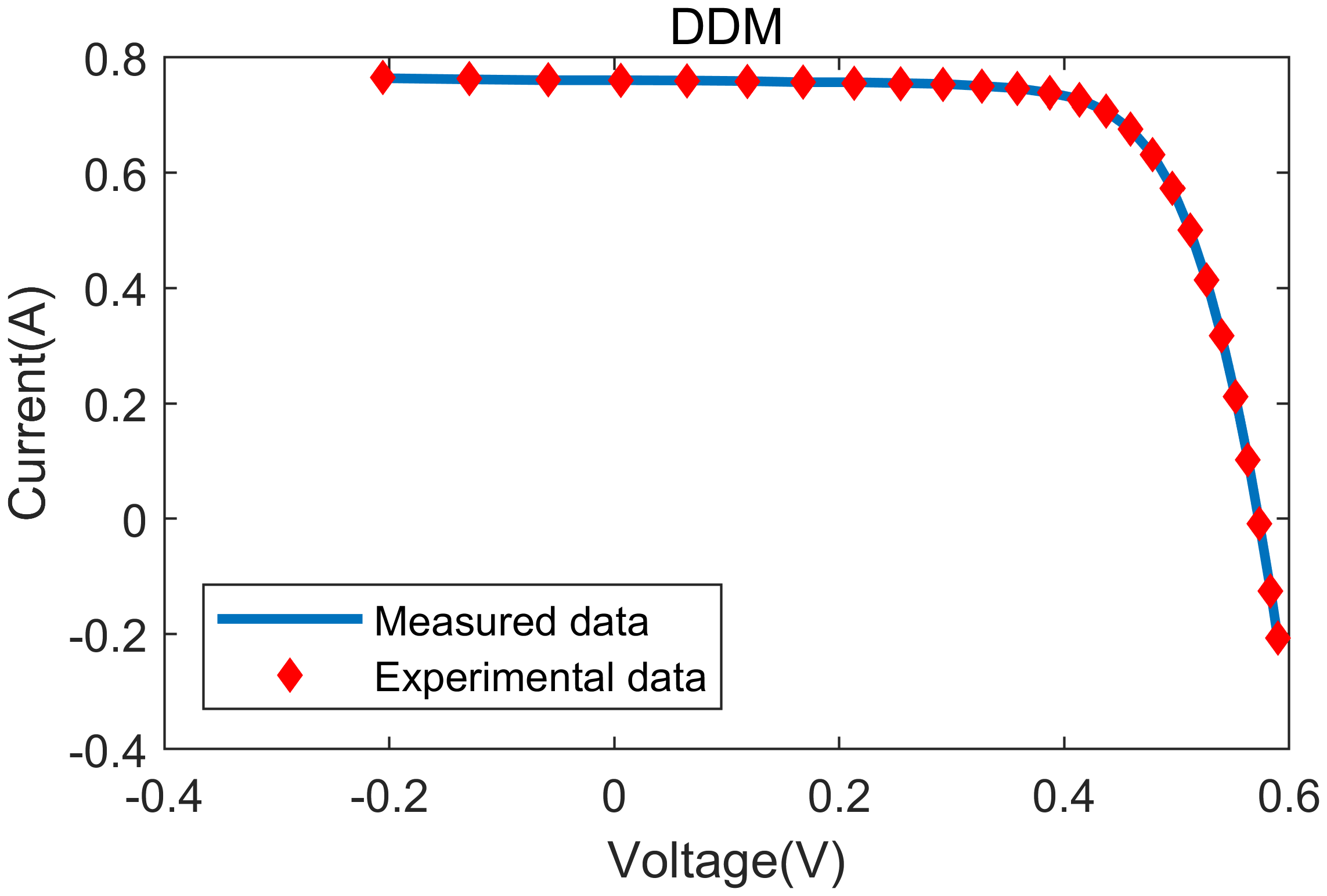}
		\subcaption{\textit{I-V} characteristics}
	\end{minipage}%
	\begin{minipage}{0.45\textwidth}
		\centering
		\includegraphics[width=\textwidth]{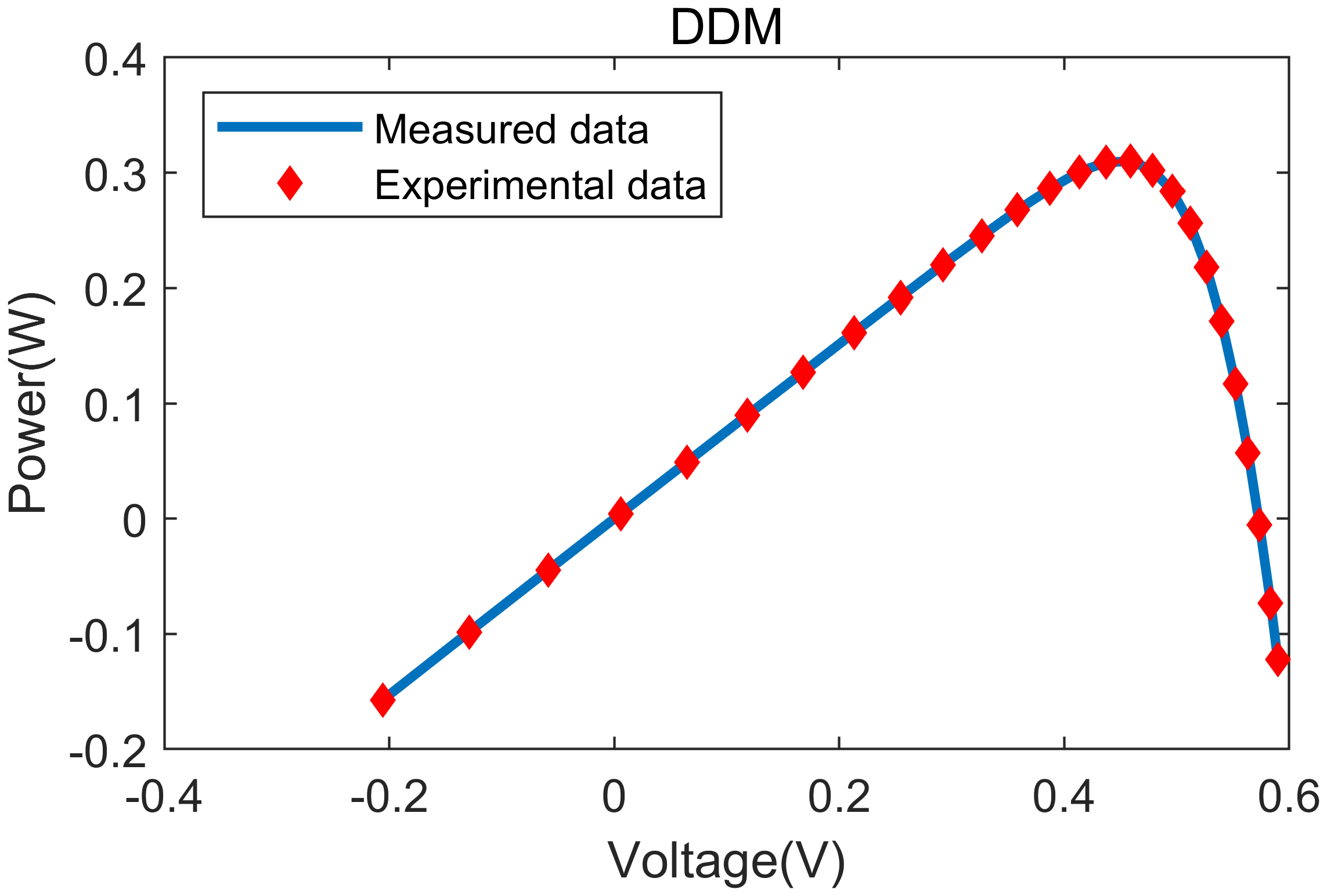}
		\subcaption{\textit{P-V} characteristics}
	\end{minipage}
	
	\caption{Measured and MFO-DBO simulated data for DDM.}
	\label{figDDM}
\end{figure}

\begin{figure}[H]
	\centering
	\begin{minipage}{0.45\textwidth}
		\centering
		\includegraphics[width=\textwidth]{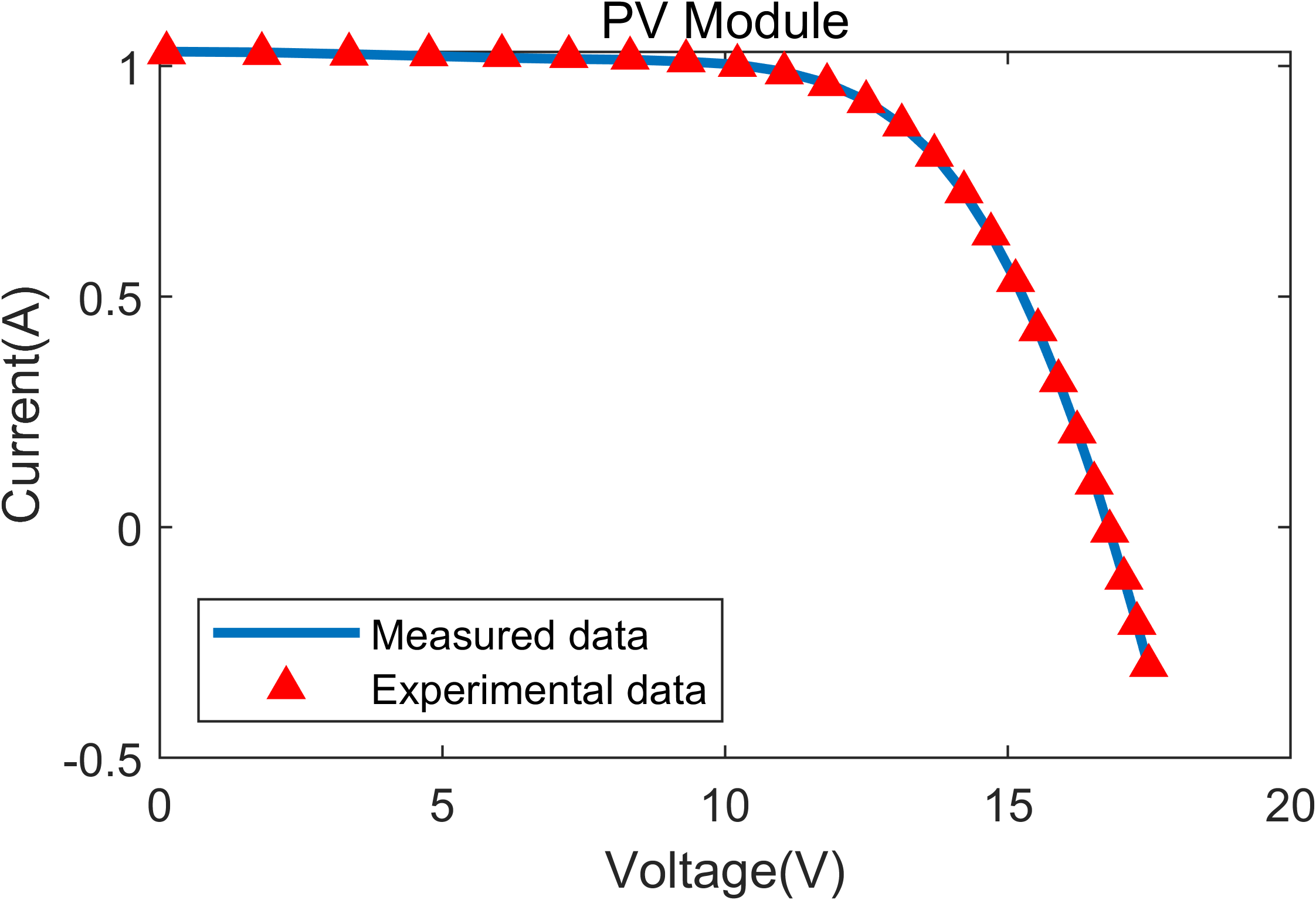}
		\subcaption{\textit{I-V} characteristics}
	\end{minipage}%
	\begin{minipage}{0.45\textwidth}
		\centering
		\includegraphics[width=\textwidth]{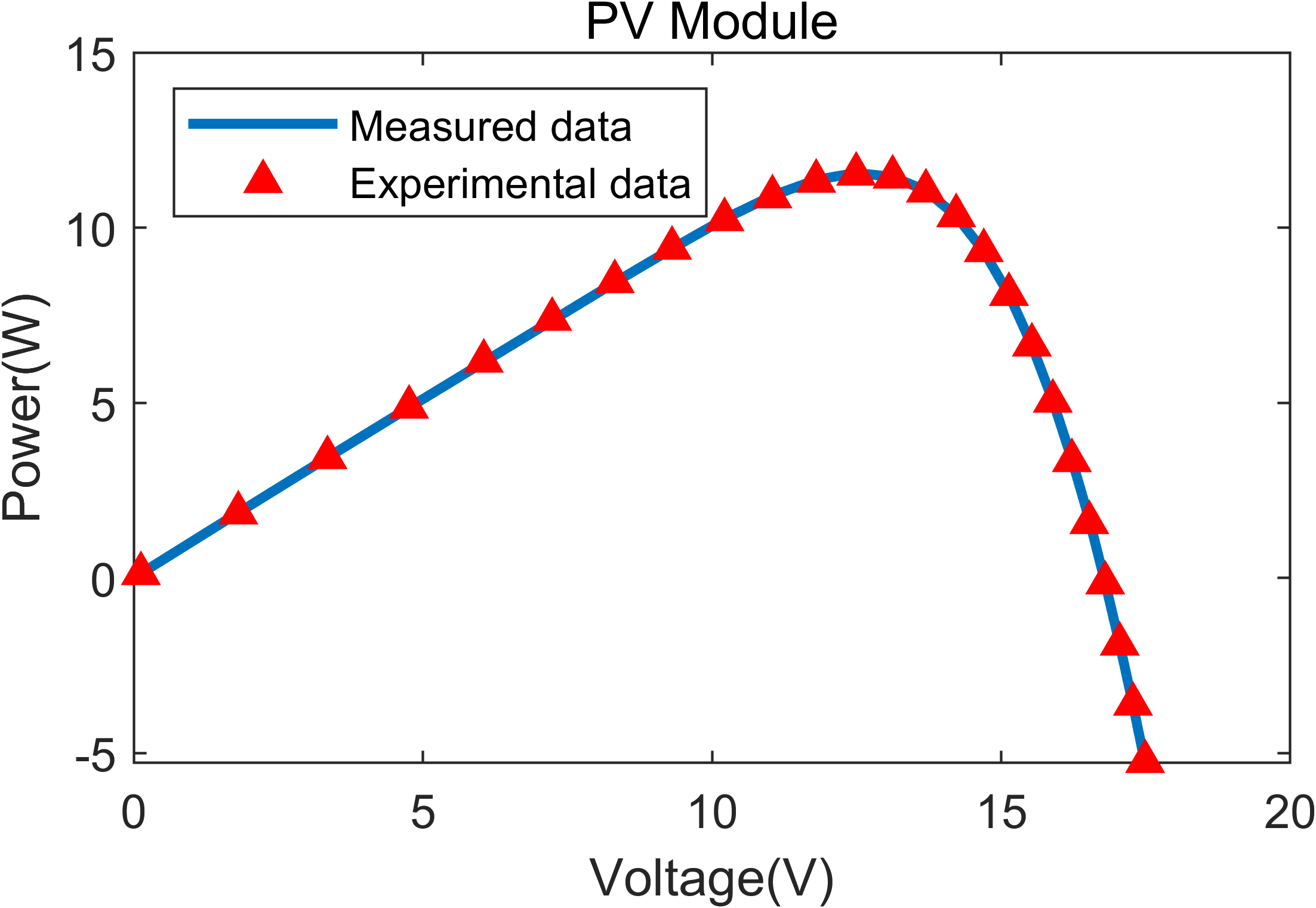}
		\subcaption{\textit{P-V} characteristics}
	\end{minipage}
	
	\caption{Measured and MFO-DBO simulated data for PV Module.}
	\label{figPV}
\end{figure}

In addition,
\Cref{figSDM,figDDM,figPV} illustrate the comparison between measured I-V data and the experimental data obtained by MFO-DBO.
As seen in these figures,
the experimental curves generated by MFO-DBO closely match the measured data,
confirming the accuracy and competitiveness of the algorithm in real-world PV parameter identification tasks.

\section {Concluding Remarks} \label{sec6}

This paper proposed a Memory-enhanced Fractional-Order Dung Beetle Optimization (MFO-DBO) algorithm to address the limitations of the classic DBO.
The main contributions can be summarized as follows:
(1) \textbf{Methodological level}: Three coordinated enhancements were introduced, including
Fractional-Order (FO) calculus to introduce long-term in the ball-rolling phase,
Fractional-Order Logistic Chaotic (FOLC) mapping for improved population diversity,
and Chaotic Perturbation (CP) to enhance global search capabilities.
(2) \textbf{Performance level}: Numerical results on the CEC2017 benchmark suite show that MFO-DBO achieves superior convergence speed, solution accuracy, and robustness compared to other advanced algorithms, while maintaining a well-balanced exploration–exploitation balance;
(3) \textbf{Application level}: The MFO-DBO algorithm demonstrates strong practical value in photovoltaic (PV) parameter identification, outperforming existing approaches in accuracy and stability.

Next, we will focus on two main directions: (i) \textit{Cross-domain applications}—leveraging the memory-preserving properties of fractional-order dynamics, the proposed algorithm holds strong potential for tasks such as fractional-order system control, chaotic time-series prediction, and parameter identification in nonlinear fractional-order chaotic systems;
(ii) \textit{Algorithmic improvements}—future extensions may include adaptive tuning of the fractional-order parameter $\delta$ and memory depth $m$, as well as integration of learning-based strategies (e.g., neural networks) to dynamically adjust perturbation strength and enhance search behavior in highly complex landscapes.

\section*{Acknowledgement}

This work was supported by National Natural Science Foundation of China (Nos. 12301405, 12271419), Shaanxi Fundamental Science Research Project for Mathematics and Physics (No. 23JSZ010)
and Fundamental Research Funds for Central Universities of China (No. ZYTS25201).

\bibliographystyle{elsarticle-num}
\bibliography{references}
\end{document}